\documentclass[10pt,twocolumn,letterpaper]{article}

\usepackage[pagenumbers]{cvpr} 

\usepackage[dvipsnames]{xcolor}

\usepackage{multirow}
\usepackage{makecell}
\usepackage{todonotes}
\usepackage{cuted}
\usepackage{tabularx}
\usepackage{chngcntr}
\usepackage{colortbl}
\usepackage[accsupp]{axessibility} 

\definecolor{cvprblue}{rgb}{0.21,0.49,0.74}
\usepackage[pagebackref,breaklinks,colorlinks,citecolor=cvprblue,hypertexnames=false]{hyperref}

\def\paperID{629} 
\def\confName{CVPR}
\def\confYear{2024}
\def\name{PanFusion}

\title{Taming Stable Diffusion for Text to 360$^{\circ}$ Panorama Image Generation}

\author{
    Cheng Zhang$^{1,3}$ \quad
    Qianyi Wu$^{1}$ \quad
    Camilo Cruz Gambardella$^{1,3}$ \quad
    Xiaoshui Huang$^{2}$\thanks{Corresponding author.}
    \\
    Dinh Phung$^{1}$ \quad
    Wanli Ouyang$^{2}$ \quad
    Jianfei Cai$^{1}$\footnotemark[1]
    \\
    $^{1}$Monash University
    $^{2}$Shanghai AI Laboratory
    $^{3}$Building 4.0 CRC, Caulfield East, Victoria, Australia
}

\begin{document}
\maketitle

\begin{strip}
    \vspace{-3em}
	\centering
	\small
    \newcommand{\rot}[1]{\rotatebox[origin=c]{90}{#1}}
    \newcommand{\pano}[1]{\raisebox{-0.5\height}{\includegraphics[width=.49\linewidth,clip,trim=-10 -5 -10 -5]{#1}}}
    \setlength\tabcolsep{0px}
    \begin{tabular}{rcc}
        \rot{MVDiffusion}
        & \pano{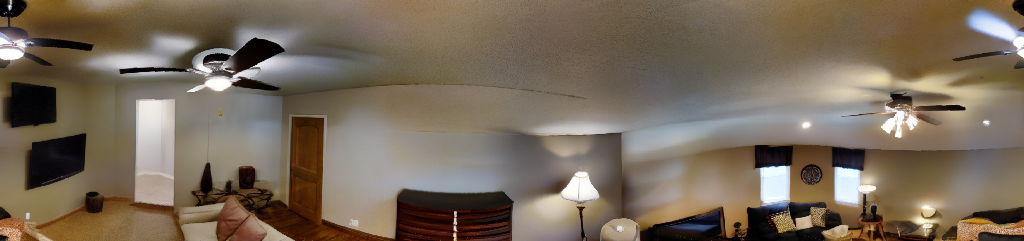}
        & \pano{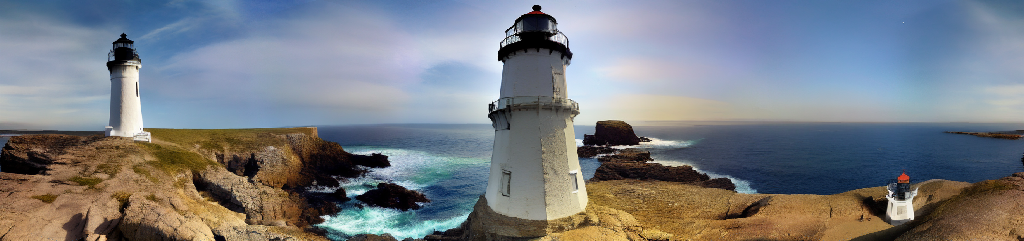}
        \\
        \rot{PanFusion (Ours)}
        & \pano{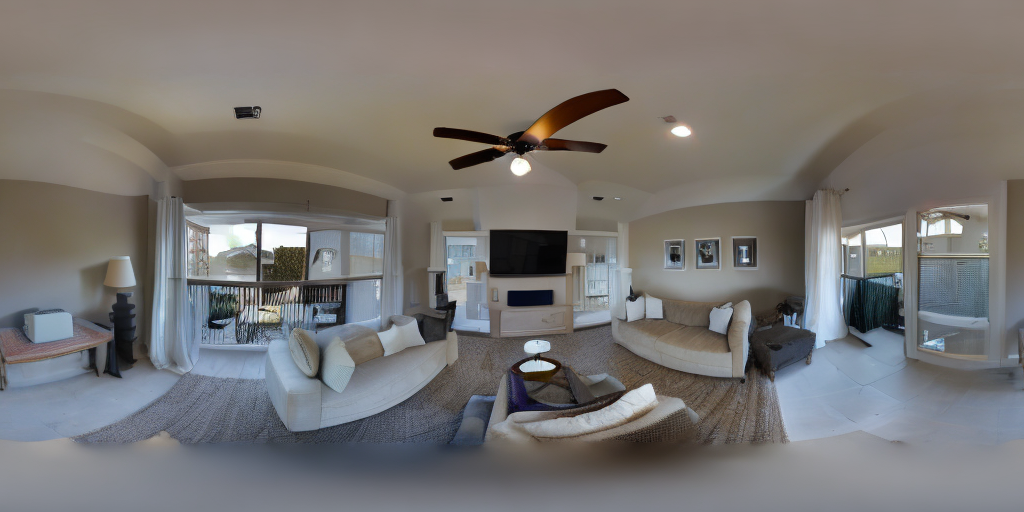}
        & \pano{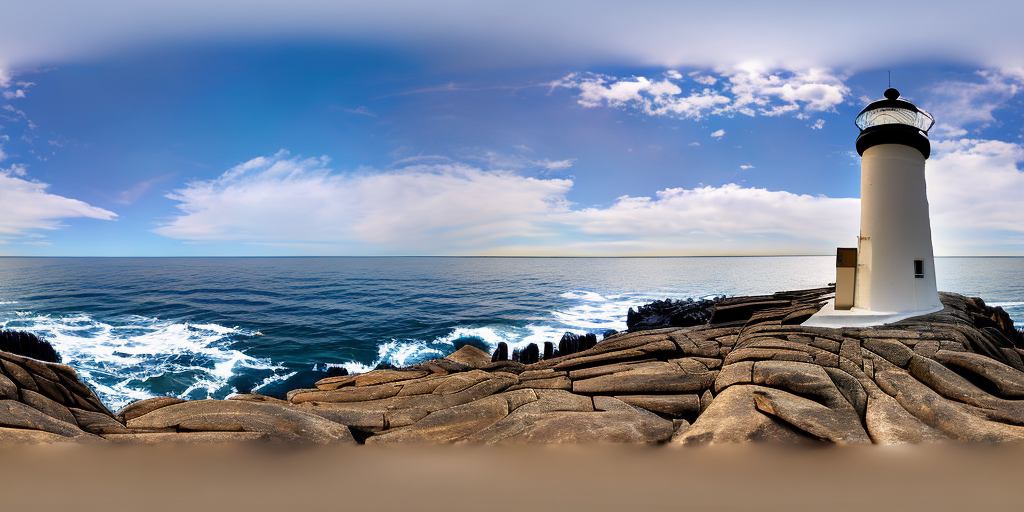}
        \\
        & ``A living room with a ceiling fan.''
        & \makecell{
            ``A historic lighthouse perched on a rugged coastline,\\
            overlooking the vast expanse of the open sea.''
        }
    \end{tabular}
    \vspace{-0.5em}
	\captionof{figure}{
        Our {\name} can generate realistic and consistent $360^\circ$ horizontal by $180^\circ$ vertical FOV panoramas from a single text prompt, compared to the limited FOV of current state-of-the-art method MVDiffusion~\cite{tang2023mvdiffusion}.
        Left: {\name} addresses the problem of repetitive elements (duplicated ``ceiling fans'') and inconsistency (the ceiling and wall in the center) of MVDiffusion.
        Right: While trained mostly on indoor scenes, {\name} can generalize well to out-of-domain outdoor prompts with more reasonable layout.
        }
        \label{fig:teaser}
    \vspace{-0.5em}
\end{strip}

\begin{abstract}
\vspace{-0.2in}
Generative models, {\eg}, Stable Diffusion, have enabled the creation of photorealistic images from text prompts.
Yet, the generation of 360-degree panorama images from text remains a challenge, particularly due to the dearth of paired text-panorama data and the domain gap between panorama and perspective images.
In this paper, we introduce a novel dual-branch diffusion model named {\name} to generate a 360-degree image from a text prompt.
We leverage the stable diffusion model as one branch to provide prior knowledge in natural image generation and register it to another panorama branch for holistic image generation.
We propose a unique cross-attention mechanism with projection awareness to minimize distortion during the collaborative denoising process.
Our experiments validate that {\name} surpasses existing methods and, thanks to its dual-branch structure, can integrate additional constraints like room layout for customized panorama outputs.
Code is available at \url{https://chengzhag.github.io/publication/panfusion}.
\end{abstract}

\vspace{-2em}
\section{Introduction}
\label{sec:intro}
%
%
%
%
Creating a $360^\circ$ panorama image from textual prompts is a nascent yet pivotal frontier in computer vision, with profound implications for applications that require extensive environmental representation, such as environmental lighting~\cite{chen2022text2light,wang2022stylelight,akimoto2022diverse}, VR/AR~\cite{yang2022neural,yang2023dreamspace}, autonomous driving~\cite{xue2018survey}, and visual navigation~\cite{li2023panogen}.
Despite recent significant strides in text-to-image synthesis, the leap to generating full $360^\circ$ horizontal by $180^\circ$ vertical field-of-view (FOV) panorama remains challenging. 

There are two major hurdles for achieving this goal.
The first hurdle is data scarcity.
The availability of text-to-panorama image pairs~\cite{tang2023mvdiffusion,lu2023autoregressive} is significantly less compared with the abundance of text-to-common image pairs~\cite{schuhmann2021laion,schuhmann2022laion}.
The dearth of data complicates the training and finetuning of generative models.
The second hurdle lies in the geometric and domain variations.
Panorama images are distinct not only in their aspect ratio ($2:1$) but also in the underlying equirectangular projection (ERP) geometry~\cite{xu2020state}.
This differs significantly from typical square images of the perspective projection that are used in most generative model training~\cite{schuhmann2021laion,schuhmann2022laion}.

To mitigate the scarcity of panorama-specific training data, the previous solutions follow a common principle that \emph{leverages the prior knowledge of the pre-trained generative model}~\cite{tang2023mvdiffusion,li2023panogen,lu2023autoregressive}.
However, taming powerful models like stable diffusion~\cite{rombach2022highresolution,runwayml} to generate high-fidelity panorama images remains a non-trivial task.
Early attempts turn to formulate the 360-degree generation as an iterative image inpainting or warping process~\cite{li2023panogen,lu2023autoregressive}.
Such solutions suffer from error accumulation and fail to handle loop closure~\cite{tang2023mvdiffusion}.
To address this, MVDiffusion~\cite{tang2023mvdiffusion} proposes to produce multiple perspective images simultaneously by introducing a correspondence-aware attention module to facilitate multiview consistency, and then stitch together the perspective images to form a complete panorama. 
Despite the improved performance, the pixel-level consistency between neighboring perspectives in MVDiffusion cannot ensure global consistency, often resulting in repetitive elements or semantic inconsistency, as illustrated in \cref{fig:teaser}. 

%

Therefore, in this paper, we propose a novel dual-branch diffusion model called \textit{\name} that is tailored to address the limitations of prior models for high-quality text to 360$^{\circ}$ panorama image generation.
Specifically, {\name} is designed to operate in both panorama and perspective domains, employing a global branch for creating a coherent panoramic ``canvas" and a local branch that focuses on rendering detail-rich multiview perspectives.
The local-global synergy of {\name} significantly improves the resulting panoramas against the prevalent issues of error propagation and visual inconsistency that prior models have struggled with. To enhance the synergy between the two branches, we further propose an Equirectangular-Perspective Projection Attention 
(EPPA) mechanism, which respects the equirectangular projection for maintaining geometric integrity throughout the generation process.
%
%
%
%
In addition, our adoption of parametric mapping for positional encoding is another leap forward, enhancing the model's spatial awareness and further ensuring the consistency of the generated panorama.
%
Moreover, the panorama branch of {\name}  
can be easily leveraged to accommodate supplementary control inputs at the panorama level, such as room layout, allowing for the creation of images that adhere to precise spatial conditions. 
We summarize our primary contributions as follows.
\begin{itemize}
    \item We pioneer a dual-branch diffusion model PanFusion, harnessing both the global panorama and local perspective latent domains, to generate high-quality, consistent $360^\circ$ panoramas from text prompts.
    \item To enhance the interaction between the two branches, we introduce an ``equirectangular-perspective projection attention'' mechanism that establishes a novel correspondence between global panorama and local perspective branches, addressing the unique projection challenges of panorama synthesis.
    \item 
    Our PanFusion not only surpasses prior models in quality and consistency but also supports extended control over the generation process with the inclusion of room layout.
    Extensive experimental results demonstrate the superiority of our proposed framework.
\end{itemize}




\begin{figure*}[t]
    \vspace{-1em}
    \centering
    \includegraphics[width=0.98\linewidth, trim={0 0 0 0}, clip]{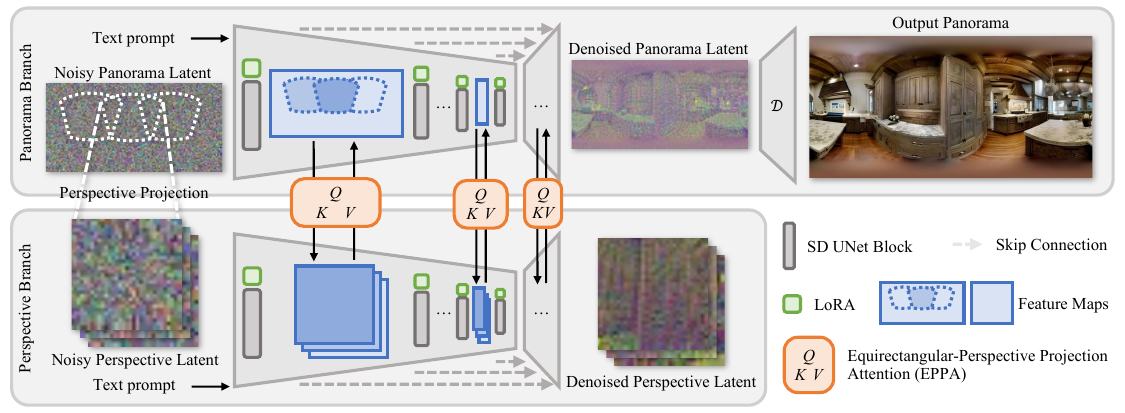}
    \vspace{-1em}
    \caption{
        Our proposed dual-branch {{\name}} pipeline.
        The panorama branch (upper) provides global layout guidance and registers the perspective information to get seamless panorama output.
        The perspective branch (lower) harnesses the rich prior knowledge of Stable Diffusion (SD) and provides guidance to alleviate distortion under perspective projection.
        Both branches employ the same UNet backbone with shared weights, while finetuned with separate LoRA layers.
        Equirectangular-Perspective Projection Attention (EPPA) modules are plugged into different layers of the UNet to pass information between the two branches.
    }\label{fig:pipeline}
    \vspace{-1em}
\end{figure*}

\section{Related Work}
\label{sec:related_work}
\noindent\textbf{Diffusion models.}
In recent years, diffusion models~\cite{yang2022diffusion,sohl2015deep,song2019generative,ho2020denoising,song2020denoising,dhariwal2021diffusion} have taken the world of image generation by storm, as they have become faster~\cite{song2020denoising,lu2022dpm,karras2022elucidating} and more capable in terms of image quality and resolution~\cite{rombach2022highresolution,saharia2022photorealistic,podell2023sdxl}.
This success has prompted the development of various applications for diffusion models, such as text-to-image~\cite{nichol2022glide,rombach2022highresolution,saharia2022photorealistic,podell2023sdxl}, image-conditioned generation~\cite{zhang2023adding,mou2023t2iadapter}, in-painting~\cite{lugmayr2022repaint,saharia2022palette} and subject-driven generation~\cite{gal2022image,ruiz2023dreambooth}.
Most of these applications try to exploit the prior knowledge of a pre-trained diffusion model to mitigate the scarcity of task-specific data, by either finetuning with techniques like LoRA~\cite{hu2021lora}, or introducing auxiliary modules to distill the knowledge.
We also adopt the same principle to harness the power of a pre-trained latent diffusion model~\cite{rombach2022highresolution} for panorama image generation.

\noindent\textbf{Panorama generation.}
Panorama image generation encompasses different settings, including panorama outpainting and text-to-panorama generation. 
Panorama outpainting~\cite{wang2022stylelight,akimoto2022diverse,wang2023360,oh2022bips,wu2023ipo,dastjerdi2022guided} focus on generating a 360-degree panorama from a partial input image. 
Various methods, such as StyleLight~\cite{wang2022stylelight} and BIPS~\cite{oh2022bips}, have addressed specific use cases, focusing on HDR environment lighting and robotic guidance scenarios. 
Recent works~\cite{wang2023360,wu2023ipo} have improved realism with diffusion models, but often lack the exploitation of rich prior information from pre-trained models, limiting generalization.
On the other hand, recent developments in generative models have opened new frontiers in synthesizing immersive visual content from textual input~\cite{hollein2023text2room,song2023roomdreamer,wu2023ipo,fang2023ctrl,wang2023prolificdreamer,wang2023perf,yang2023dreamspace,yu2023long,wang2023customizing}. 
As an image-based representation, generating panorama from text has gained much attention.
Text2Light~\cite{chen2022text2light} adopt the VQGAN~\cite{esser2021taming} structure to synthesize an HDR panorama image from text.
To generate in arbitrary resolution with pre-trained diffusion models, DiffCollage~\cite{zhange2023diffcollage}, MultiDiffusion~\cite{bar2023multidiffusion} and SyncDiffusion~\cite{lee2023syncdiffusion} propose to fuse the diffusion paths, while PanoGen~\cite{li2023panogen} solves by iteratively inpainting.
However, they failed to model the equirectangular projection of 360-degree panoramas. 
Lu et al.~\cite{lu2023autoregressive} adopts an autoregressive framework, but suffers from inefficient issues.
MVDiffusion~\cite{tang2023mvdiffusion} designs a correspondence-aware attention module to produce multi-view images simultaneously that can be stitched together but results in repetitive elements and inconsistency.
In contrast, our proposed PanFusion, a dual-branch framework, addresses the limitations of existing methods by considering both global panorama views and local perspective views, providing a comprehensive solution for text-driven 360-degree panorama image generation.

\section{Method}
\label{sec:method}


\subsection{Preliminary}

First introduced in~\cite{sohl2015deep}, Diffusion models~\cite{song2019generative, ho2020denoising, song2020denoising} aim to generate images from a noise distribution by iterative denoising with a learned prior distribution.
However, the early diffusion models operate in the image space, which is of high dimension and complex.
We have recently witnessed the huge success of latent diffusion models~\cite{rombach2022highresolution} that first transform an image ${x}$ to a latent representation ${z}$ with a learned encoder ${\mathcal{E}}$ and then train a UNet~\cite{ronneberger2015u} model ${\epsilon}_{\theta}$ parameterized by ${\theta}$ for the reverse process in the latent space, formulating the training objective as:
\begin{equation}\label{eq:ldm_loss}
    \mathcal{L} = \mathbb{E}_{{\mathcal{E}}({x}),{t},{\epsilon},{y}} \left[ || {\epsilon} - {\epsilon}_{\theta}({z}_{t},{t},\tau({y})) ||_2 \right],
\end{equation}
where ${\tau}({y})$ is an encoding of input condition ${y}$ (\eg, text prompt or image), ${z}_{t}$ is the latent map at time step $t$, and ${\epsilon}$ is sampled from a Gaussian noise.
To sample from the model, ${z}_{t}$ is first initialized from standard Gaussian distribution, then the reverse process is applied iteratively to generate ${z}_{0}$, which is finally decoded into image space with a decoder ${\mathcal{D}}$.


\subsection{Dual-Branch Diffusion Model}\label{sec:dual_branch}

Directly employing pre-trained latent diffusion models~\cite{rombach2022highresolution}, \eg, Stable Diffusion (SD)~\cite{runwayml}, to generate panorama from multiple perspective images, either in an iterative manner~\cite{li2023panogen,hollein2023text2room,fridman2023scenescape} or in a synchronized way~\cite{tang2023mvdiffusion,bar2023multidiffusion}, would fail to handle loop closure~\cite{tang2023mvdiffusion} or produce repetitive elements (\cref{fig:teaser}) due to lack of global understanding.
To address this issue, we propose a dual-branch diffusion model, which consists of a panorama branch and a perspective branch both based on the UNet of SD as shown in \cref{fig:pipeline}.
The panorama branch is designed to provide global layout guidance and to register the perspective information to get the final panorama without stitching,
while the perspective branch is designed to exploit the rich perspective image-generation capabilities of SD and to provide guidance to alleviate distortion under perspective projection.
The two branches work together during the diffusion process to generate a denoised panorama latent map.
Finally, this latent map runs through the pre-trained decoder ${\mathcal{D}}$ of SD to produce the final panorama image.

\noindent \textbf{Panorama Branch.}
Given a text prompt ${y}$, our goal is to generate a panorama ${x} \in \mathbb{R}^{3 \times {H} \times {W}}$, where ${W} = 2{H}, {H} = {512}$.
To account for the difference between the target resolution and the one that SD is trained on, we introduce LoRA~\cite{hu2021lora} layers to adapt the model to the new resolution.
SD with LoRA can already serve as a strong baseline for panorama generation, but the results are not loop-consistent.
Previous works~\cite{wu2023ipo,fang2023ctrl} have attempted to overcome this problem using 90 degree rotations of the latent map at each diffusion step.
However, the seams are still obvious~\cite{wu2023ipo}.
On close inspection of the SD model we have observed that the loop inconsistency is mainly caused by the convolutional layers in the UNet backbone, due to the lack of a mechanism to pass information between the two ends of the panorama image.
Therefore, we introduce an adaptation to the UNet by adding additional circular padding~\cite{zhuang2022acdnet,wang2023360,shum2023conditional} before each convolutional layer, and then cropping the output feature maps to the original size.
In addition, we also add circular padding to the latent map before decoding to mitigate the less apparent loop inconsistency caused by the decoder.
The combination of the techniques outlined above – latent rotation and circular padding – enable the generation of loop-consistent results with negligible computational cost, and can thus
serve as another strong baseline.
However, these alone do not make full use of the perspective generation capabilities that SD possess.

\noindent \textbf{Perspective Branch.}
The perspective branch aims to exploit the outstanding capabilities that SD has shown in the generation perspective images.
Since it does not generate panoramas directly, it can operate at a lower resolution.
Specifically, we set its resolution to $H/2 \times H/2$ and add LoRA layers to adapt to the new resolution.
To evenly distribute perspective cameras and fully cover the panorama image, we sample $N=20$ cameras with poses ${R}^{i} \in SO(3), i \in 1,\cdots,N$ on an icosahedron, similar to~\cite{rey2022360monodepth,peng2023high}, and set the $\text{FOV} = 90^\circ$.
We input the same prompt ${y}$ to both branches, and leave it to the model to decide how to exploit the prompt.

\noindent \textbf{Joint Latent Map Initialization.}
Previous work~\cite{mao2023guided} has shown that the latent map initialization can affect the layout of the generated image, \ie, the initial noise can be modified to manipulate the layout of the generated image.
We find this is particularly important for multi-image generation with correspondence, as the overlapping regions in different views tend to generate different elements if noise is sampled independently.
Under this observation, we propose to jointly sample noise for the panorama and perspective latent maps by projecting noise from the panorama to the perspective.
To be more specific, we first initialize panorama latent map ${z}_{T}^{*} \in \mathbb{R}^{4 \times {H}/{f} \times {W}/{f}}$ as Gaussian noise, then initialize perspective latent maps with ${z}_{T}^{i} = \text{P}({z}_{T}^{*}, {R}^{i}, \text{FOV}, ({H}/2{f}, {H}/2{f})), {z}_{T}^{i} \in \mathbb{R}^{4 \times {H}/2{f} \times {H}/2{f}}$, where $\text{P}(\cdot)$ is the function that projects ${z}_{T}^{*}$ to a perspective view, and ${f}$ is the down-sampling factor the encoder ${\mathcal{E}}$ used in SD.
We use nearest interpolation for projection as it introduces fewer artifacts than bilinear interpolation.

\subsection{EPP Attention}
\label{sec:epp_attention}
\begin{figure}[t]
    \centering
    \includegraphics[width=1.0\linewidth, trim={0 0 0 0}, clip]{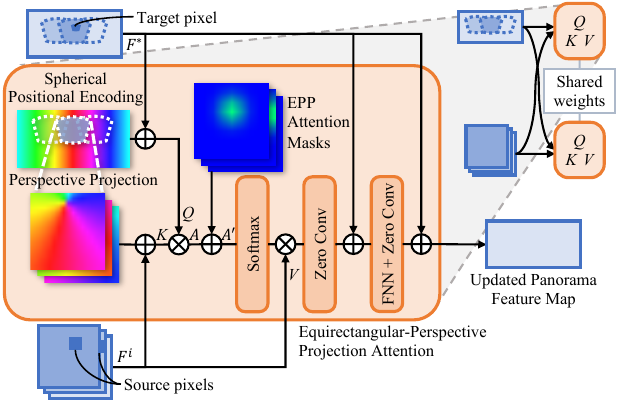}
    \vspace{-1.5em}
    \caption{
        Equirectangular-Perspective Projection (EPP) Attention. As EPP attention module is designed to be bijective to pass information in both directions, we only illustrated the direction of registering perspective information to panorama.
    }\label{fig:attention}
    \vspace{-1em}
\end{figure}


To pass the guidance between the perspective and panorama branches, a naive solution is to project the feature maps in different layers from one branch to the other, similar to~\cite{tang2023mvdiffusion}.
However, this will introduce information loss during interpolation and will restrict the receptive field to a small region around corresponding pixels.
Instead, we propose an Equirectangular-Perspective Projection Attention (EPPA) module to implicitly pass the guidance in a cross-attention way as shown in~\cref{fig:pipeline}.
The EPPA operates on feature maps in different layers of two UNet branches, denoted as ${F}^{*} \in \mathbb{R}^{{c} \times {h} \times {w}}$ and ${F}^{i} \in \mathbb{R}^{{c} \times {h}/2 \times {h}/2}$, where ${c}$ is the channel dimension, ${h}$ and ${w}$ are the height and width of the feature maps, respectively.
In addition, while cross-attention bypasses interpolation and provides benefits from the global receptive field of the attention mechanism, it is unaware of the projection between different formats.
To address this issue, as shown in~\cref{fig:attention}, we introduce two key components: EPP spherical positional encoding and EPP attention mask.

\noindent \textbf{EPP Spherical Positional Encoding.}
Since the EPPA module tries to associate two different formats, it is important to add positional information in the same space so that the attention mechanism can learn the correspondence.
We achieve this by introducing Spherical Positional Encoding (SPE)~\cite{chen2022text2light,mildenhall2021nerf} to the EPPA module.
The $\text{SPE}({\theta}, {\phi}) = \left( \gamma({\theta}), \gamma({\phi}) \right)$ function maps polar coordinates $({\theta}, {\phi})$ to a higher dimensional space $\mathbb{R}^{4L}$ with Fourier positional encoding:
\begin{equation}
    \footnotesize
    \gamma({\theta}) = \left[ \sin(2^0 \pi {\theta}), \cos(2^0 \pi {\theta}), \cdots, \sin(2^{L-1} \pi {\theta}), \cos(2^{L-1} \pi {\theta}) \right].
\end{equation}
Here we set ${L} = {c} / 4$ so that $\text{SPE}({\theta}, {\phi}) \in \mathbb{R}^{c}$.
In the EPPA module, we apply SPE by first computing an SPE map for the panorama feature map, then project it to each perspective feature map with projection function $\text{P}(\cdot)$, so that corresponding pixels in different formats share the same SPE vector.
Finally, the SPE maps are added to the feature maps accordingly and go through a linear layer to get the query ${Q}$ and key ${K}$, following a matrix product to get the affinity matrix ${A}$.
Take the perspective-to-panorama direction as an example as shown in~\cref{fig:attention}, where ${Q} \in \mathbb{R}^{{c} \times {h} \times {w}}$ and ${K} \in \mathbb{R}^{N \times {c} \times {h}/2 \times {h}/2}$ are reshaped and multiplied to get the affinity matrix ${A} \in \mathbb{R}^{{h}{w} \times {N}{h}^2/4}$.

\noindent \textbf{EPP Attention Mask.}
In addition to the SPE, we also propose an EPP attention mask to encourage the attention mechanism to focus around the corresponding pixels, inspired by~\cite{tseng2023consistent}.
For example, for a target pixel in the panorama feature map as in~\cref{fig:attention}, we would like to focus on registering the information from corresponding source pixels in the perspective feature maps.
We achieve this by enhancing the affinity matrix ${A}$ with a soft mask ${M} \in \mathbb{R}^{{h}{w} \times {N}{h}^2/4}$ highlighting the corresponding pixels.
To generate ${M}$, we first use $\text{P}(\cdot)$ to project a binary mask ${M}_{{j},{k}}$ for each pixel $({j},{k})$ in panorama feature map to each perspective view ${i}$.
Then we apply a Gaussian kernel to smooth the masks and normalized them to $\left[-1, 1\right]$.
The masks are stacked and reshaped into ${M}$, which is then added to ${A}$ to get the enhanced affinity matrix ${A}^{\prime}$.
The rest goes the same as the vanilla attention mechanism, where a softmax function is applied to ${A}^{\prime}$ to get the attention weights, which are finally multiplied with the value ${V}$ to get the output.

Inspired by~\cite{tang2023mvdiffusion,zhang2023adding}, we add zero-initialized $1 \times 1$ convolutional layers to the output of cross-attention and add it as a residual to the target feature map.
This ensures the UNet stays unmodified at the beginning of training and can be gradually adapted to the EPPA modules.
We add independent EPPA modules after each down-sampling layer and before each up-sampling layer of UNet to connect two branches, detailed in~\cref{sec:suppl_arch} of the supplementary material.
Considering that the guidance is passed in both directions following the same equirectangular-perspective projection rule, which is bijective in nature, we share the weights of the EPPA modules in the two directions.

\subsection{Layout-Conditioned Generation}\label{sec:layout_cond}
One of the most important applications for panorama generation is to generate panorama according to a given room layout.
This is particularly useful for panorama novel view synthesis~\cite{xu2021layout} and can potentially benefit indoor 3D scene generation~\cite{fang2023ctrl,hollein2023text2room}.
However, this problem is not well researched for diffusion-based panorama generation, largely due to the difficulty of introducing layout constraints while exploiting the rich prior knowledge of SD in perspective format at the same time.
For panorama generation from multi-view~\cite{tang2023mvdiffusion,li2023panogen}, one naive solution is to project layout condition into different views to locally condition the generation of perspective images.
Instead, for our dual-branch diffusion model, we can naturally exploit the global nature of the panorama branch to enforce a much stronger layout constraint.
Specifically, we render the layout condition as a distance map, then use it as the input of a ControlNet~\cite{zhang2023adding} to condition the panorama branch.

\subsection{Training}
During training, we employ the same technique for latent map initialization to jointly sample noise for the panorama and perspective view, which we denote as ${\epsilon}^{*}$ and ${\epsilon}^{i}$, respectively.
Given a GT panorama ${x}^{*}$, we apply supervision on the panorama branch with the loss in \cref{eq:ldm_loss}:
\begin{equation}
    \mathcal{L}^{*} = \mathbb{E}_{{\mathcal{E}}({x}^{*}),{t},{\epsilon}^{*},{y}} \left[ || {\epsilon}^{*} - {\epsilon}_{\theta}^{*}({z}_{t}^{*},{t},\tau({y})) ||_2 \right],
\end{equation}
where ${\epsilon}_{\theta}^{*}$ is the predicted noise from the panorama branch.
To encourage the synchronization between the two branches, we also apply supervision on the perspective branch with the following loss:
\begin{equation}
    \mathcal{L}^{i} = \mathbb{E}_{{\mathcal{E}}({x}^{i}),{t},{\epsilon}^{i},{y}} \left[ || {\epsilon}^{i} - {\epsilon}_{\theta}^{i}({z}_{t}^{i},{t},\tau({y})) ||_2 \right],
\end{equation}
where ${x}^{i}$ is a perspective image projected from ${x}^{*}$ and ${\epsilon}_{\theta}^{i}$ is the predicted noise from the perspective branch.
We joint train the EPPA modules and the LoRA layers in the two branches by combining the above two losses as ${\mathcal{L}} = \mathcal{L}^{*} + \frac{1}{N} \sum_{i=1}^{N} \mathcal{L}^{i}$.   Note that the SD UNet blocks remain frozen.

\section{Experiment}
\label{sec:exp}

\begin{figure*}[t]
    \vspace{-1em}
	\centering
	\small
    \includegraphics[width=1.\linewidth,clip,trim=0 0 0 0]{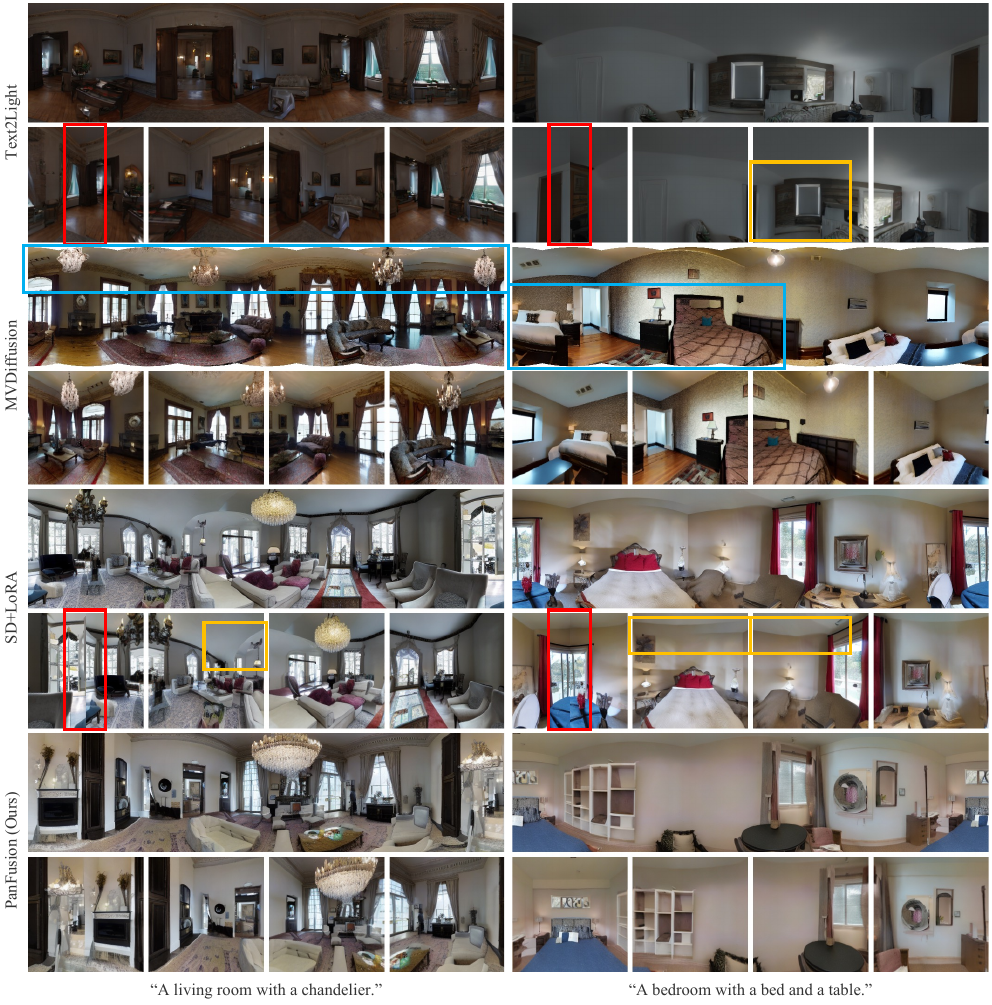}
    \vspace{-2em}
    \caption{
        Qualitative comparisons of text-conditioned panorama generation.
        We show panoramas cropped to the vertical FoV of MVDiffusion~\cite{tang2023mvdiffusion}.
        Below each panorama, we show 4 evenly spaced perspective projections, with the first view crossing the left and right boundaries.
        We highlight the \textcolor{red}{loop inconsistency}, \textcolor{orange}{distorted lines} and \textcolor{cyan}{repetitive objects and unreasonable furniture layout} of baseline methods with corresponding colors of boxes, which are addressed by our method.
        More results can be found in \cref{sec:suppl_comp} of the supplementary.
    }\label{fig:comp_lowres}
    \vspace{-1em}
\end{figure*}

\begin{table*}[t]
    \centering
    \vspace{-1em}
    \begin{tabular}{lccccccccc}
        \multirow{2}{*}{Method}
        & \multicolumn{4}{c}{Panorama}
        & \multicolumn{2}{c}{20 Views}
        & \multicolumn{3}{c}{Horizontal 8 Views~\cite{tang2023mvdiffusion}}
        \\
        \cmidrule(r){2-5} \cmidrule(r){6-7} \cmidrule(r){8-10}
        & FAED $\downarrow$
        & FID $\downarrow$ & IS $\uparrow$ & CS $\uparrow$
        & FID $\downarrow$ & IS $\uparrow$
        & FID $\downarrow$ & IS $\uparrow$ & CS $\uparrow$
        \\
        \midrule
        Text2Light~\cite{chen2022text2light}
        & 97.24
        & 76.5 & 3.60 & 27.48
        & 36.25 & 5.67
        & 43.66 & 4.92 & \underline{25.88}
        \\
        MVDiffusion~\cite{tang2023mvdiffusion}
        & -
        & - & - & -
        & - & -
        & 25.27 & \textbf{6.90} & \textbf{26.34}
        \\
        SD+LoRA~\cite{rombach2022highresolution,hu2021lora}
        & \underline{7.19}
        & 51.69 & \underline{4.40} & \textbf{28.83}
        & \underline{19.32} & \underline{6.90}
        & 20.68 & 6.48 & 24.77
        \\
        Pano Branch
        & 7.90
        & \underline{50.40} & \textbf{4.54} & \underline{28.67}
        & 20.10 & \textbf{7.06}
        & \underline{20.56} & 6.37 & 24.85
        \\
        {\name} (Ours)
        & \textbf{6.04}
        & \textbf{46.47} & 4.36 & 28.58
        & \textbf{17.04} & 6.85
        & \textbf{19.88} & \underline{6.50} & 24.98
    \end{tabular}
    \vspace{-0.5em}
    \caption{
        Comparison with SoTA methods.
        We evaluate the panorama image quality with Fr\'echet Auto-Encoder Distance (FAED), Fr\'echet Inception Distance (FID), Inception Score (IS), and CLIP Score (CS).
        In addition, we evaluate the perspective image quality in two settings.
        We first randomly sample 20 views, which is the closest to the real-world scenario where the user can freely navigate the panorama to view the scene from different perspectives.
        We then follow MVDiffusion~\cite{tang2023mvdiffusion} to horizontally sample 8 evenly spaced views.
    }\label{tbl:comp_lowres}
    \vspace{-0.5em}
\end{table*}
\begin{figure*}[t]
    \centering
    \includegraphics[width=1.\linewidth]{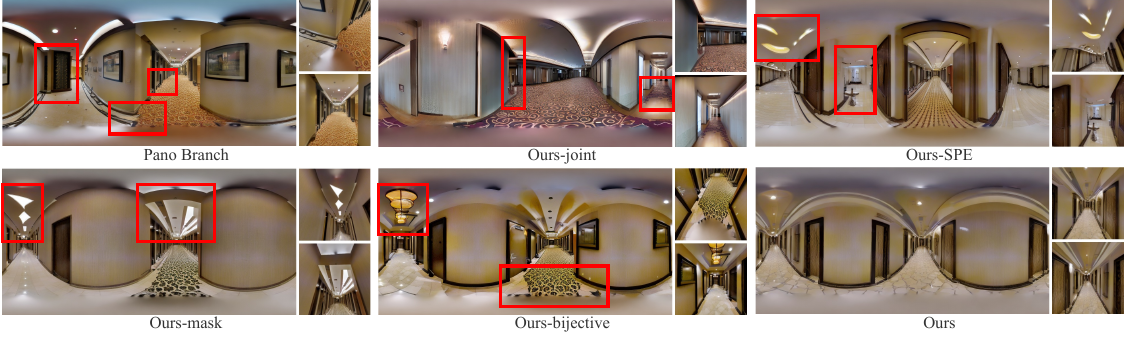}
    \vspace{-2em}
    \caption{
        Ablation study.
        Artifacts are highlighted with red boxes and projected to perspective views.
        Prompt: ``A hallway in a hotel.''
    }\label{fig:ablation}
    \vspace{-1em}
\end{figure*}

\subsection{Experimental Setup}\label{sec:exp_setup}

\noindent\textbf{Dataset.}
We follow the MVDiffusion~\cite{tang2023mvdiffusion} to use the Matterport3D dataset~\cite{chang2017matterport3d}, which has 10,800 panoramic images with 2,295 room layout annotations.
We employ BLIP-2~\cite{li2023blip2} to generate a short description for each image.

\noindent\textbf{Implementation Details.}
For text-conditioned generation, the training and inference schedules are kept the same as MVDiffusion~\cite{tang2023mvdiffusion} to make a fair comparison.
For text-layout conditioned generation, we train the additional ControlNet with other parameters fixed.

\noindent\textbf{Evaluation Metrics.}
Following previous works, we evaluate image quality in panorama~\cite{chen2022text2light} and perspective~\cite{tang2023mvdiffusion} domain.
For layout-conditioned generation, we propose a new metric to evaluate how well the generated panorama follows input layout.
Specifically, we use the following metrics:
\begin{itemize}[leftmargin=0pt, wide=0pt]
    \item \textit{Panorama.}
    We follow Text2Light~\cite{chen2022text2light} to report Fr\'echet Inception Distance (FID)~\cite{heusel2017gans} and Inception Score (IS)~\cite{salimans2016improved} on panoramas to measure realism and diversity.
    Additionally, CLIP Score (CS)~\cite{radford2021learning} is used to evaluate the text-image consistency.
    While FID is widely used for image generation, it relies on an Inception network~\cite{szegedy2016rethinking} trained on perspective images, thus less applicable for panorama images.
    Therefore, a variant of FID customized for panorama, Fr\'echet Auto-Encoder Distance (FAED)~\cite{oh2022bips}, is used to better compare the realism.
    \item \textit{Perspective.}
    To simulate the real-world scenario where the user can freely navigate a panorama by viewing from different perspective views,
    we also report FID and IS for 20 randomly sampled views 
    to compare with methods that generate 180$^{\circ}$ vertical FOV.
    We also follow MVDiffusion~\cite{tang2023mvdiffusion} to report FID, IS and CS scores on 8 horizontally sampled views. 
    It is worth noting that this group of metrics favors MVDiffusion by measuring its direct outputs, while ours involve interpolation for perspective views.
    \item \textit{Layout Consistency.}
    We propose a layout consistency metric, which employs a layout estimation network HorizonNet~\cite{sun2019horizonnet} to estimate the room layout from the generated panorama and then compute its 2D IoU and 3D IoU~\cite{sun2019horizonnet} with the input layout condition.
\end{itemize}
See Supplementary~\cref{sec:suppl_exp} for more details of the above.

\begin{table}[t]
    \centering
    \resizebox{1.\columnwidth}{!}{
    \begin{tabular}{lcccccc}
        \multirow{2}{*}{Method}
        & \multicolumn{4}{c}{Panorama}
        & \multicolumn{2}{c}{20 Views}
        \\
        \cmidrule(r){2-5} \cmidrule(r){6-7}
        & FAED $\downarrow$
        & FID $\downarrow$ & IS $\uparrow$ & CS $\uparrow$
        & FID $\downarrow$ & IS $\uparrow$
        \\ 
        \midrule
        Pano Branch
        & 7.90
        & 50.40 & 4.54 & \textbf{28.66}
        & 20.10 & \underline{7.06}
        \\
        Ours-joint
        & 7.75
        & 59.43 & 4.34 & 27.94
        & 29.68 & 6.82
        \\
        Ours-SPE
        & \underline{6.63}
        & 49.55 & \textbf{4.65} & \textbf{28.66}
        & 20.82 & 7.00
        \\
        Ours-mask
        & 6.77
        & \textbf{45.49} & 4.35 & \textbf{28.66}
        & \textbf{15.81} & 6.81
        \\
        Ours-bijective
        & 7.36
        & 48.35 & \underline{4.58} & \textbf{28.66}
        & 18.78 & \textbf{7.12}
        \\
        Ours
        & \textbf{6.04}
        & \underline{46.47} & 4.36 & \underline{28.58}
        & \underline{17.04} & 6.85
        \\
    \end{tabular}
    }
    \vspace{-1em}
    \caption{
        Ablation study.
        We compare the ablated versions of our method on both panorama and perspective domains. Here, ``-'' indicates removing the subsequent component.
    }\label{tbl:ablation}
    \vspace{-1em}
\end{table}

\subsection{Comparisons with Previous Methods}\label{sec:comp}
\noindent\textbf{Baselines.}
We compare our PanFusion with the following baselines (see Supplementary~\cref{sec:suppl_exp} for details).
\begin{itemize}[leftmargin=0pt, wide=0pt]
    \item \textit{MVDiffusion}~\cite{tang2023mvdiffusion} utilizes a multi-view diffusion model to generate 8 horizontal views that can be stitched into a panorama with vertical FOV of 90$^{\circ}$.
    It requires separate prompts for training while providing an option to generate from a single prompt.
    \item \textit{Text2Light}~\cite{chen2022text2light} generates a 180$^{\circ}$ vertical FOV panorama from a text prompt in a two-stage auto-regressive manner.
    \item \textit{SD+LoRA} is our baseline model that uses LoRA~\cite{hu2021lora} to finetune a Stable Diffusion~\cite{rombach2022highresolution} on panorama images.
    \item \textit{Pano Branch} is SD+LoRA with additional modifications as described in \cref{sec:dual_branch} to ensure loop consistency.
\end{itemize}

\noindent\textbf{Quantitative Results.} \cref{tbl:comp_lowres} shows the quantitative comparison results.
Here, we assign the highest value to realism in image generation, measured in FAED and FID.
On these two metrics, our method outperforms baseline methods in both panorama and perspective.
For IS, our method's performance is slightly lower than baselines.
This is likely due to the fact that IS evaluates diversity of objects in generated images, using a classifier, and our model, unlike the baselines, tends to not generate unexpected objects.
Similarly, it is possible to say that baseline models present slightly higher CS is due to the repetition of objects reinforcing alignment with prompts.
Considering SD+LoRA is superior to Pano Branch on FAED and is on par in other metrics, we only qualitatively compare with SD+LoRA below.

\noindent\textbf{Qualitative Results.} \cref{fig:comp_lowres} shows the qualitative comparison results. \textcolor{red}{Loop inconsistency} can be observed for Text2Light and SD+LoRA due to lack of message passing between left-right boundaries.
They also suffer from \textcolor{orange}{distorted lines} in perspective views, meaning the generated panorama does not follow the correct equirectangular projection.
MVDiffusion on the other hand suffers from \textcolor{cyan}{repetitive objects and unreasonable furniture layout},
which is likely due to the lack of global context.
Our method generates the most realistic scenes and aligns to text condition the best, also with less distortion in perspective views.

\begin{figure*}[t]
    \vspace{-1em}
    \centering
    \includegraphics[width=\textwidth]{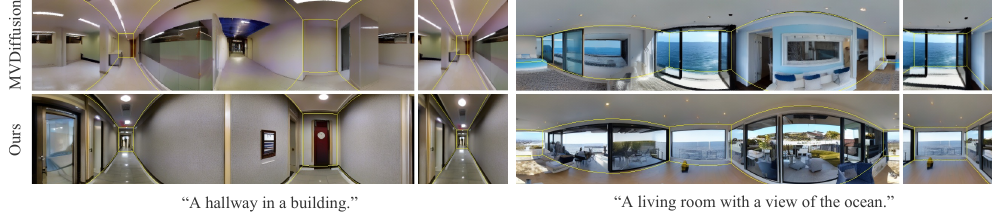}
    \vspace{-2em}
    \caption{
        Layout-conditioned generation comparisons. We showcase how layout-conditioned panorama generation can benefit from our dual-branch structure.
        The input layout is drawn on the genereated panorama as yellow lines, while the panorama is cropped to the FoV of MVDiffusion and projected to one view to highlight the layout consistency.
    }\label{fig:layout_cond}
    \vspace{-1.5em}
\end{figure*}

\subsection{Ablation Study}
\label{sec:ablation}
In~\cref{sec:comp} and~\cref{tbl:comp_lowres}, we show that our full model outperforms Pano Branch, the baseline model of our method without the perspective branch.
Here, as shown in \cref{tbl:ablation} 
and \cref{fig:ablation},
we further conduct an ablation study to validate the effectiveness of each component in our method.
For a consistent comparison, we keep the layout similar between different ablated versions by sampling the same noise for latent map initialization, exploiting the observation of~\cite{mao2023guided}.

\noindent\textbf{Joint latent map initialization.}
We ablate the joint latent map initialization by initializing the latent maps of the panorama and perspective branches separately (\textbf{Ours-joint}). 
Significant performance drop can be observed in all metrics and qualitative results, demonstrating the importance of joint latent map initialization.
Interestingly, Ours-joint is even worse than Pano Branch in FID. It is likely due to that joint latent map initialization helps the corresponding pixels to share a similar noise distribution from the beginning of the diffusion process, which is essential for EPPA to align the content of overlapped regions.

\noindent\textbf{EPP SPE and attention mask.}
We ablate spherical positional encoding (\textbf{Ours-SPE}) and attention mask (\textbf{Ours-mask}) from EPPA module.
From \cref{tbl:ablation}, we can see that missing SPE hurts FAED and FID, which is likely due to that SPE helps the model to learn the relative position of the pixels between the two branches.
Missing attention mask gets FID better, but it hurts FAED, which is more accurate in evaluating the panorama quality as it is customized for the target dataset.
Both result in clear artifacts around point light sources, inconsistent textures of the floor, and distortions in the highlighted projections, as shown in \cref{fig:ablation}.

\noindent\textbf{Bijective EPPA.}
We ablate the bijective EPPA (\textbf{Ours-bijective}) by using separate parameters for the two directions in the EPPA module.
Both FAED and FID get worse for Ours-bijective.
In addition, the ablated version struggles to generate consistent textures for the floor and ceiling for the two directions of the hallway in \cref{fig:ablation}.
On the contrary, our full model generates floor and ceiling with consistent style, showing a better global understanding of the scene.


\begin{table}[t]
    \centering
    \resizebox{1.\columnwidth}{!}{
    \begin{tabular}{lccccc}
        \multirow{2}{*}{Method}
        & \multicolumn{2}{c}{Layout Consistency}
        & \multicolumn{3}{c}{Horizontal 8 Views~\cite{tang2023mvdiffusion}}
        \\
        \cmidrule(r){2-3} \cmidrule(r){4-6}
        & 3D IoU $\uparrow$ & 2D IoU $\uparrow$
        & FID $\downarrow$ & IS $\uparrow$ & CS $\uparrow$
        \\
        \midrule
        MVDiffusion~\cite{tang2023mvdiffusion} 
        & 61.06 & 64.43
        & 28.83 & \textbf{5.60} & \textbf{26.79}
        \\
        {\name} (Ours) 
        & \textbf{68.46} & \textbf{71.82}
        & \textbf{22.58} & 5.10 & 26.04
        \\
        \midrule
        GT image
        & 74.31 & 77.15
        & - & - & -
    \end{tabular}
    }
    \vspace{-1em}
    \caption{
        Layout-conditioned comparisons.
        We evaluate layout consistency between the condition and the layout of each generated image extracted by a layout estimator HorizonNet~\cite{sun2019horizonnet}.
        ``GT image'' indicates the upper bound of the layout estimator.
    }\label{tbl:layout_cond}
    \vspace{-1em}
\end{table}

\subsection{Application: Layout-Conditioned Generation}
To showcase the advantage of our method in generating panorama images with additional layout conditions, we build a baseline model by adding a ControlNet to MVDiffusion as described in \cref{sec:layout_cond}.
We render the layout condition into a distance map and then project it to perspective views to condition its generation of multi-view images.
The training settings are kept the same as our PanFusion. 
As shown in \cref{tbl:layout_cond}, our method outperforms the baseline model in layout consistency while keeping the realism advantage of perspective projections.
We visualize the layout condition as wireframes overlaid on generated panorama images in \cref{fig:layout_cond}, where we can see
our generated panorama images follow their layout conditions better, especially as highlighted in perspective views.
More details can be found in \cref{sec:suppl_layout_cond} of the supplementary material.

\section{Conclusion}
\label{sec:conclusion}

\begin{figure}[t]
    \centering
	\small
    \setlength\tabcolsep{0px}
    \newcommand{\pano}[1]{\includegraphics[width=.5\linewidth,clip,trim=-4 0 -4 0]{#1}}
    \begin{tabular}{cc}
        \pano{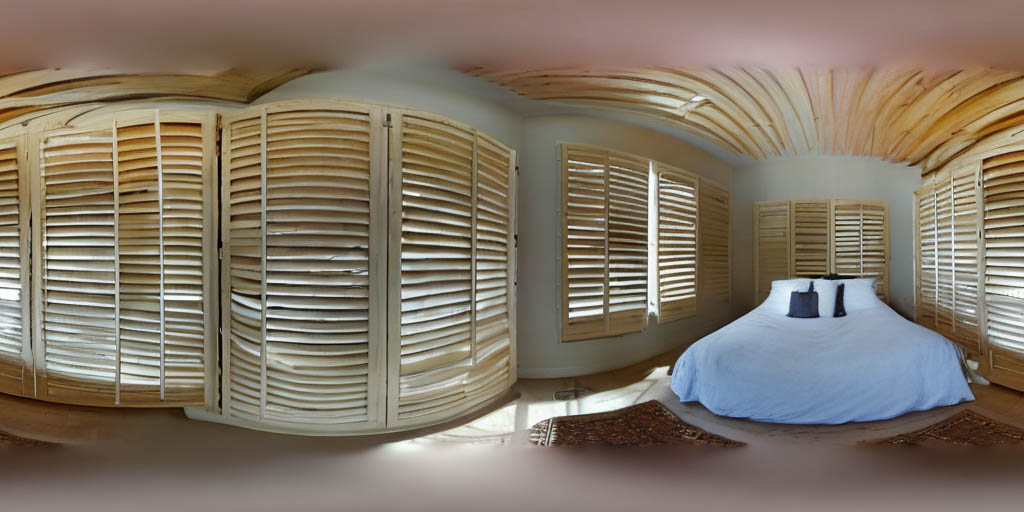}
        & \pano{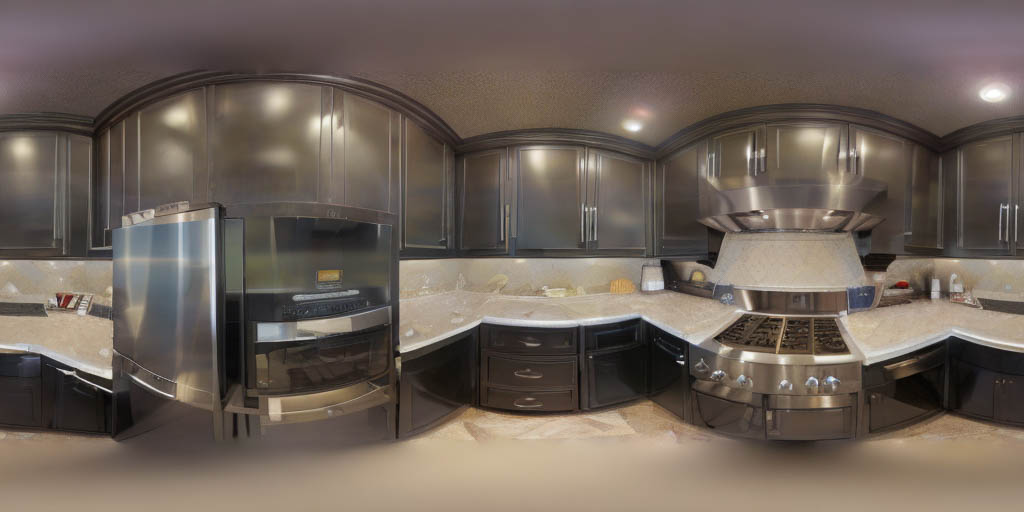}
    \end{tabular}
    \vspace{-1em}
    \caption{
        Failure case.
        Our method sometimes generates indoor scenes with no entrance.
    }\label{fig:failure}
    \vspace{-1em}
\end{figure}

In this paper, we have proposed {\name}, a novel text-to-360$^{\circ}$-panorama image generation method that can generate high-quality panorama images from a single text prompt. Particularly, a dual-branch diffusion architecture has been introduced to harness prior knowledge of Stable Diffusion in the perspective domain while addressing the repetitive elements and inconsistency issues observed in previous works.
An EPPA module has been further introduced to enhance the information passing between the two branches.
We have also extended our PanFusion 
for the application of layout-conditioned panorama generation.
Comprehensive experiments have demonstrated that {\name} can generate high-quality panorama images with better realism and layout consistency than previous methods.

\noindent\textbf{Limitations.}
Although the dual-branch architecture of {\name} combines the advantages in both panorama and perspective domains, it comes with a cost of higher computational complexity.
Additionally, our method sometimes fails to generate entrance for indoor scenes, as shown in \cref{fig:failure}, which is essential for use cases like virtual tour.

\noindent\textbf{Acknowledgement:}
This research is supported by Building 4.0 CRC and the National Key R\&D Program of China (NO.2022ZD0160101), and was partially done at Shanghai AI Laboratory.

{
    \small
    \bibliographystyle{ieeenat_fullname}
    \bibliography{main}
}

\clearpage
\setcounter{page}{1}
\maketitlesupplementary

\counterwithin{figure}{section}
\counterwithin{table}{section}
\renewcommand\thesection{\Alph{section}}
\renewcommand\thetable{\thesection.\arabic{table}}
\renewcommand\thefigure{\thesection.\arabic{figure}}
\setcounter{section}{0}


The supplementary material is organized as follows.
In \cref{sec:suppl_arch}, we provide more details about the network architecture.
In \cref{sec:suppl_exp}, we provide more details about the experiment setup and baseline methods.
In \cref{sec:suppl_comp}, we provide more qualitative comparisons.
In \cref{sec:suppl_layout_cond}, we provide more details about the layout conditioned generation.
In \cref{sec:suppl_generalization}, we provide more generalization results to out-domain prompts.

\section{Network Architecture}\label{sec:suppl_arch}
{
\newcounter{layer_idx}
\setcounter{layer_idx}{0}
\newcommand{\rowindex}{(\stepcounter{layer_idx}\arabic{layer_idx})}

\begin{table}[t]
    \centering
    \resizebox{.92\columnwidth}{!}{
    \begin{tabular}{ccccc}
        \multirow{2}{*}{\#} & \multirow{2}{*}{Layer} & \multicolumn{2}{c}{Output} & \multirow{2}{*}{\makecell{Additional \\ Inputs}} \\
        \cmidrule(r){3-4}
        & & Pers Branch ($ 20 \times $) & Pano Branch & \\
        \midrule
        \rowindex & Latent Map & $4 \times 32 \times 32$ & $4 \times 64 \times 128$ & \\
        \rowindex & Conv. & $320 \times 32 \times 32$ & $320 \times 64 \times 128$ & \\ 
        \midrule
        \multicolumn{5}{c}{CrossAttnDownBlock1} \\
        \midrule
        \rowindex & ResBlock & $320 \times 32 \times 32$ & $320 \times 64 \times 128$ & Time emb. \\
        \rowindex & AttnBlock & $320 \times 32 \times 32$ & $320 \times 64 \times 128$ & Prompt emb. \\ 
        \rowindex & ResBlock & $320 \times 32 \times 32$ & $320 \times 64 \times 128$ & Time emb. \\
        \rowindex & AttnBlock & $320 \times 32 \times 32$ & $320 \times 64 \times 128$ & Prompt emb. \\ 
        \rowindex & DownSampler & $320 \times 16 \times 16$ & $320 \times 32 \times 64$ & \\ 
        \rowcolor{orange!50}
        \rowindex & EPPA & $320 \times 16 \times 16$ & $320 \times 32 \times 64$ & \\
        \midrule
        \multicolumn{5}{c}{CrossAttnDownBlock2} \\
        \midrule
        \rowindex & ResBlock & $640 \times 16 \times 16$ & $640 \times 32 \times 64$ & Time emb. \\
        \rowindex & AttnBlock & $640 \times 16 \times 16$ & $640 \times 32 \times 64$ & Prompt emb. \\ 
        \rowindex & ResBlock & $640 \times 16 \times 16$ & $640 \times 32 \times 64$ & Time emb. \\
        \rowindex & AttnBlock & $640 \times 16 \times 16$ & $640 \times 32 \times 64$ & Prompt emb. \\ 
        \rowindex & DownSampler & $640 \times 8 \times 8$ & $640 \times 16 \times 32$ & \\ 
        \rowcolor{orange!50}
        \rowindex & EPPA & $640 \times 8 \times 8$ & $640 \times 16 \times 32$ & \\
        \midrule
        \multicolumn{5}{c}{CrossAttnDownBlock3} \\
        \midrule
        \rowindex & ResBlock & $1280 \times 8 \times 8$ & $1280 \times 16 \times 32$ & Time emb. \\
        \rowindex & AttnBlock & $1280 \times 8 \times 8$ & $1280 \times 16 \times 32$ & Prompt emb. \\ 
        \rowindex & ResBlock & $1280 \times 8 \times 8$ & $1280 \times 16 \times 32$ & Time emb. \\
        \rowindex & AttnBlock & $1280 \times 8 \times 8$ & $1280 \times 16 \times 32$ & Prompt emb. \\ 
        \rowindex & DownSampler & $1280 \times 4 \times 4$ & $1280 \times 8 \times 16$ & \\ 
        \rowcolor{orange!50}
        \rowindex & EPPA & $1280 \times 4 \times 4$ & $1280 \times 8 \times 16$ & \\
        \midrule
        \multicolumn{5}{c}{DownBlock} \\
        \midrule
        \rowindex & ResBlock & $1280 \times 4 \times 4$ & $1280 \times 8 \times 16$ & Time emb. \\ 
        \rowindex & ResBlock & $1280 \times 4 \times 4$ & $1280 \times 8 \times 16$ & Time emb. \\ 
        \midrule
        \multicolumn{5}{c}{MidBlock} \\
        \midrule
        \rowindex & ResBlock & $1280 \times 4 \times 4$ & $1280 \times 8 \times 16$ & Time emb. \\
        \rowindex & AttnBlock & $1280 \times 4 \times 4$ & $1280 \times 8 \times 16$ & Prompt emb. \\
        \rowindex & ResBlock & $1280 \times 4 \times 4$ & $1280 \times 8 \times 16$ & Time emb. \\
        \rowcolor{orange!50}
        \rowindex & EPPA & $1280 \times 4 \times 4$ & $1280 \times 8 \times 16$ & \\
        \midrule
        \multicolumn{5}{c}{UpBlock} \\
        \midrule
        \rowindex & ResBlock & $1280 \times 4 \times 4$ & $1280 \times 8 \times 16$ & (22), Time emb. \\
        \rowindex & ResBlock & $1280 \times 4 \times 4$ & $1280 \times 8 \times 16$ & (21), Time emb. \\
        \rowindex & ResBlock & $1280 \times 4 \times 4$ & $1280 \times 8 \times 16$ & (19), Time emb. \\
        \rowcolor{orange!50}
        \rowindex & EPPA & $1280 \times 4 \times 4$ & $1280 \times 8 \times 16$ & \\
        \rowindex & UpSampler & $1280 \times 8 \times 8$ & $1280 \times 16 \times 32$ & \\
        \midrule
        \multicolumn{5}{c}{CrossAttnUpBlock1} \\
        \midrule
        \rowindex & ResBlock & $1280 \times 8 \times 8$ & $1280 \times 16 \times 32$ & (18), Time emb. \\
        \rowindex & AttnBlock & $1280 \times 8 \times 8$ & $1280 \times 16 \times 32$ & Prompt emb. \\
        \rowindex & ResBlock & $1280 \times 8 \times 8$ & $1280 \times 16 \times 32$ & (16), Time emb. \\
        \rowindex & AttnBlock & $1280 \times 8 \times 8$ & $1280 \times 16 \times 32$ & Prompt emb. \\
        \rowindex & ResBlock & $1280 \times 8 \times 8$ & $1280 \times 16 \times 32$ & (13), Time emb. \\
        \rowindex & AttnBlock & $1280 \times 8 \times 8$ & $1280 \times 16 \times 32$ & Prompt emb. \\
        \rowcolor{orange!50}
        \rowindex & EPPA & $1280 \times 8 \times 8$ & $1280 \times 16 \times 32$ & \\
        \rowindex & UpSampler & $1280 \times 16 \times 16$ & $1280 \times 32 \times 64$ & \\
        \midrule
        \multicolumn{5}{c}{CrossAttnUpBlock2} \\
        \midrule
        \rowindex & ResBlock & $640 \times 16 \times 16$ & $640 \times 32 \times 64$ & (12), Time emb. \\
        \rowindex & AttnBlock & $640 \times 16 \times 16$ & $640 \times 32 \times 64$ & Prompt emb. \\
        \rowindex & ResBlock & $640 \times 16 \times 16$ & $640 \times 32 \times 64$ & (10), Time emb. \\
        \rowindex & AttnBlock & $640 \times 16 \times 16$ & $640 \times 32 \times 64$ & Prompt emb. \\
        \rowindex & ResBlock & $640 \times 16 \times 16$ & $640 \times 32 \times 64$ & (7), Time emb. \\
        \rowindex & AttnBlock & $640 \times 16 \times 16$ & $640 \times 32 \times 64$ & Prompt emb. \\
        \rowcolor{orange!50}
        \rowindex & EPPA & $640 \times 16 \times 16$ & $640 \times 32 \times 64$ & \\
        \rowindex & UpSampler & $640 \times 32 \times 32$ & $640 \times 64 \times 128$ & \\
        \midrule
        \multicolumn{5}{c}{CrossAttnUpBlock3} \\
        \midrule
        \rowindex & ResBlock & $320 \times 32 \times 32$ & $320 \times 64 \times 128$ & (6), Time emb. \\
        \rowindex & AttnBlock & $320 \times 32 \times 32$ & $320 \times 64 \times 128$ & Prompt emb. \\
        \rowindex & ResBlock & $320 \times 32 \times 32$ & $320 \times 64 \times 128$ & (4), Time emb. \\
        \rowindex & AttnBlock & $320 \times 32 \times 32$ & $320 \times 64 \times 128$ & Prompt emb. \\
        \rowindex & ResBlock & $320 \times 32 \times 32$ & $320 \times 64 \times 128$ & (2), Time emb. \\
        \rowindex & AttnBlock & $320 \times 32 \times 32$ & $320 \times 64 \times 128$ & Prompt emb. \\
        \midrule
        \rowindex & GroupNorm & $320 \times 32 \times 32$ & $320 \times 64 \times 128$ & \\
        \rowindex & SiLU & $320 \times 32 \times 32$ & $320 \times 64 \times 128$ & \\
        \rowindex & Conv. & $4 \times 32 \times 32$ & $4 \times 64 \times 128$ & \\
    \end{tabular}
    }
    \caption{
        Detailed PanFusion pipeline.
        We highlight the inserted EPPA modules in orange.
    }\label{tbl:pipeline}
\end{table}
}

In \cref{sec:method} of the main paper, we introduced our dual-branch architecture for panorama generation.
Here we provide more details about the position of inserting EPPA into the UNet of SD and the feature dimension of each layer in \cref{tbl:pipeline}.
We found that inserting EPPA earlier than the first DownSampler (Table~A.1(8)) and later than the last UpSampler (Table~A.1(47)) is memory-consuming due to large feature maps with no better performance.
Therefore, we insert EPPA right after the DownSampler and before the UpSampler of each block.

\section{Experiment Details}\label{sec:suppl_exp}

As mentioned in \cref{sec:exp} of the main paper, here we provide more details about the experiment setup and baseline methods.

\noindent\textbf{Dataset.}
\begin{figure*}[t]
    \centering
    \begin{subfigure}[t]{0.49\linewidth}
        \centering
        \includegraphics[width=1.0\linewidth, trim={0 0 0 0}, clip]{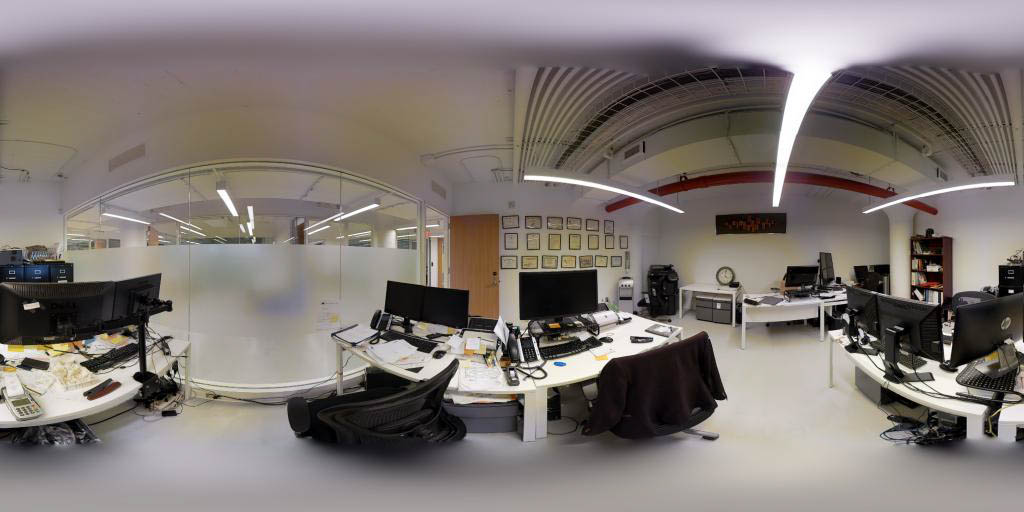}
        \caption{
            Panorama image.
        }\label{fig:mp3d_pano}
    \end{subfigure}
    \begin{subfigure}[t]{0.49\linewidth}
        \centering
        \includegraphics[width=1.0\linewidth, trim={0 0 0 0}, clip]{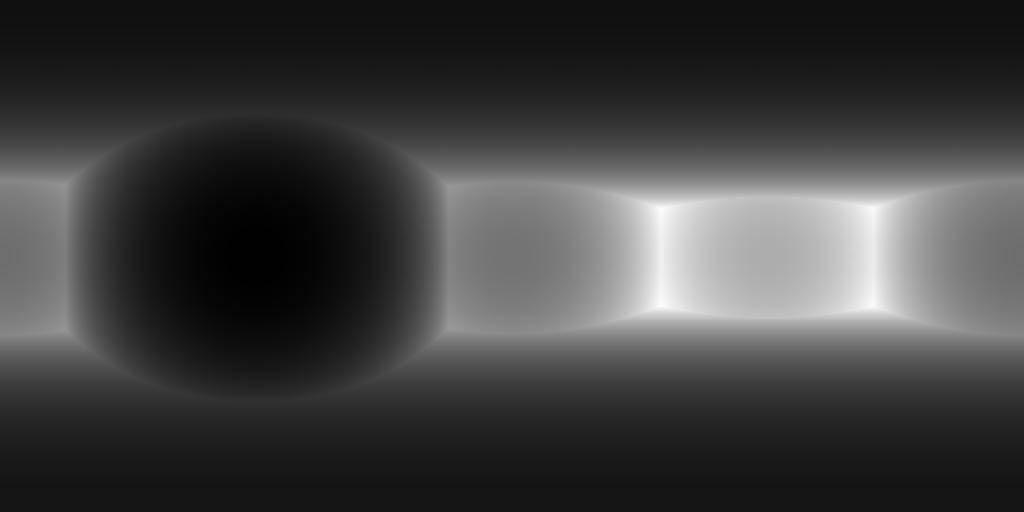}
        \caption{
            Layout distance map.
        }\label{fig:mp3d_layout}
    \end{subfigure}
    \vspace{-0.5em}
    \caption{
        An example of a panoramic image from the Matterport3D dataset~\cite{chang2017matterport3d} (a) and its room layout rendered as distance map (b).
        Regions near the upper and lower edges of the panoramic image are blurry in the original dataset.
    }\label{fig:mp3d}
\end{figure*}

Matterport3D dataset~\cite{chang2017matterport3d} is a large-scale scene understanding dataset with 10,800 panoramic images from 90 building-scale scenes.
For text conditioned generation, we utilize BLIP-2~\cite{li2023blip2} to generate a short description of the full image with a prompt of ``a 360 - degree view of''.
We use the same data split as~\cite{tang2023mvdiffusion}, which contains 9,820 for training and 1,092 for evaluation.
We note that the original Matterport3D dataset contains blurry regions near the upper and lower edges, as shown in \cref{fig:mp3d_pano}.
Therefore, our model is trained to generate images with similar blurry regions.
For text and layout conditioned generation, we use the MatterportLayout~\cite{wang2020layoutmp3d,zou2021manhattan} dataset, which annotates room layout for 2,295 indoor panoramic images in the Matterport3D dataset, with 1,648 for training, 191 for validation, and 459 for testing.

\noindent\textbf{Implementation Details.}
We implement our model in PyTorch based on the implementation of Stable Diffusion~\cite{rombach2022highresolution} from Diffusers~\cite{von-platen-etal-2022-diffusers}.
When training our dual-branch model for text-conditioned generation, we jointly train the EPPA module and finetune the two branches with rank-4 LoRA to the new resolution.
We randomly sample 20 views as the input of the perspective branch to encourage EPPA to understand the correspondence provided by SPE and attention mask instead of remembering the fixed camera poses.
Following MVDiffusion~\cite{tang2023mvdiffusion}, we train for 10 epochs with the AdamW optimizer, using a batch size of 4 and learning rate of 2e-4 for training, and a DDIM sampler~\cite{song2020denoising} is used with a step size of 50 for inference.
When training an additional ControlNet for text-layout conditioned generation, we extend training to 100 epochs due to less room layout annotations.
Training is conducted on 4 NVIDIA A100 GPUs and takes about 8 hours for text conditioned generation and 15 hours for text-layout conditioned generation.

\noindent\textbf{Perspective transformation details.}
In the main paper, we denote the transformation from equirectangular panorama ${I}^{*} \in \mathbb{R}^{C \times H \times W}$ to perspective image ${I} \in \mathbb{R}^{C \times h \times w}$ as ${I} = \text{P}\left( {I}^{*}, {R}, \text{FoV}, \left( {h}, {w} \right) \right)$,
where the rotation matrix ${R} \in SO(3)$ describes the camera extrinsic matrix
and $\text{FoV}$ and image size $\left( {h}, {w} \right)$ define the camera intrinsic matrix ${K}$.
Specifically, given a pixel ${p} \in \mathbb{R}^{2}$ on the image plane of ${I}$,
we shoot a ray ${K}^{-1} \left[ {p}, 1 \right]^{T}$ from the camera center,
and then transform it to the 3D coordinate of panorama as ${v} = {R}^{-1} {K}^{-1} \left[ {p}, 1 \right]^{T}$.
Subsequently, its corresponding pixel ${p}^{*}$ on the image plane of ${I}^{*}$ can be computed as:
\begin{equation}
    \resizebox{1.\linewidth}{!}{
        $
        \notag
        {p}^{*} = \left[ \frac{{W} \left( atan2 \left( {v}_{y}, {v}_{x} \right) + \pi \right) }{{2\pi}}, \frac{{H} \left( atan2 \left( {v}_{z}, \sqrt{{v}_{x}^{2} + {v}_{y}^{2}} \right) + \pi / 2 \right) }{\pi} \right],
        $
    }
\end{equation}
which is used to bilinearly interpolate ${I}$ from ${I}^{*}$.
Note that we use different symbols here for easy explanation.

\noindent\textbf{Evaluation Metrics.}
Previous works MVDiffusion~\cite{tang2023mvdiffusion} and Text2Light~\cite{chen2022text2light} both address the problem of text conditioned image generation, but in different domains.
MVDiffusion generates 8 horizontal views with 90$^{\circ}$ FoV, thus limiting the evaluation to perspective images.
Text2Light generates a full 180$^{\circ}$ vertical FoV, therefore focusing on evaluating the panorama quality.
Ours is closer to the latter, but to showcase the effectiveness of our proposed method, we conduct a comparison in both.
We also detail the implementation of layout consistency evaluation in the following.
\begin{itemize}[leftmargin=0pt, wide=0pt]
    \item \textit{In the panorama domain},
    we value Fr\'echet Auto-Encoder Distance (FAED)~\cite{oh2022bips} more, since it is customized for panorama and uses an auto-encoder trained on the target dataset as the feature extractor.
    Specifically, we train the auto-encoder similar to~\cite{oh2022bips} but with RGB images instead of RGBD by removing the depth branch.
    The auto-encoder is trained on the training set of the Matterport3D dataset for 60 epochs with Adam optimizer and batch size of 4.
    An exponential learning rate scheduler is used with an initial learning rate of 1e-4 and decay rate of 0.99 for every epoch.
    \item \textit{In the perspective domain},
    the CS is measured between the perspective image and the text prompt captioned from GT view using BLIP-2~\cite{li2023blip2}.
    \item \textit{When evaluating layout consistency},
    to make the comparison fair for MVDiffusion, we mask out pixels outside its vertical FoV before feeding the generated panorama to HorizonNet for our method, so that we do not benefit from the larger FoV when estimating the layout.
    We finetune HorizonNet on the masked training set of MatterportLayout dataset for 100 epochs with Adam optimizer and batch size of 4.
    The initial learning rate is set to 1e-4 and halved if the validation loss does not decrease for 10 epochs.
\end{itemize}

\noindent\textbf{Additional comparison with previous methods.}
\begin{table}[t]
    \centering
    \resizebox{1.\linewidth}{!}{
    \begin{tabular}{lccccccccc}
        \multirow{2}{*}{Method}
        & \multicolumn{3}{c}{Horizontal 8 Views~\cite{tang2023mvdiffusion}}
        \\
        \cmidrule(r){2-4}
        & FID $\downarrow$ & IS $\uparrow$ & CS $\uparrow$
        \\
        \midrule
        MVDiffusion~\cite{tang2023mvdiffusion}
        & 25.27 & 6.90 & 26.34
        \\
        MVDiffusion (projection)
        & 32.56 & 6.40 & 25.70
        \\
        MVDiffusion+LoRA
        & 21.76 & 6.55 & 25.22
        \\
        MVDiffusion+LoRA (projection)
        & 30.04 & 5.69 & 24.90
        \\
        {\name} (Ours)
        & 19.88 & 6.50 & 24.98
    \end{tabular}
    }
    \vspace{-0.5em}
    \caption{
        More quantitative comparison.
        We compare our method with MVDiffusion~\cite{tang2023mvdiffusion} in different settings.
        MVDiffusion with projection considers stitching and projection, which is closer to our setting.
        We also finetune MVDiffusion with LoRA~\cite{hu2021lora} on low resolution to have a fair comparison for time efficiency and layout-conditioned generation.
    }\label{tbl:more_comp}
\end{table}
We follow~\cite{tang2023mvdiffusion} to use the released weights of Text2Light~\cite{chen2022text2light} and MVDiffusion~\cite{tang2023mvdiffusion} as two of our baselines.
For Text2Light, we use its first stage without super-resolution inverse tone mapping stage to get panoramic images at a resolution of $512 \times 1024$, which takes 80.6 seconds per image on a single NVIDIA A100 GPU.
For MVDiffusion, we use its direct outputs for quantitative comparison in main paper \cref{tbl:comp_lowres}.
This favors MVDiffusion by avoiding inconsistency in stitching and interpolating in projection.
Therefore, to make a comprehensive comparison, we additionally evaluate MVDiffusion in different settings in \cref{tbl:more_comp}.
We detail these settings in the following.
\begin{itemize}[leftmargin=0pt, wide=0pt]
    \item \textit{MVDiffusion} is in its original setting that does not involve stitching and projection, and is used for comparison in main paper \cref{tbl:comp_lowres}.
    It outputs 8 horizontal views with 90$^{\circ}$ FoV at a resolution of $512 \times 512$, which takes 102.2 seconds.
    One only difference from the original MVDiffusion paper is that we downsample the output images to $256 \times 256$ before evaluation to match the resolution of GT images.
    \item \textit{MVDiffusion (projection)} uses the same weight as MVDiffusion, but we stitch its outputs into a panorama and then project the panorama back to perspective views for evaluation.
    This strictly follows our setting of panorama generation by considering the inconsistency between the perspective images.
    The performance drops significantly, which shows that inconsistency is a major issue for MVDiffusion.
    \item \textit{MVDiffusion+LoRA} is MVDiffusion finetuned with LoRA on a lower resolution at $256 \times 256$.
    With lower resolution, the inference time is reduced to 27.4 seconds for a panorama with 90$^{\circ}$ vertical FoV at the resolution of $256 \times 1024$, while our method takes 15.1 seconds to generate a panorama with 180$^{\circ}$ vertical FoV at the resolution of $512 \times 1024$.
    This setting skips the stitching and projection thus does not reflect the actual panorama generation ability of MVDiffusion.
    \item \textit{MVDiffusion+LoRA (projection)} follows the evaluation setting of MVDiffusion (projection) but uses the same weight as MVDiffusion+LoRA.
    The FID is better than MVDiffusion (projection), but still significantly worse than ours.
    This version is used for layout conditioned generation in \cref{tbl:layout_cond} of the main paper, detailed in \cref{sec:suppl_layout_cond}.
\end{itemize}

While our method achieves better realism than baseline methods, it comes with a cost of higher computational complexity as discussed in \cref{sec:conclusion} of the main paper.
Specifically, the average inference time is 2.8 and 2.9 seconds per panorama for SD+LoRA and Pano Branch, respectively.
However, we note that our model can be further optimized for higher speed as a significant amount of numpy operations are used for the EPPA module.

\section{Loop Consistency Analysis}

\begin{figure}[t]
    \centering
    \footnotesize
    \setlength\tabcolsep{0px}
    \renewcommand\arraystretch{0.5}
    \newcommand{\crop}[1]{\includegraphics[width=.16\linewidth,clip,trim=-2 0 -2 -4]{figure/loop_consistency/#1}}
    \begin{tabular}{cccccc}
        \crop{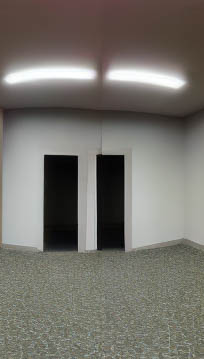}
        & \crop{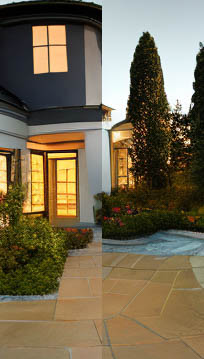}
        & \crop{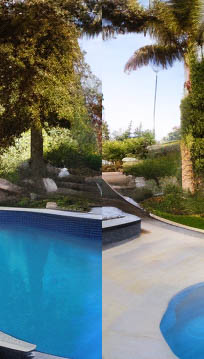}
        & \crop{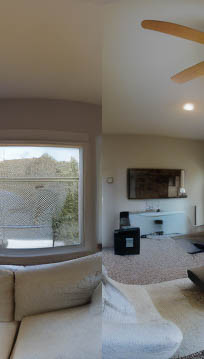}
        & \crop{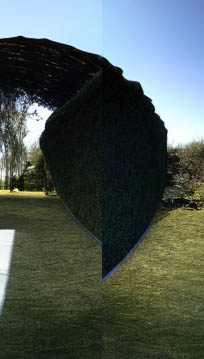}
        & \crop{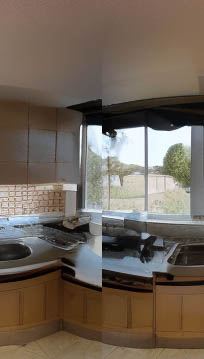}
        \\
        \multicolumn{6}{c}{(a) SD+LoRA}
        \\
        \crop{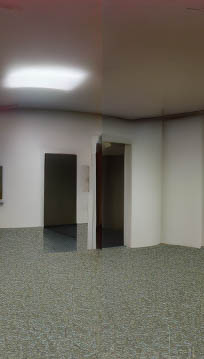}
        & \crop{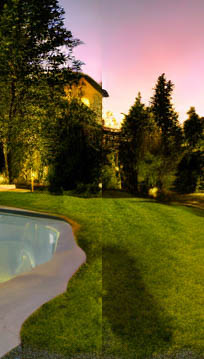}
        & \crop{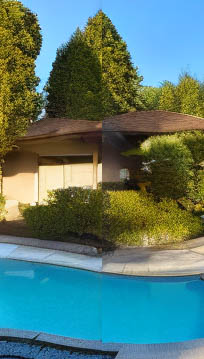}
        & \crop{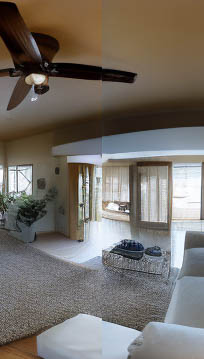}
        & \crop{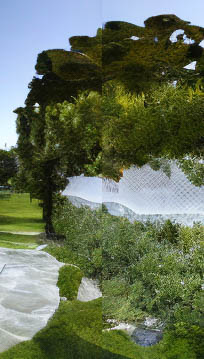}
        & \crop{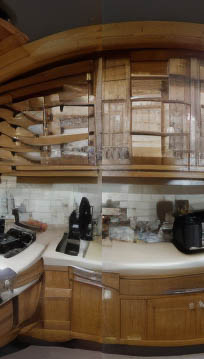}
        \\
        \multicolumn{6}{c}{(b) SD+LoRA + latent rotation}
        \\
        \crop{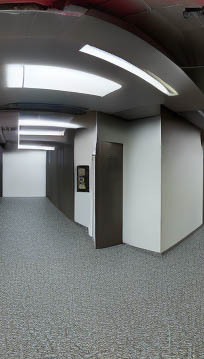}
        & \crop{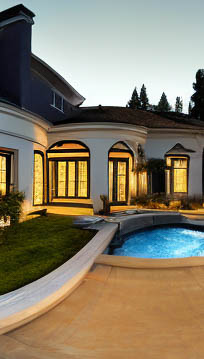}
        & \crop{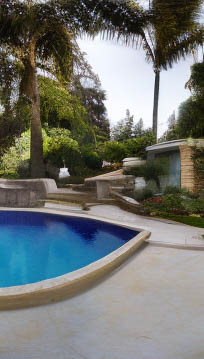}
        & \crop{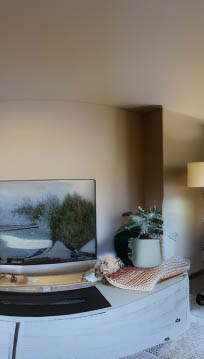}
        & \crop{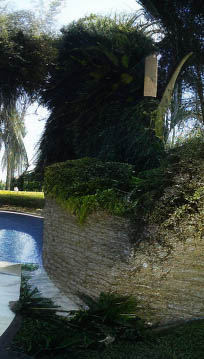}
        & \crop{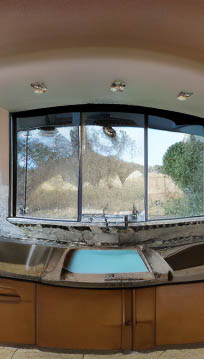}
        \\
        \multicolumn{6}{c}{(c) Pano Branch (SD+LoRA + latent rotation + circular padding)}
        \\
        \crop{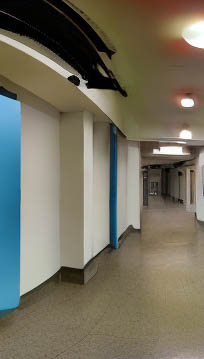}
        & \crop{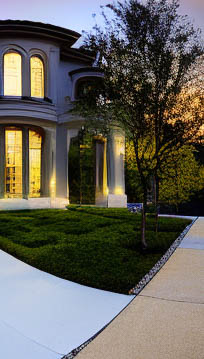}
        & \crop{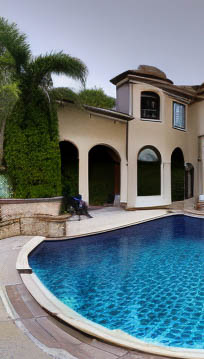}
        & \crop{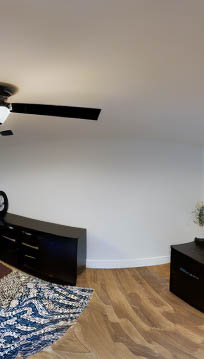}
        & \crop{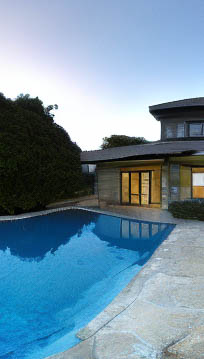}
        & \crop{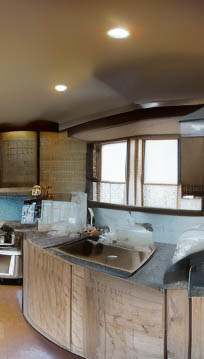}
        \\
        \multicolumn{6}{c}{(d) Pano Branch - latent rotation (SD+LoRA + circular padding)}
    \end{tabular}
    \caption{
        Loop consistency analysis.
        We stitch both ends of each generated panorama.
        Here, each column corresponding to one same input text.
        It is shown that latent rotation (b) can only mitigate loop inconsistency of SD+LoRA (a), while the results with circular padding combined (c) or alone (d) are more seamless.
    }\label{fig:loop_consistency}
\end{figure}

In \cref{sec:dual_branch}, we describe two techniques to eliminate loop inconsistency, \ie, latent rotation and circular padding.
Qualitative results in \cref{fig:loop_consistency} show the stitched ends of generated panoramas with each column corresponding to one input text.
We can see that latent rotation (b) can only mitigate loop inconsistency of SD+LoRA (a), while the results with circular padding combined (c) or alone (d) are more seamless.

\section{Repetition analysis.}
In \cref{sec:comp}, we qualitatively highlight the repetition issue of MVDiffusion.
Here, we try to evaluate the repetition by projecting panorama to cubemap and computing a score
$\text{RS} \left( {I}_{i}, {I}_{j} \right) = max \left( 100 * \cos \left( {E}_{i}, {E}_{j} \right), 0 \right)$ between each pair of 4 horizontal views,
where ${E}_{*}$ is the CLIP embedding of image ${I}_{*}$.
RS is averaged over all image pairs of 1,092 test samples, with higher values indicating more repetition.
It is shown in \cref{tbl:repetition_suppl} that our method has the lowest RS 
while MVDiffusion has the most repetition.

\begin{table}[t]
    \centering
    \resizebox{1.\columnwidth}{!}{
    \begin{tabular}{lcccccc}
        & Text2Light & MVDiffusion & PanFusion (Ours) & GT image
        \\
        \midrule
        RS $\downarrow$ & 88.81 & 90.79 & 88.13 & 86.49
    \end{tabular}
    }
    \caption{
        Repetition analysis. Inspired by CLIP Score~\cite{radford2021learning}, we report the repetition score (RS) that measures the similarity between different parts of the generated panorama images.
        Lower RS indicates less repetition.
    }\label{tbl:repetition_suppl}
\end{table}

\section{More Qualitative Comparisons}\label{sec:suppl_comp}
\newcommand{\compfig}[5]{
    \begin{figure*}[t]
        \vspace{-2em}
        \centering
        \small
        \setlength\tabcolsep{0px}
        \newcommand{\pano}[1]{
            \raisebox{-0.5\height}{\includegraphics[width=.47\linewidth,clip,trim=-4 -4 0 0]{##1}}
        }
        \begin{tabular}{rcc}
            \rotatebox[origin=c]{90}{
                \makecell[c]{
                    \newcolumntype{C}[1]{>{\hsize=##1\hsize\centering\arraybackslash}X}
                    \begin{tabularx}{1.25\linewidth}{C{1.1}C{1.1}C{0.7}C{1.1}}
                        PanFusion (Ours) & SD+LoRA & MVDiffusion & Text2Light
                    \end{tabularx}
                }
            }
            & \pano{figure/more_comp/#1}
            & \pano{figure/more_comp/#2}
            \\
            & ``#3.''
            & ``#4.''
        \end{tabular}
        \vspace{-0.5em}
        \caption{More qualitative comparisons.}\label{fig:more_comp_#5}
        \vspace{-0.5em}
    \end{figure*}
}

\compfig
{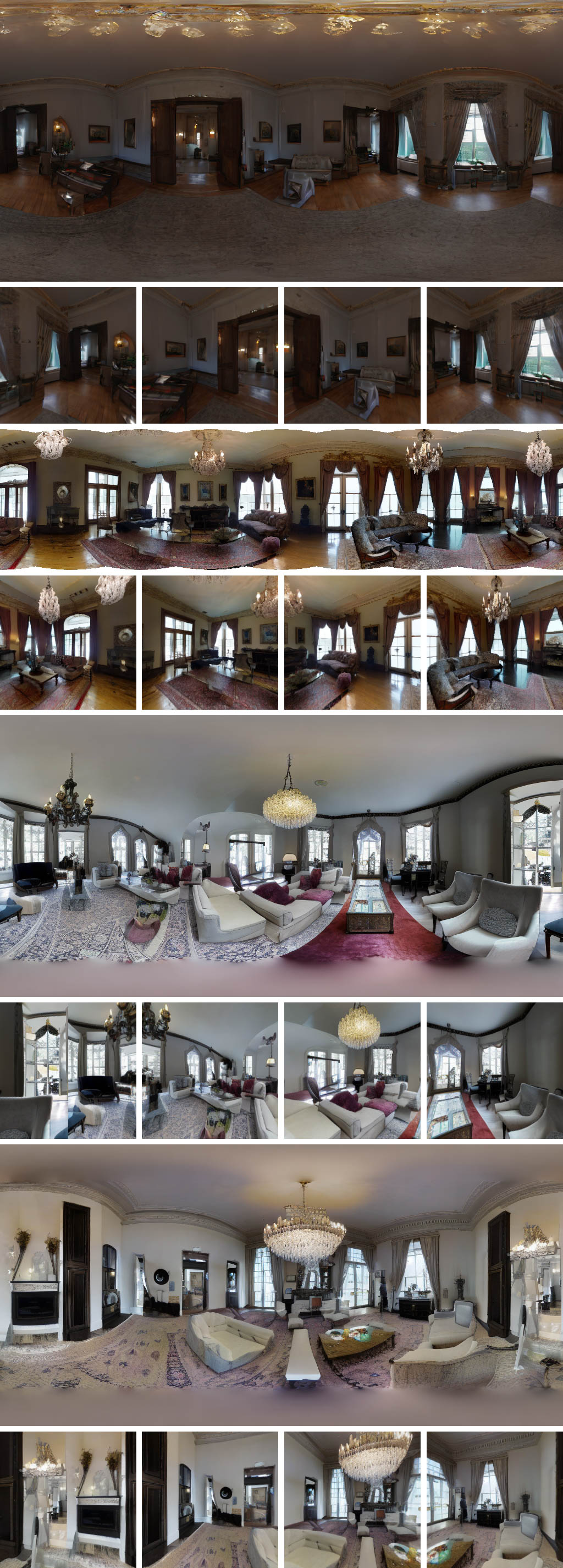}
{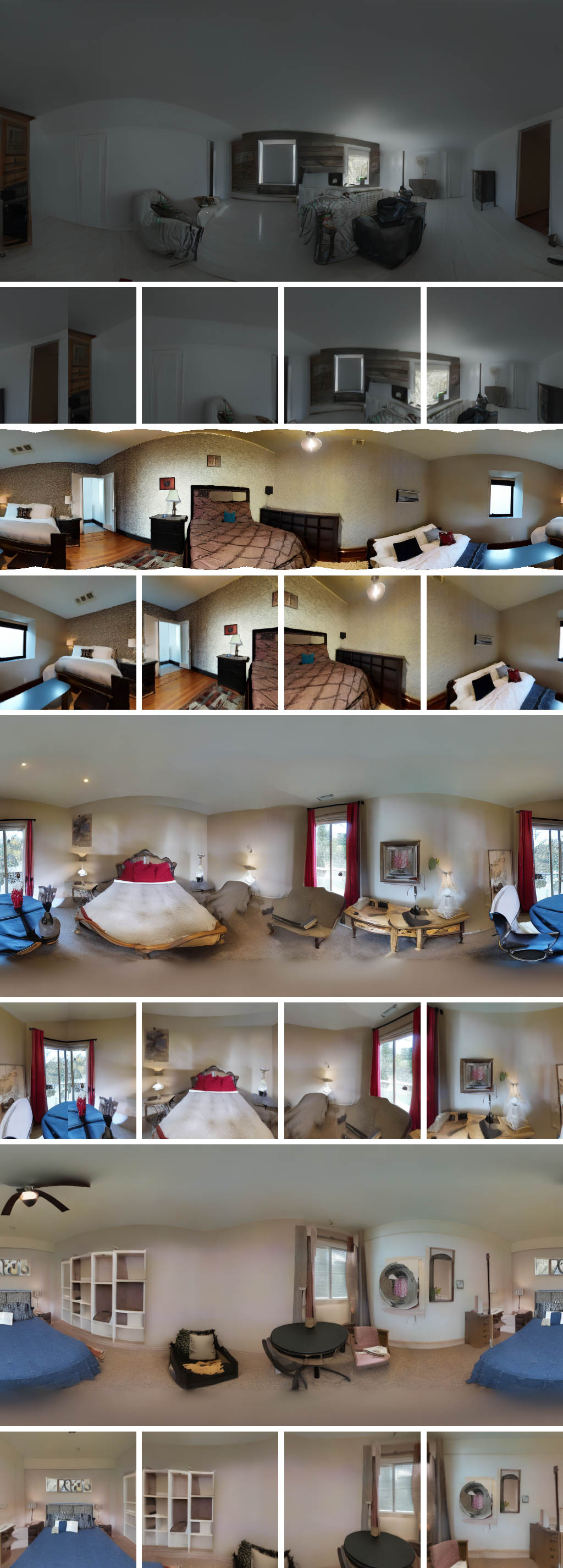}
{A living room with a chandelier}
{A bedroom with a bed and a table}
{1}

\compfig
{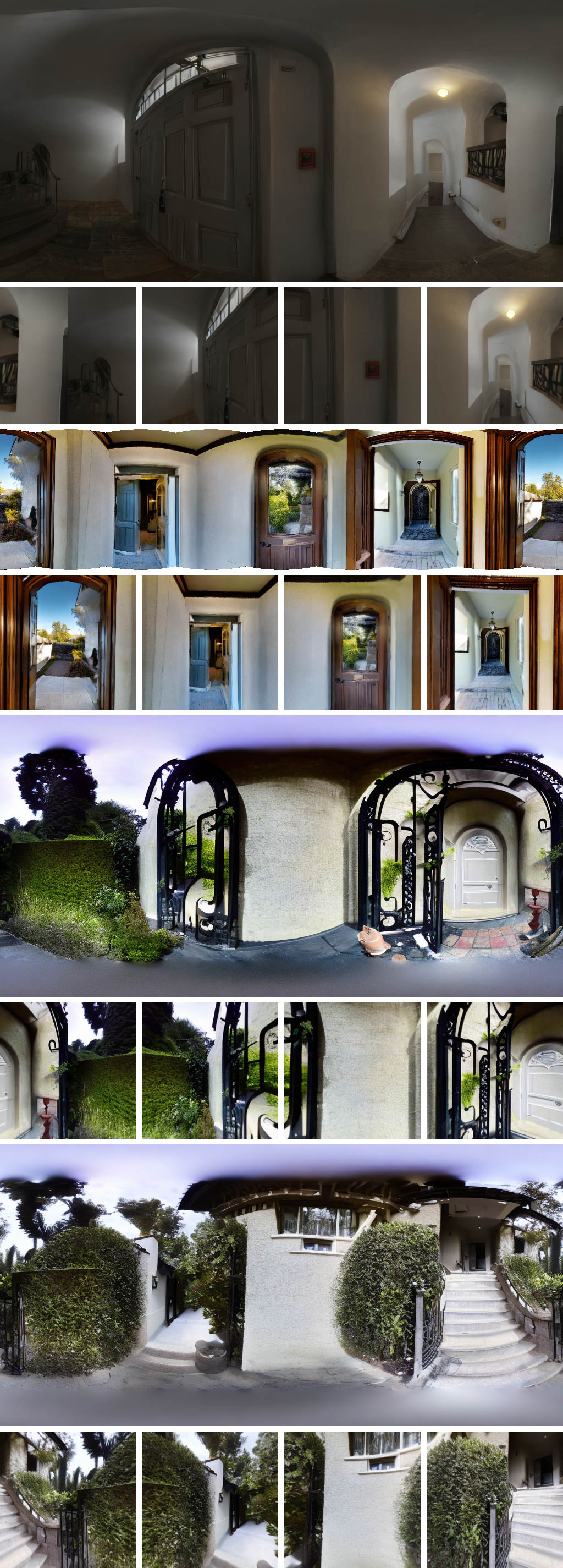}
{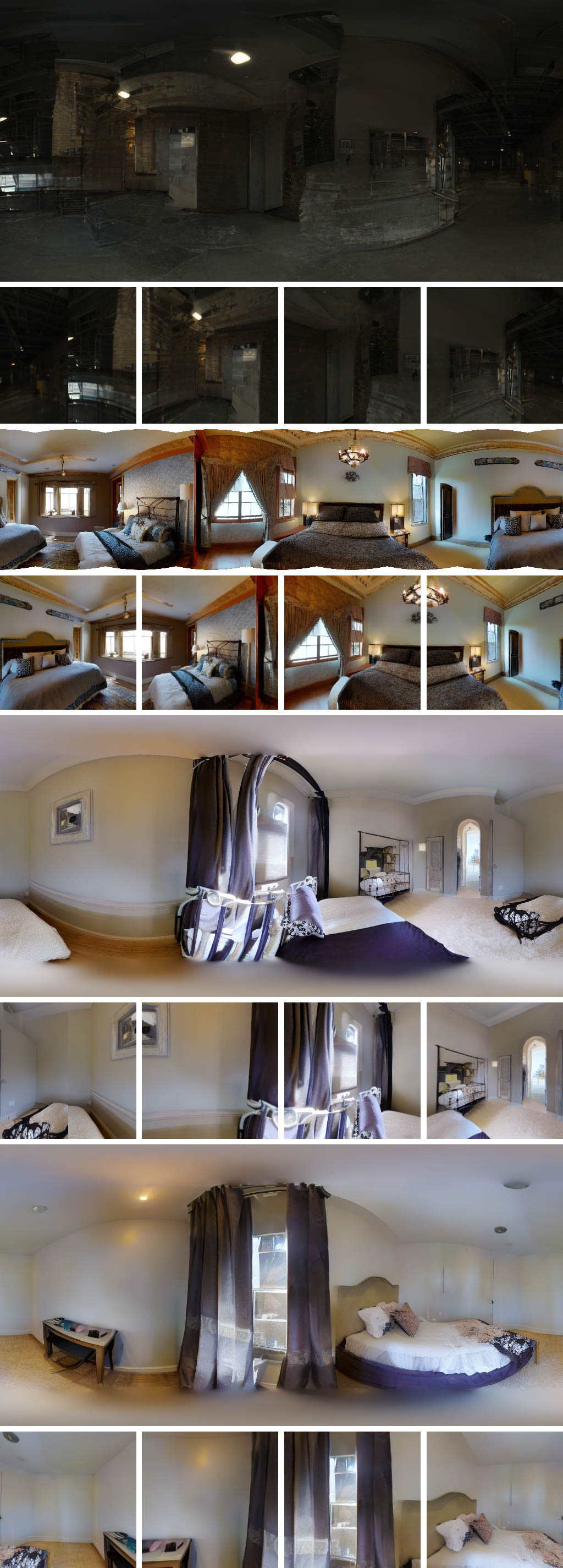}
{An entrance to a house}
{A bedroom with a bed}
{2}

\compfig
{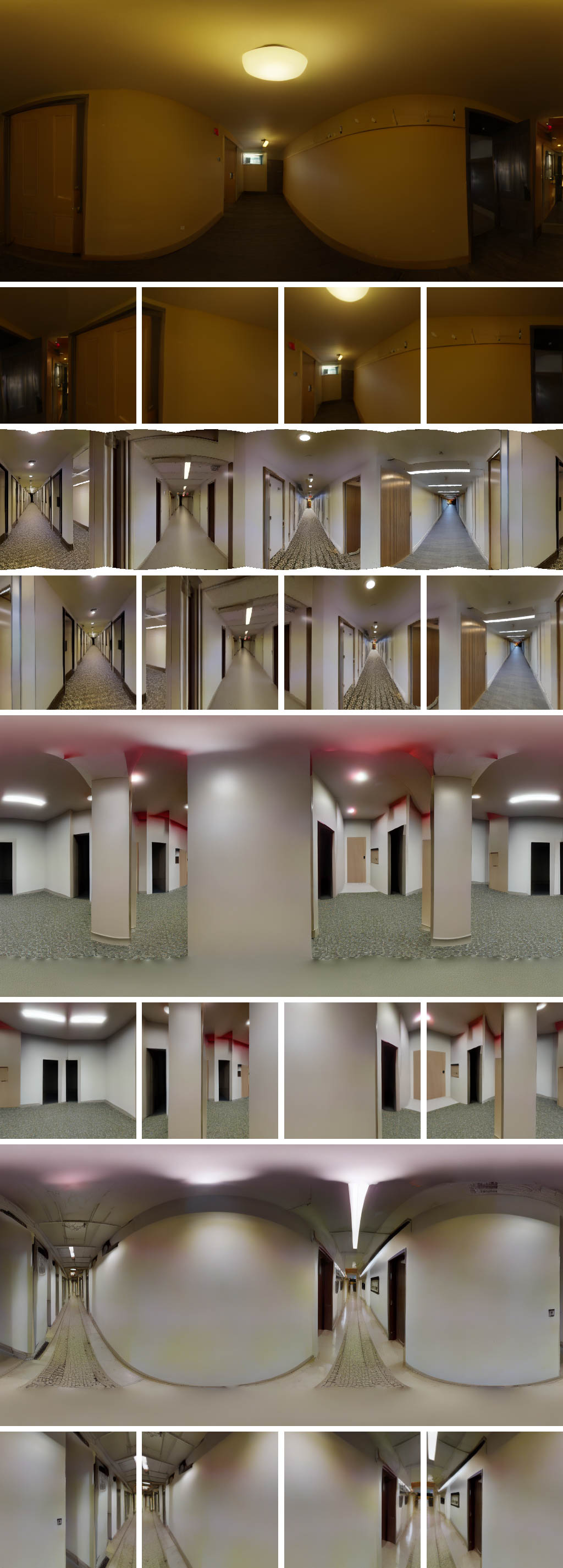}
{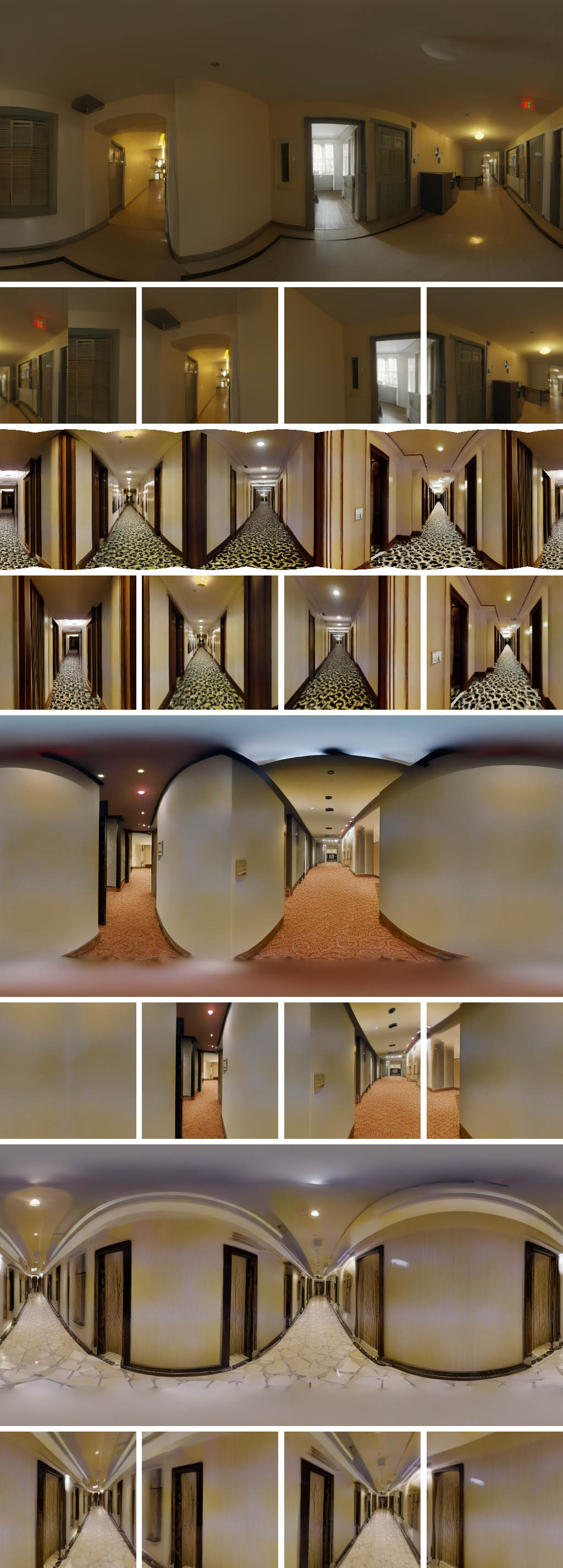}
{A hallway in a building}
{A hallway in a hotel}
{3}

\compfig
{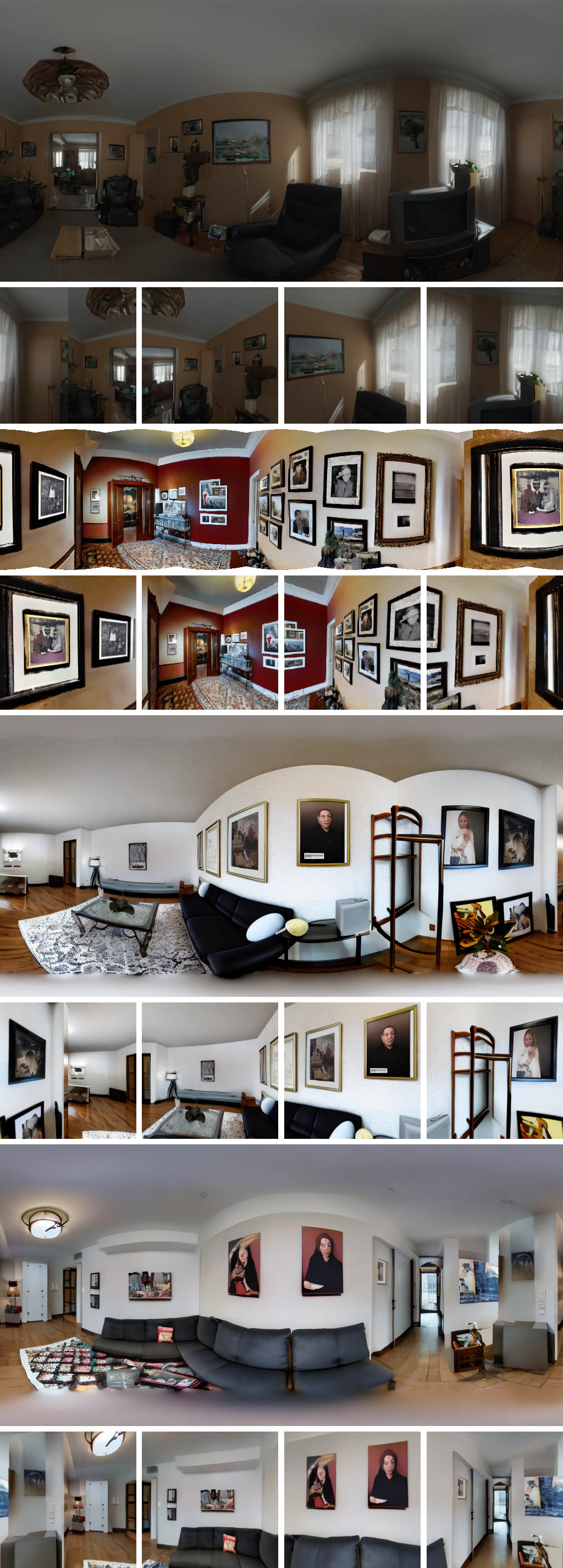}
{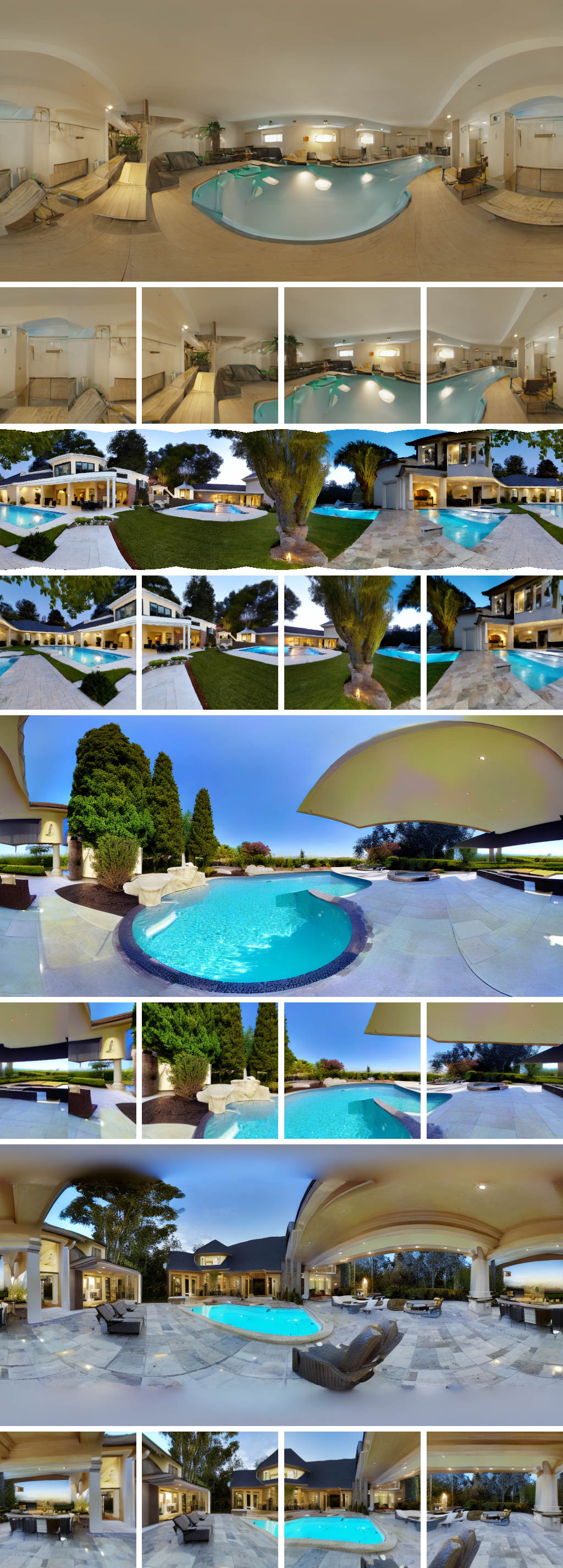}
{A living room with pictures on the wall}
{A home with a pool and patio}
{4}

\compfig
{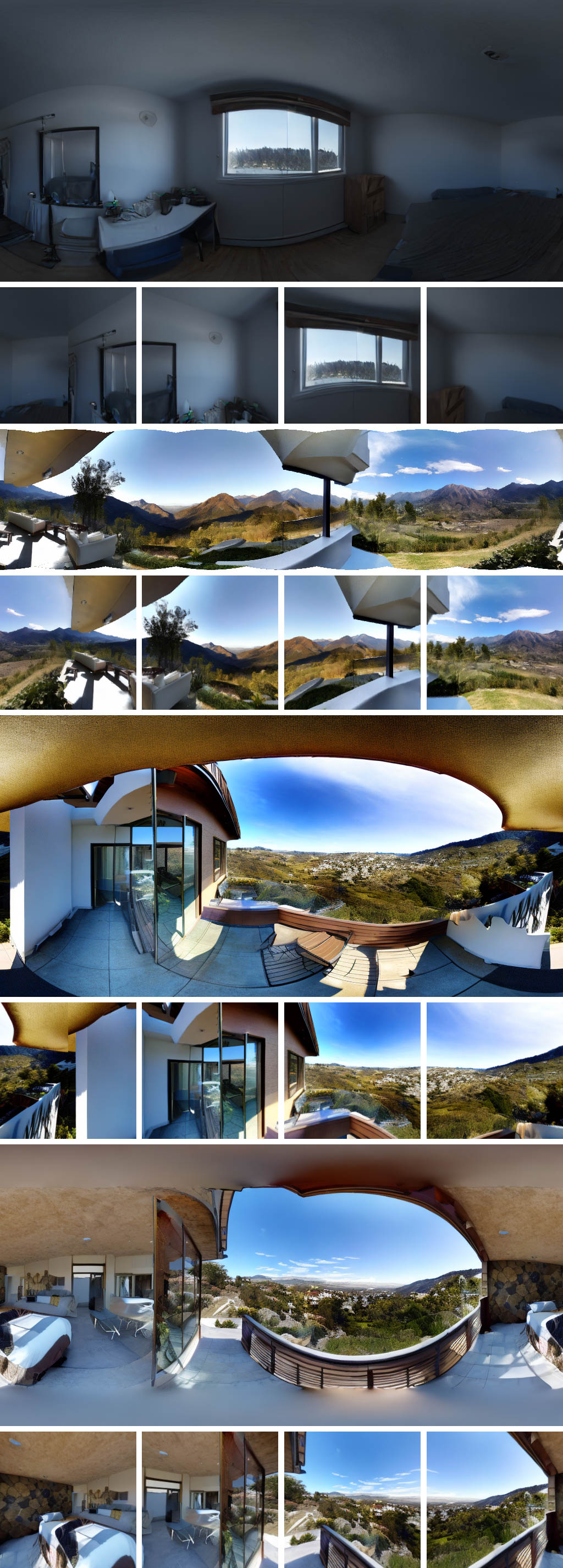}
{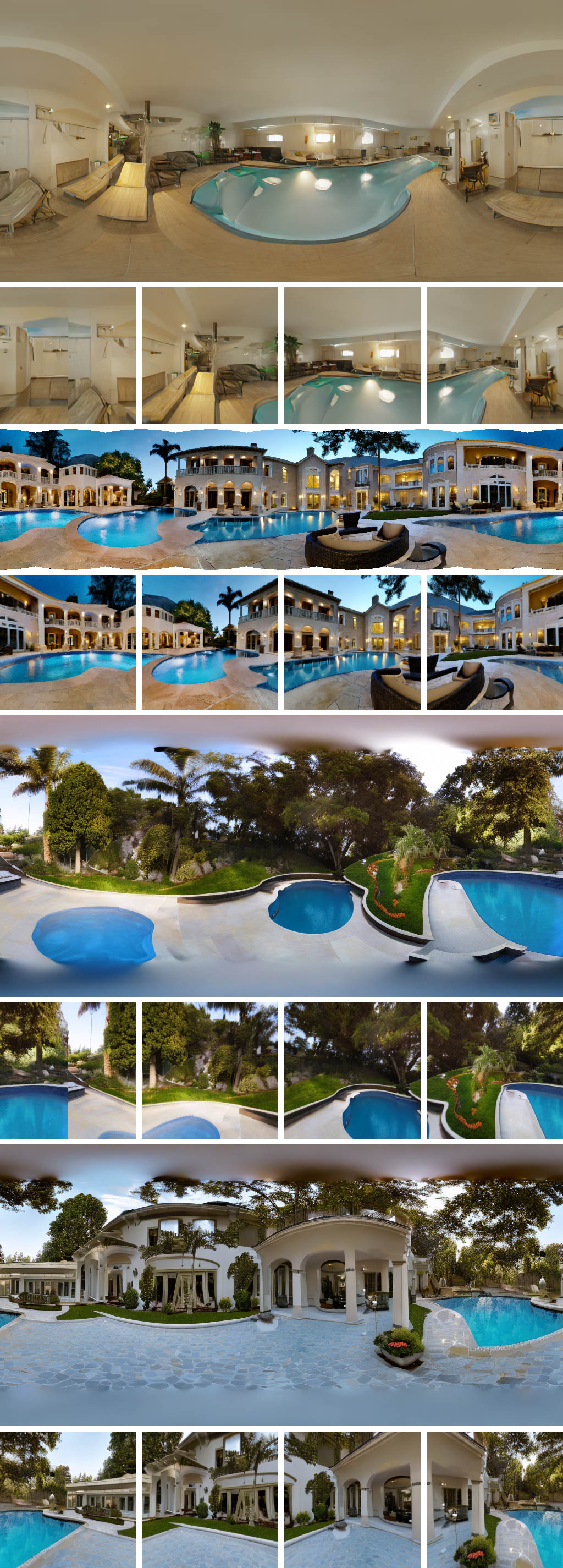}
{A house with a view of the mountains}
{A large home with a pool}
{5}

\compfig
{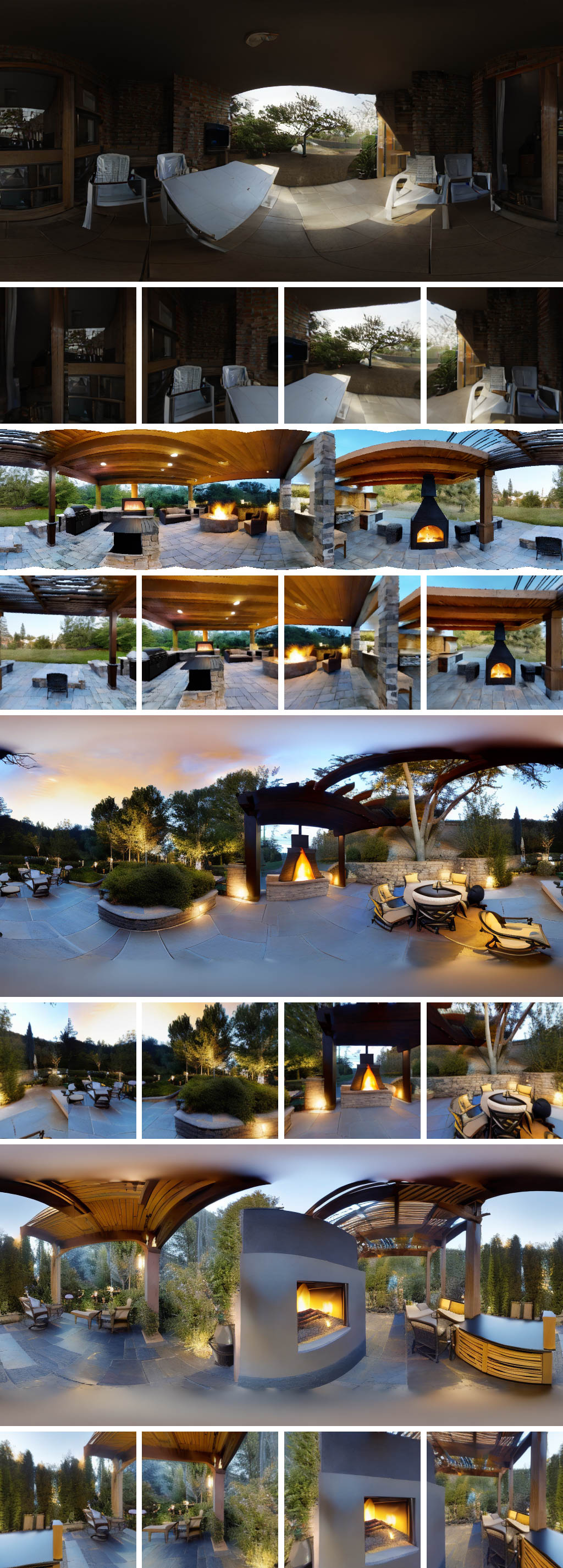}
{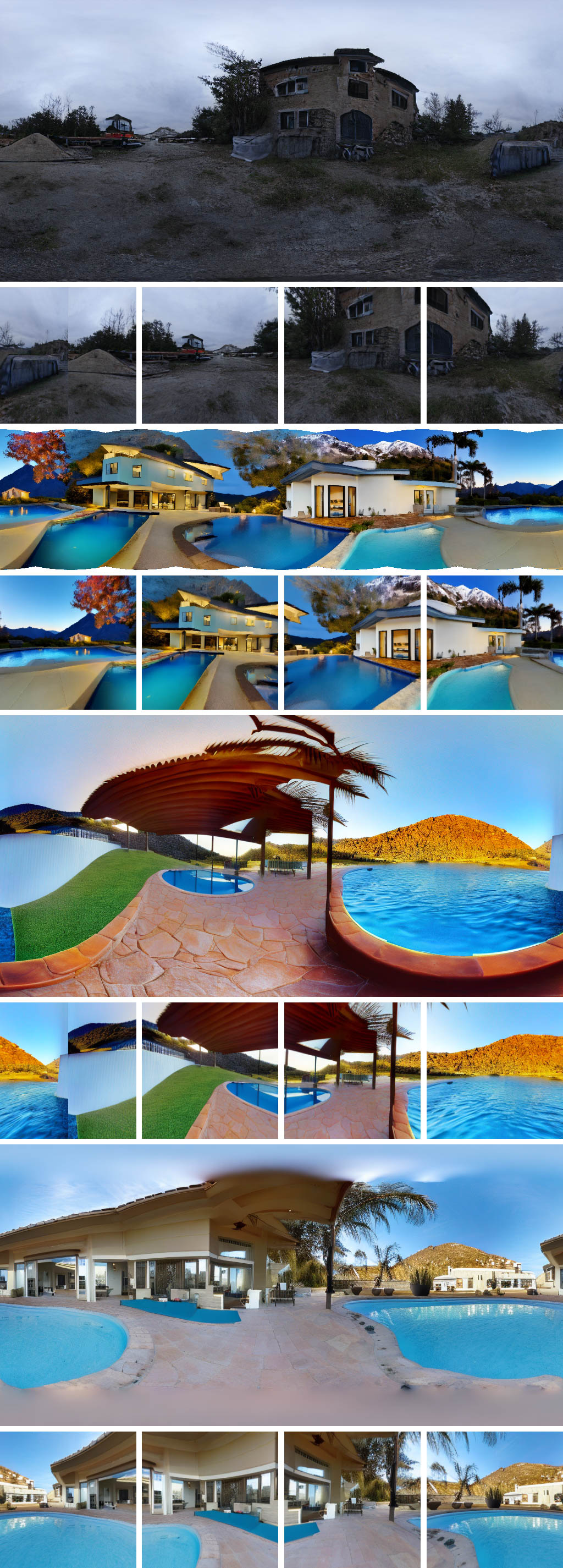}
{An outdoor patio with a fireplace}
{A house with a pool and mountains in the background}
{6}

\compfig
{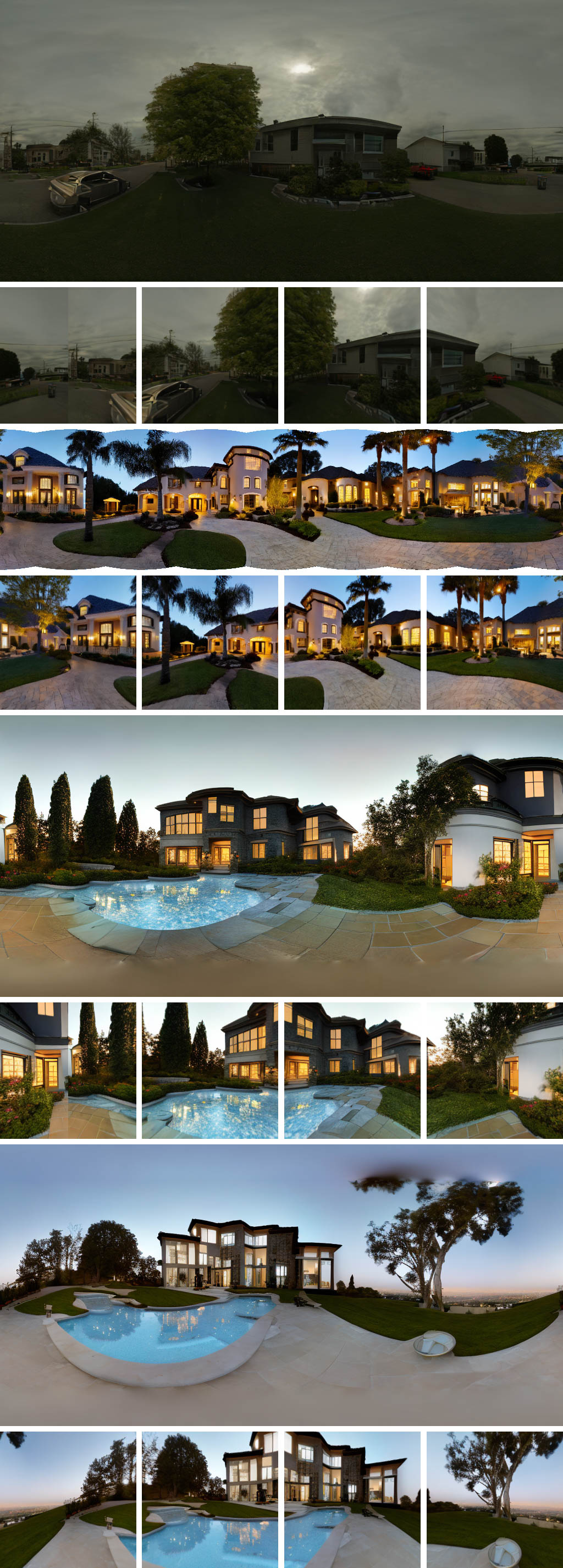}
{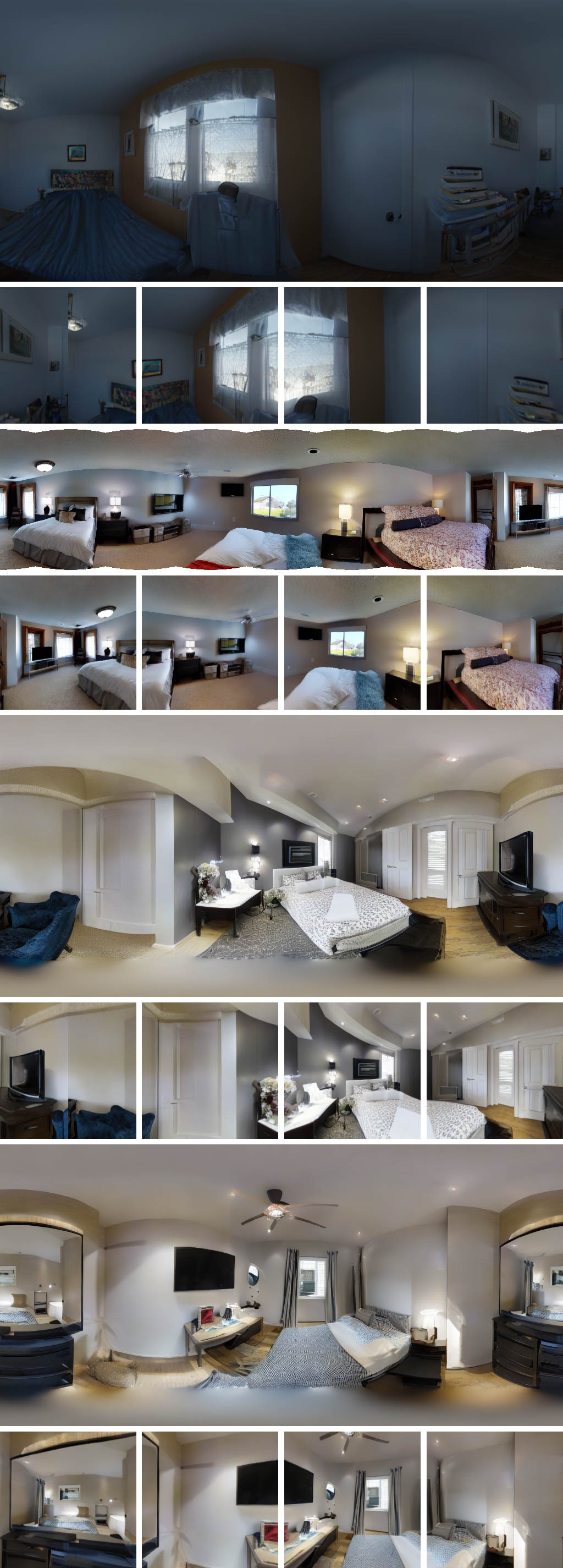}
{A luxury home at dusk}
{A bedroom with a bed and TV}
{7}

\compfig
{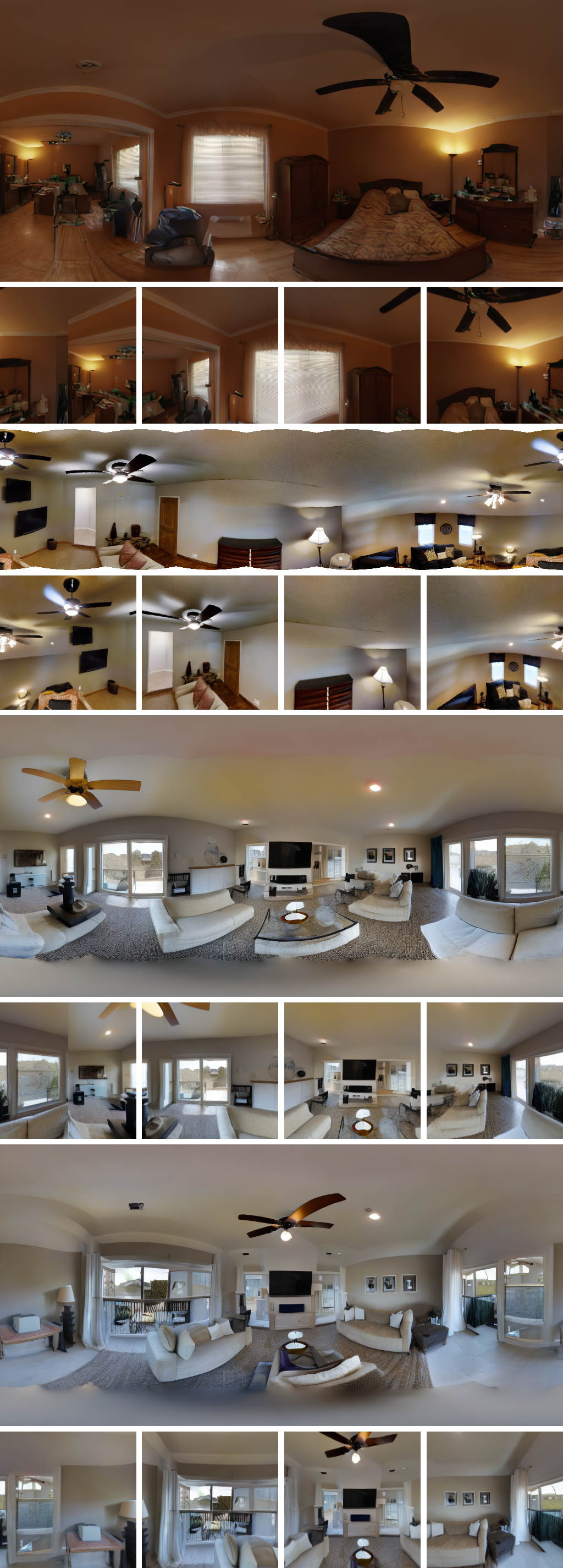}
{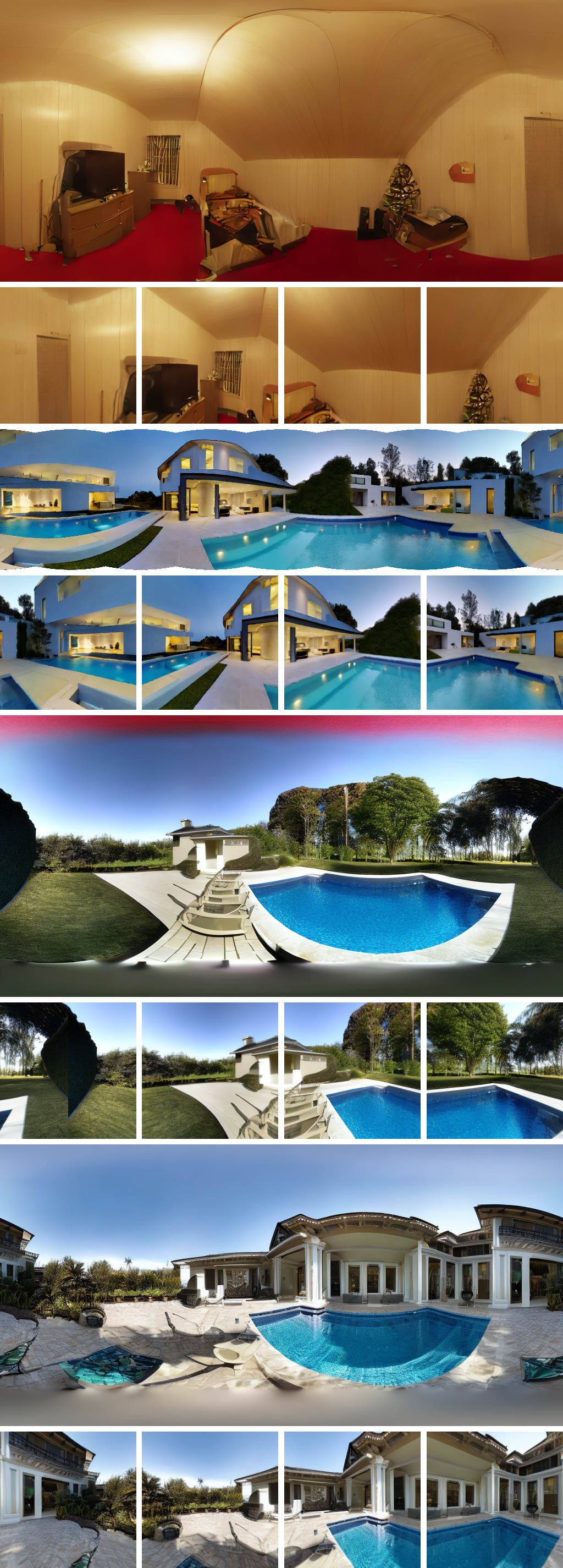}
{A living room with a ceiling fan}
{A house with a pool}
{8}

\compfig
{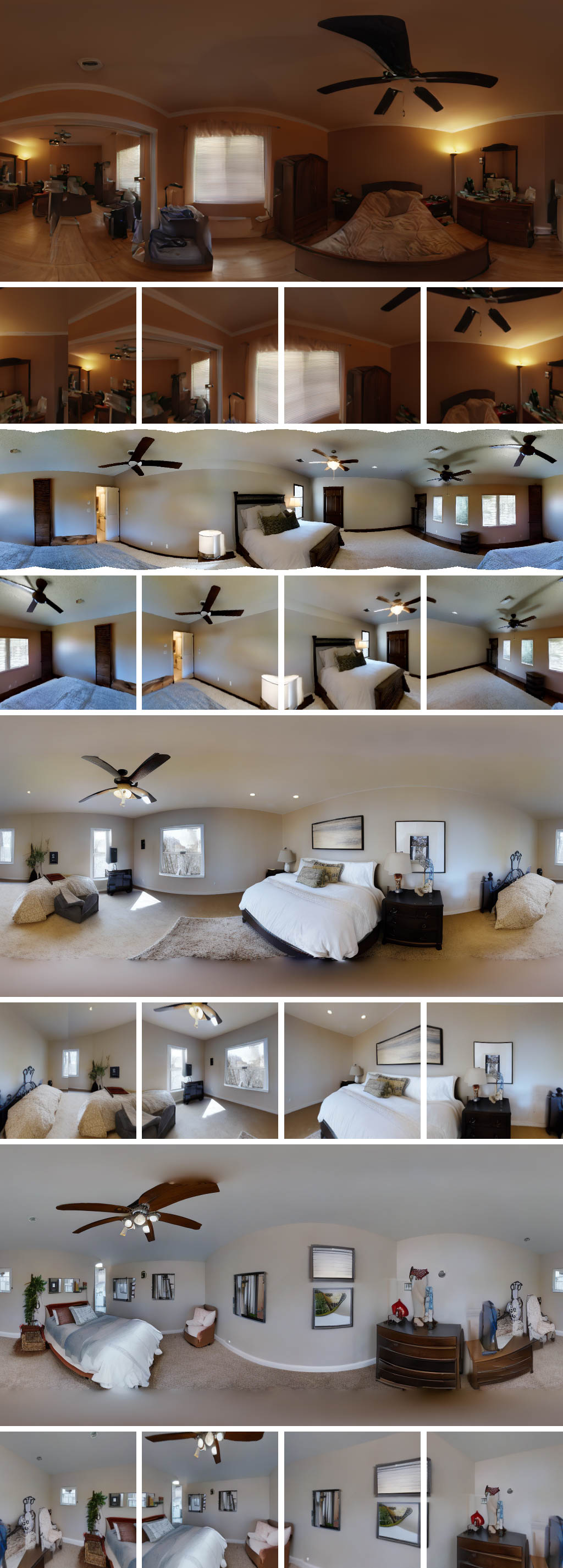}
{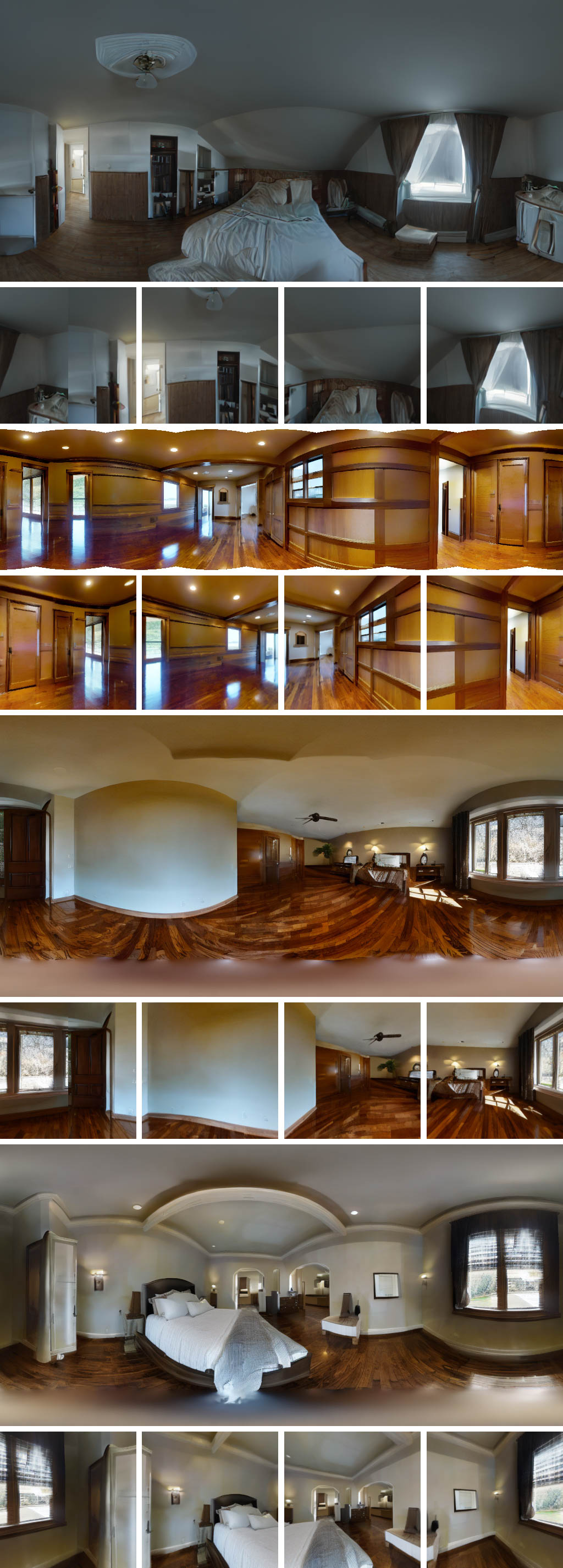}
{A bedroom with a ceiling fan}
{A bedroom with hardwood floors}
{9}

\compfig
{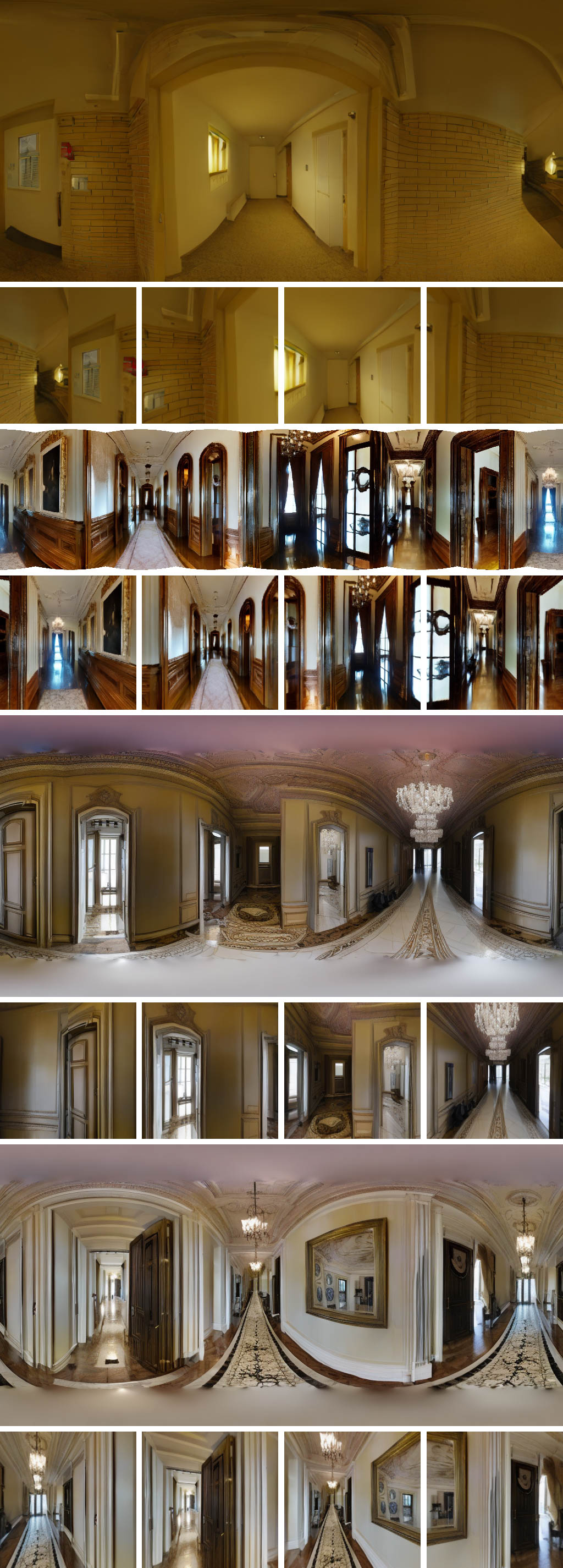}
{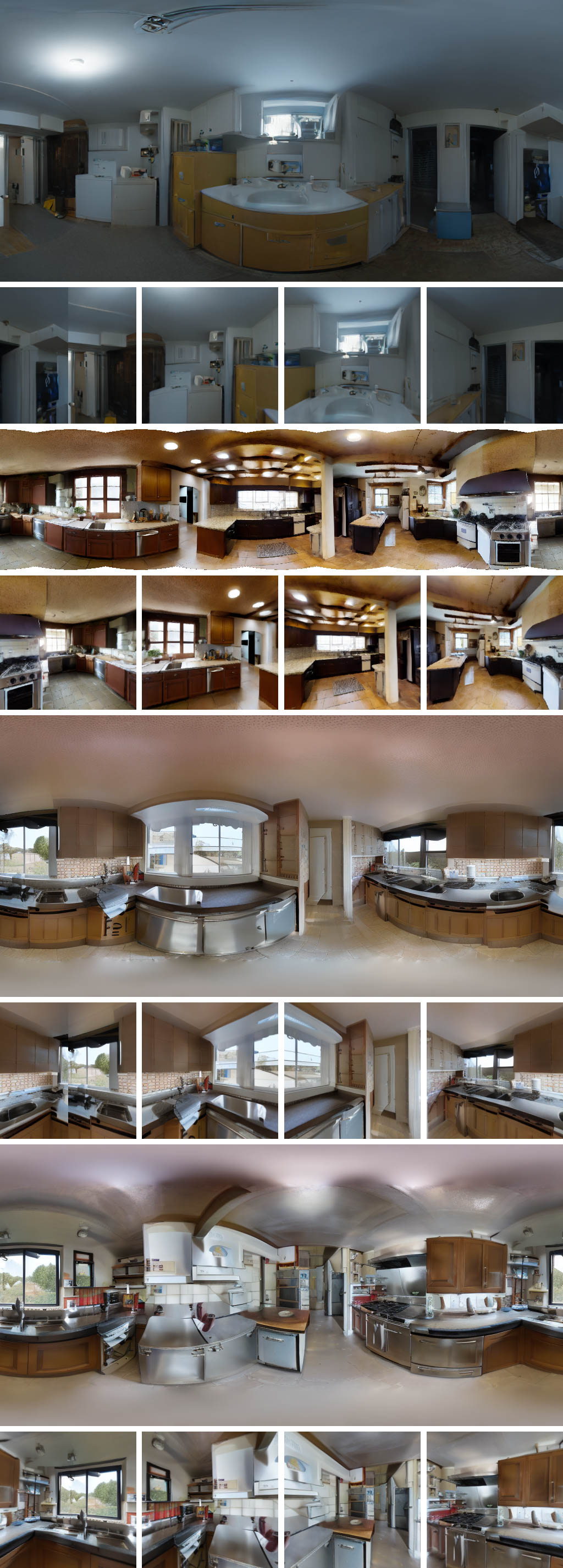}
{A hallway in a mansion}
{The inside of a kitchen}
{10}


In \cref{sec:comp} \cref{fig:comp_lowres} of the main paper, we compared our method with previous methods qualitatively.
Due to space limitations, we cropped the generated images to the vertical FoV of MVDiffusion for all methods.
Here we provide more qualitative comparisons without cropping in \crefrange{fig:more_comp_1}{fig:more_comp_10}, where \cref{fig:more_comp_1} has the same prompts as \cref{fig:comp_lowres} in the main paper.
Similarly, we evenly sample 4 horizontal views from the generated panorama for each panorama, in which the first view crosses the left and right borders to show how loop consistency is handled.

\section{Layout Conditioned Generation Details}\label{sec:suppl_layout_cond}
{
\newcommand{\layoutfig}[2]{
    \begin{figure*}[t]
        \vspace{-2em}
        \centering
        \small
        \setlength\tabcolsep{0px}
        \begin{tabular}{rc}
            #1
        \end{tabular}
        \vspace{-0.5em}
        \caption{More layout-conditioned generation comparisons.}\label{fig:more_layout_cond_#2}
        \vspace{-0.5em}
    \end{figure*}
}

\newcommand{\pano}[1]{
    \raisebox{-0.5\height}{\includegraphics[width=.9\linewidth,clip,trim=-10 -10 0 0]{#1}}
}

\newcommand{\row}[2]{
    \rotatebox[origin=c]{90}{
        \makecell[c]{
            \newcolumntype{C}[1]{>{\hsize=##1\hsize\centering\arraybackslash}X}
            \begin{tabularx}{.35\linewidth}{C{1}C{1}}
                PanFusion (Ours) & MVDiffusion
            \end{tabularx}
        }
    }
    & \pano{figure/more_layout_cond/#1}
    \\
    & ``#2.''
    \\
}

\layoutfig
{
    \row
    {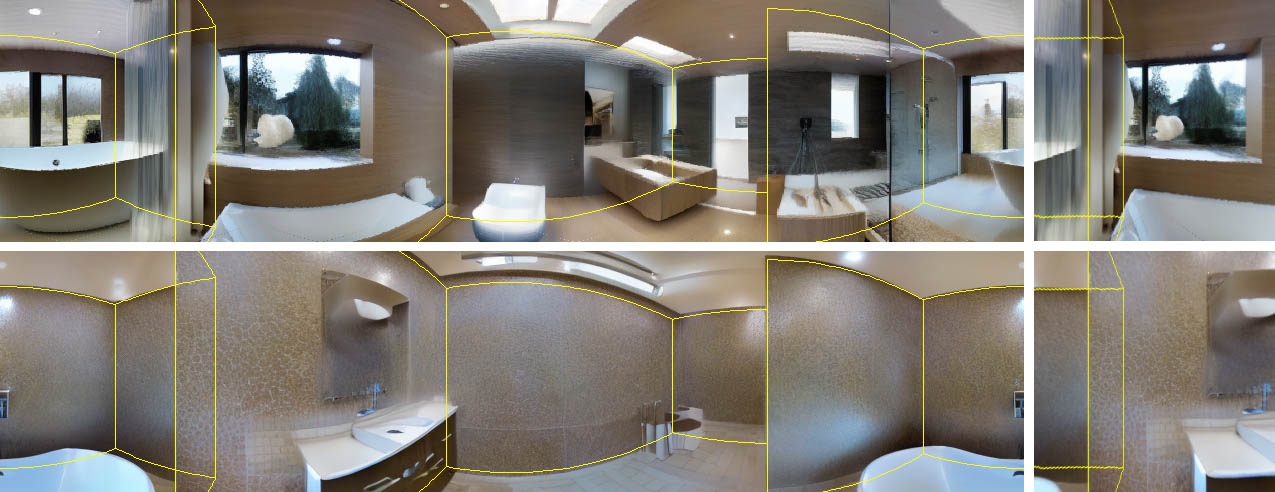}
    {A bathroom with a tub and sink}
    \row
    {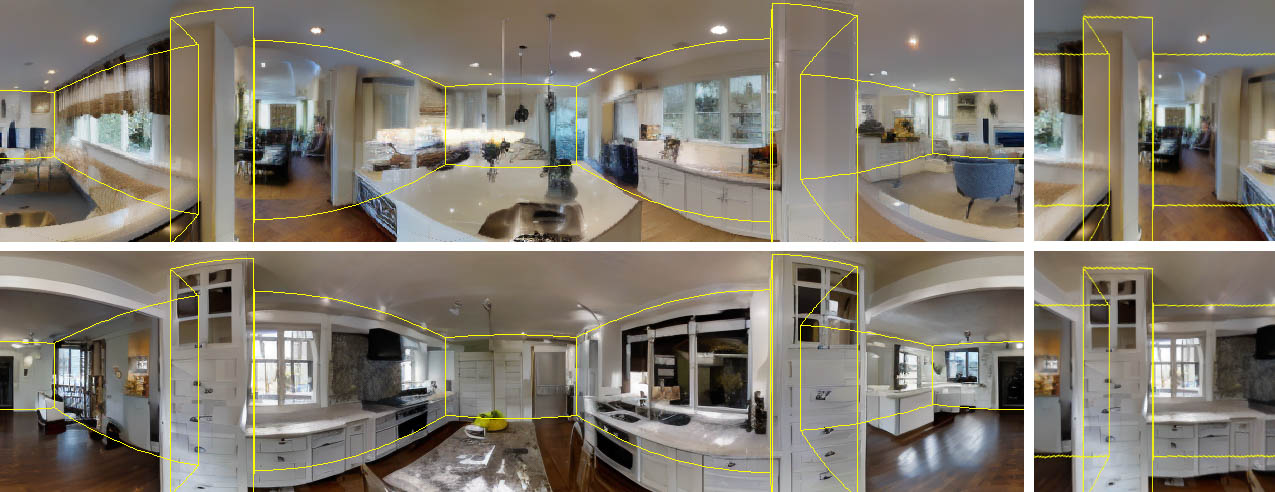}
    {A kitchen and dining room}
    \row
    {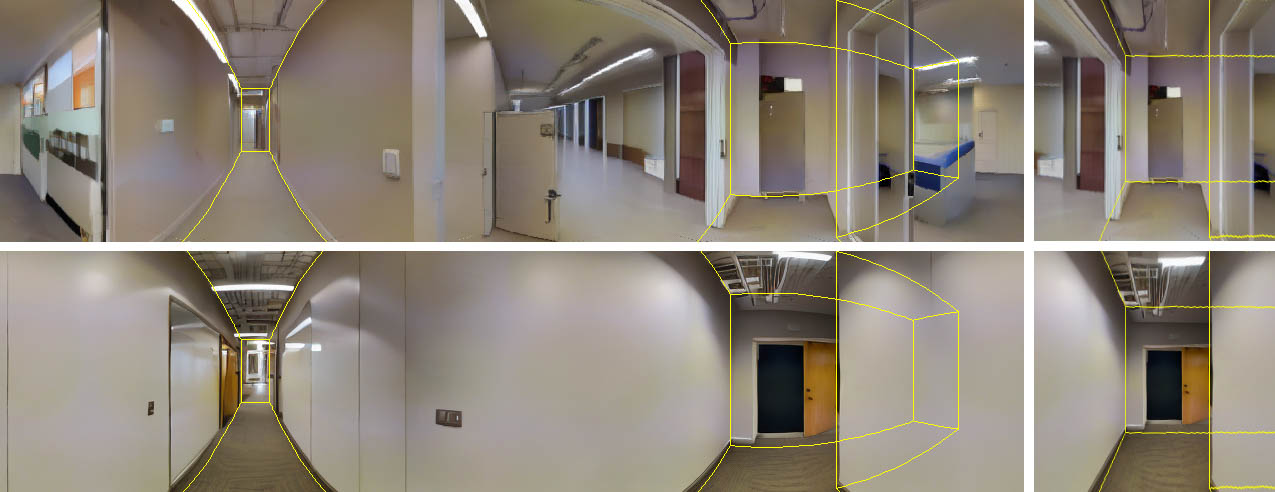}
    {A hallway in an office}
}{1}

\layoutfig
{
    \row
    {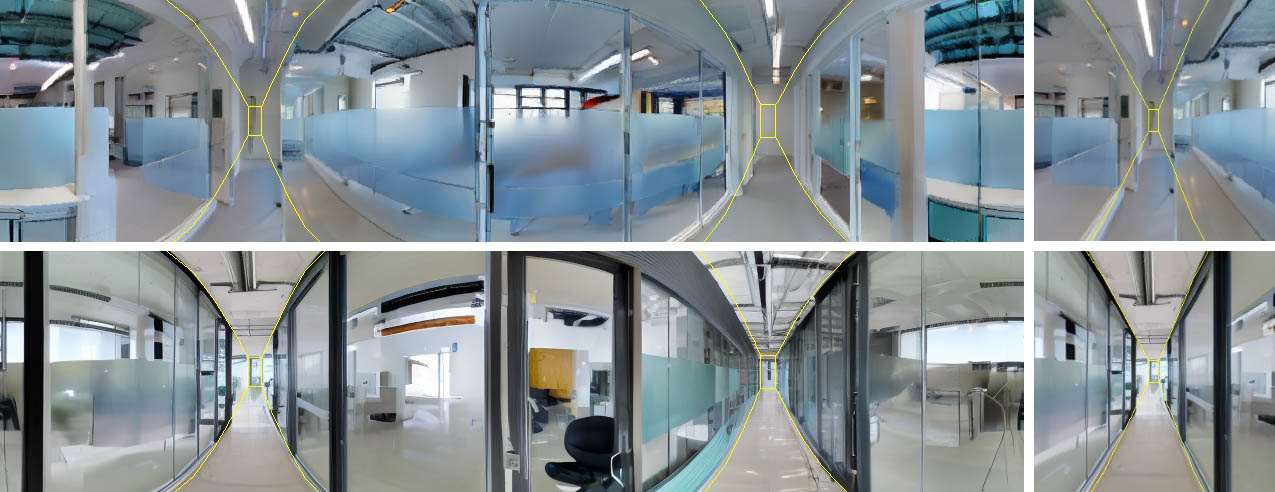}
    {An office with glass walls}
    \row
    {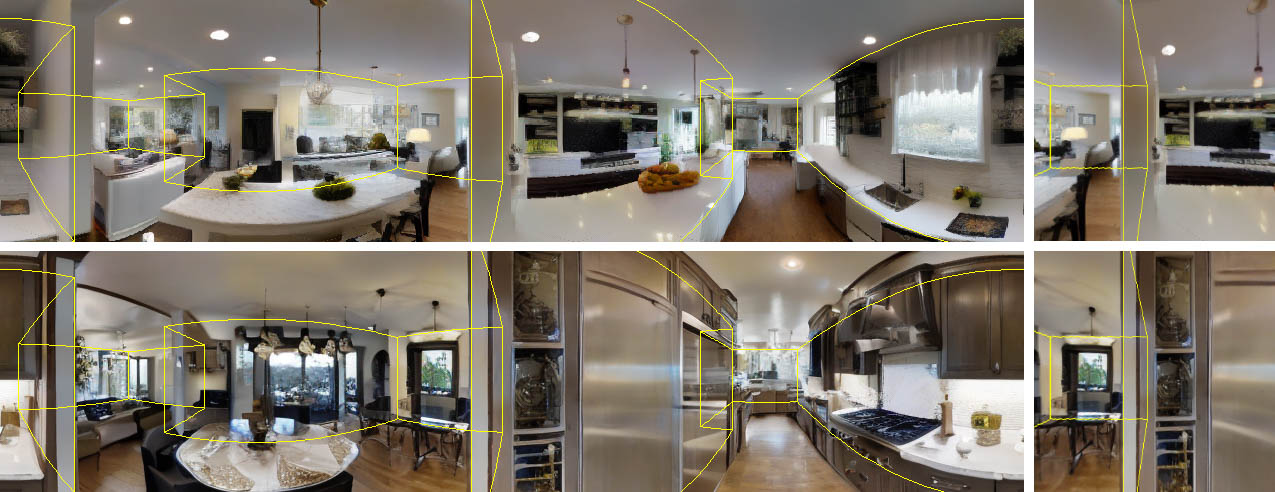}
    {A kitchen and dining room}
    \row
    {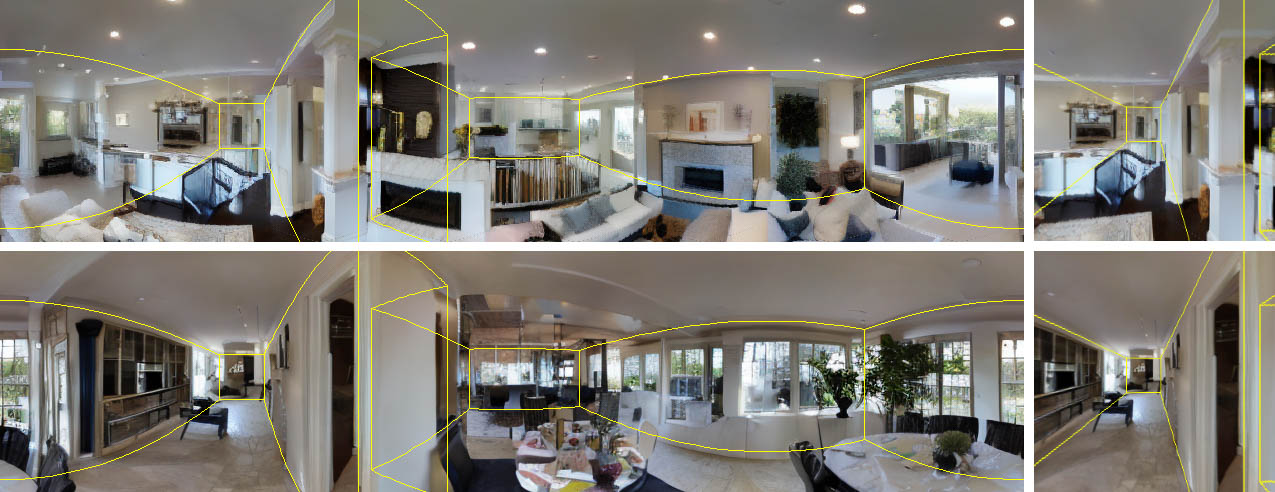}
    {A living room and dining room}
}{2}

}

\begin{table}[t]
    \centering
    \resizebox{1.\linewidth}{!}{
    \begin{tabular}{lccccc}
        \multirow{2}{*}{Method}
        & \multicolumn{2}{c}{Layout Consistency}
        & \multicolumn{3}{c}{Horizontal 8 Views~\cite{tang2023mvdiffusion}}
        \\
        \cmidrule(r){2-3} \cmidrule(r){4-6}
        & 3D IoU $\uparrow$ & 2D IoU $\uparrow$
        & FID $\downarrow$ & IS $\uparrow$ & CS $\uparrow$
        \\
        \midrule
        SD+LoRA~\cite{rombach2022highresolution,hu2021lora}
        & 68.02 & 71.41
        & \textbf{21.39} & 5.03 & 25.84
        \\
        {\name} (Ours) 
        & \textbf{68.46} & \textbf{71.82}
        & 22.58 & \textbf{5.10} & \textbf{26.04}
    \end{tabular}
    }
    \vspace{-1em}
    \caption{
        Layout-conditioned comparison with SD+LoRA.
        Our method achieves comparable or better results.
    }\label{tbl:more_layout_cond}
\end{table}

In \cref{sec:layout_cond} of the main paper, we showcased the benefits of our dual-branch method with the application of layout conditioned generation.
Specifically, the room layout is rendered as a distance map, as shown in \cref{fig:mp3d_layout}, and normalized to the range of $[-1, 1]$ as an additional spatial condition.
To add layout condition to MVDiffusion, we follow~\cite{song2023roomdreamer} to project the layout condition to perspective views as a distance map instead of a depth map to ensure consistency among overlapped regions.
However, when training the ControlNet for MVDiffusion at the original resolution of $512 \times 512$, it suffers from gradient explosion.
Instead, we found finetuning MVDiffusion with LoRA on a lower resolution of $256 \times 256$ can make the training of the ControlNet converge, and also improve the realism of MVDiffusion. 
Therefore, we use MVDiffusion+LoRA as the base model for layout conditioned generation in main paper \cref{tbl:layout_cond} to serve as a stronger baseline.
In \crefrange{fig:more_layout_cond_1}{fig:more_layout_cond_2}, we provide more quantitative comparison with MVDiffusion.
We also compare with SD+LoRA in \cref{tbl:more_layout_cond} to show that our method can get comparable or better results.

\section{Generalization to Out-domain Prompts}\label{sec:suppl_generalization}
{
\newcommand{\genfig}[2]{
    \begin{figure*}[t]
        \vspace{-2em}
        \centering
        \small
        \setlength\tabcolsep{0px}
        \begin{tabular}{cc}
            #1
        \end{tabular}
        \vspace{-0.5em}
        \caption{Generalization to out-domain prompts.}\label{fig:generalization_#2}
        \vspace{-0.5em}
    \end{figure*}
}

\newcommand{\pano}[1]{
    \raisebox{-0.5\height}{\includegraphics[width=0.45\linewidth,clip,trim=-4 -2 -4 -2]{#1}}
}

\newcommand{\row}[4]{
    \pano{figure/generalization/#1}
    & \pano{figure/generalization/#2}
    \\
    \parbox[t]{.45\linewidth}{\centering ``#3''}
    & \parbox[t]{.45\linewidth}{\centering ``#4''}
    \\
}

\genfig
{
    \row
    {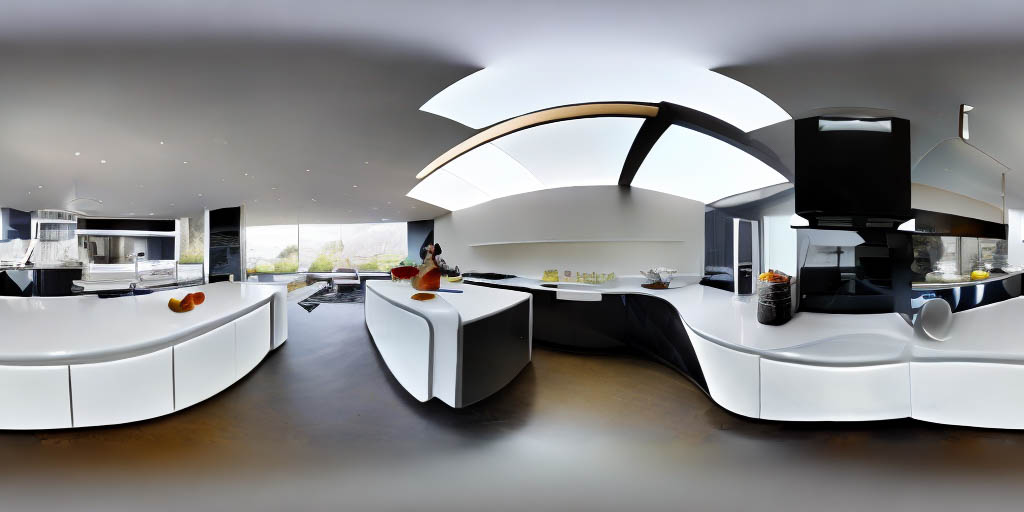}
    {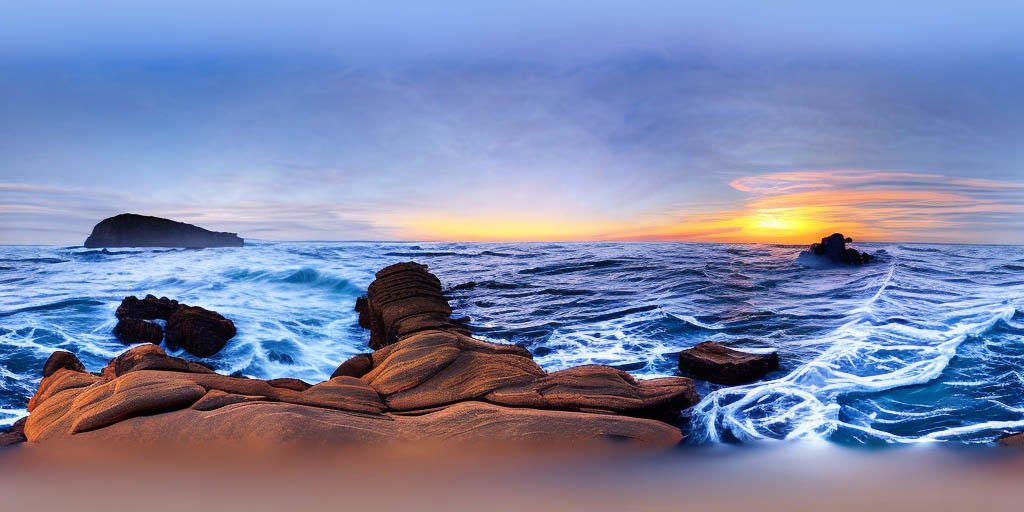}
    {A futuristic kitchen.}
    {Coastal cliff at sunset, waves crashing on rugged rocks.}
    \row
    {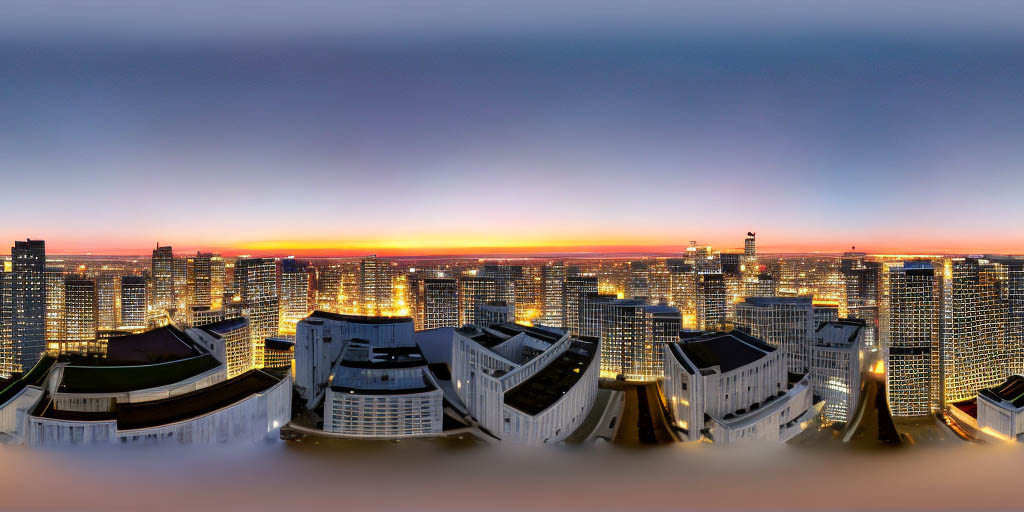}
    {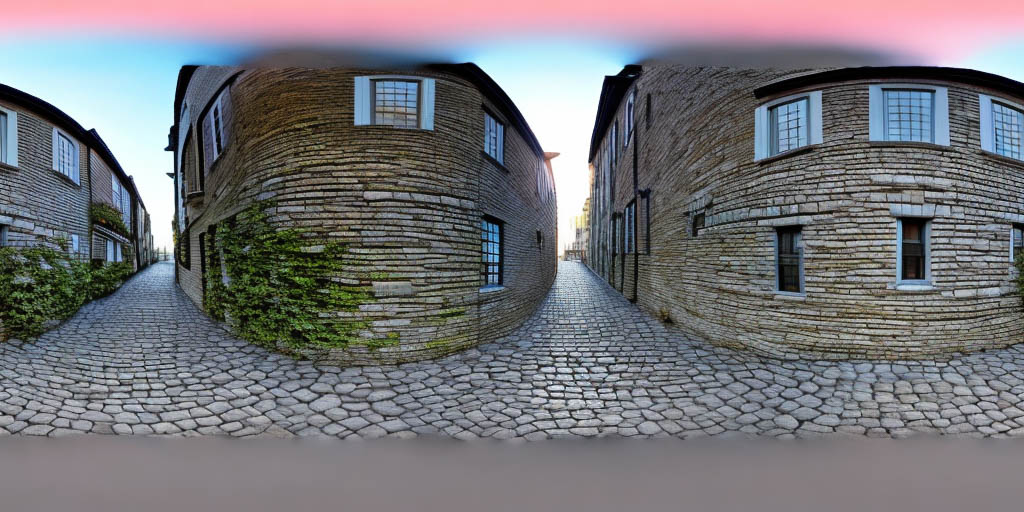}
    {Urban skyline at twilight, city lights twinkling in the distance}
    {Cobblestone alley, historic architecture bathed in soft morning light.}
    \row
    {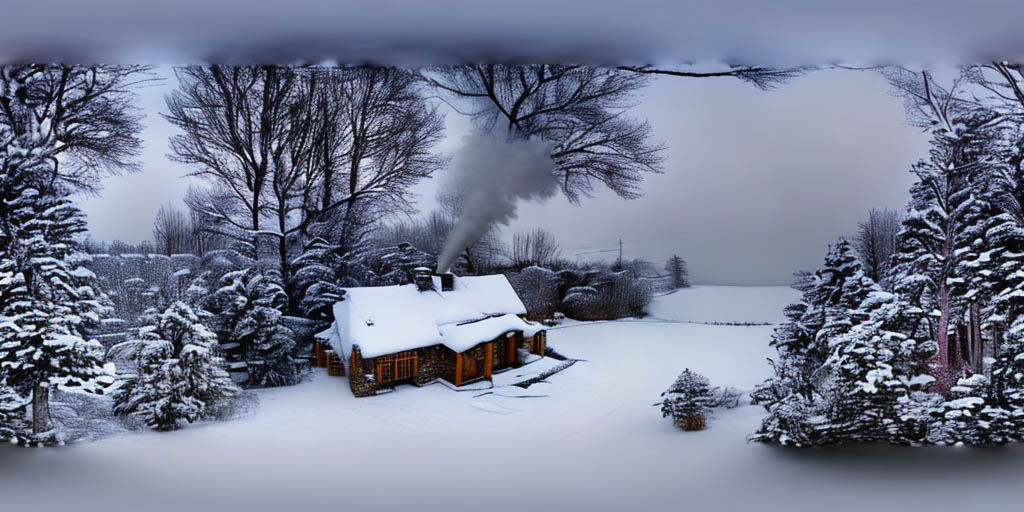}
    {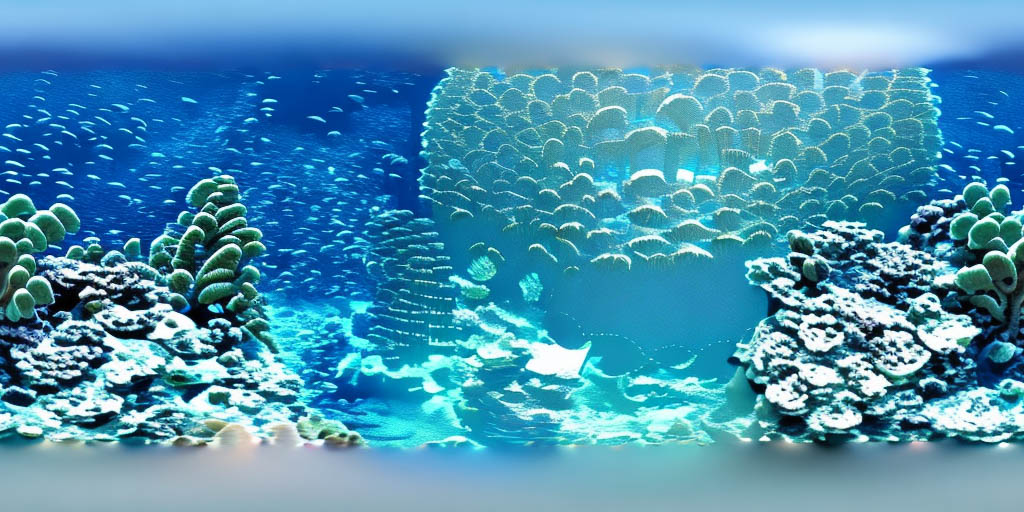}
    {Snow-covered cottage, smoke rising from a charming stone chimney.}
    {An underwater scene, where coral reefs teem with colorful fish beneath the clear blue ocean.}
    \row
    {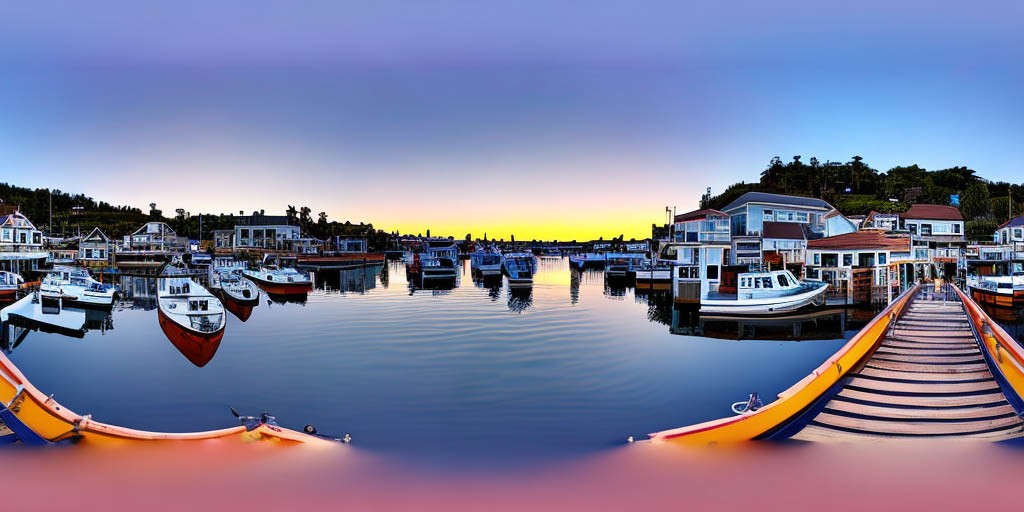}
    {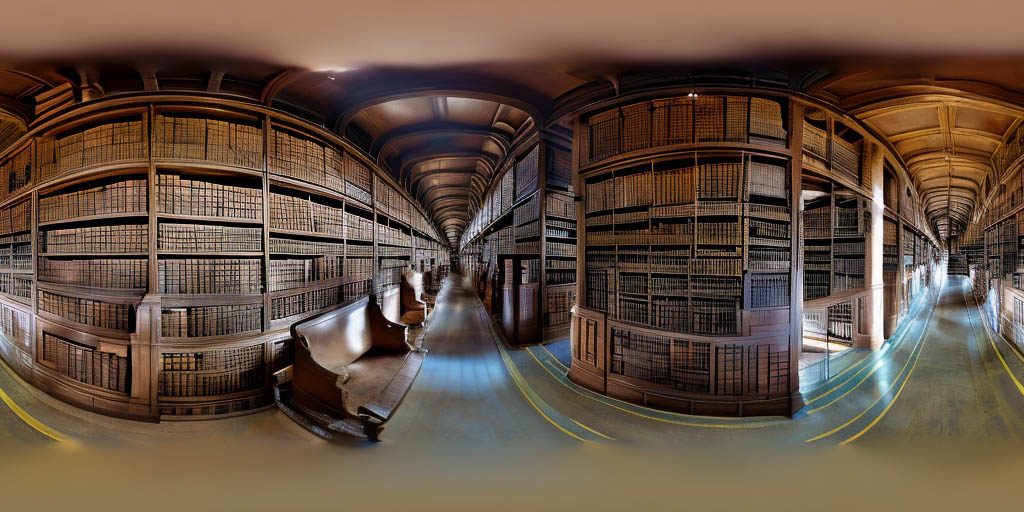}
    {A peaceful coastal village at sunrise, with fishing boats docked along the quiet harbor.}
    {The interior of a historic library, filled with rows of antique books, leather-bound and dust-covered.}
    \row
    {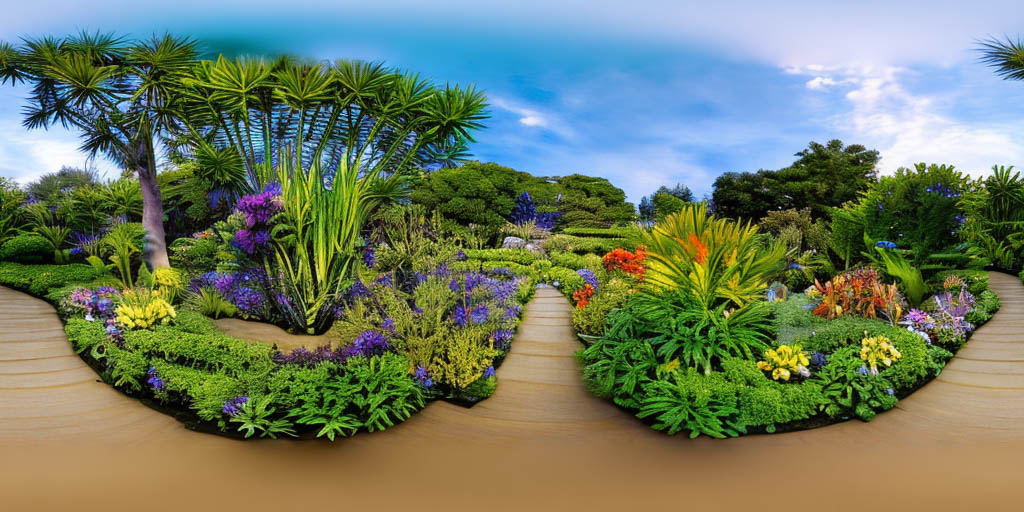}
    {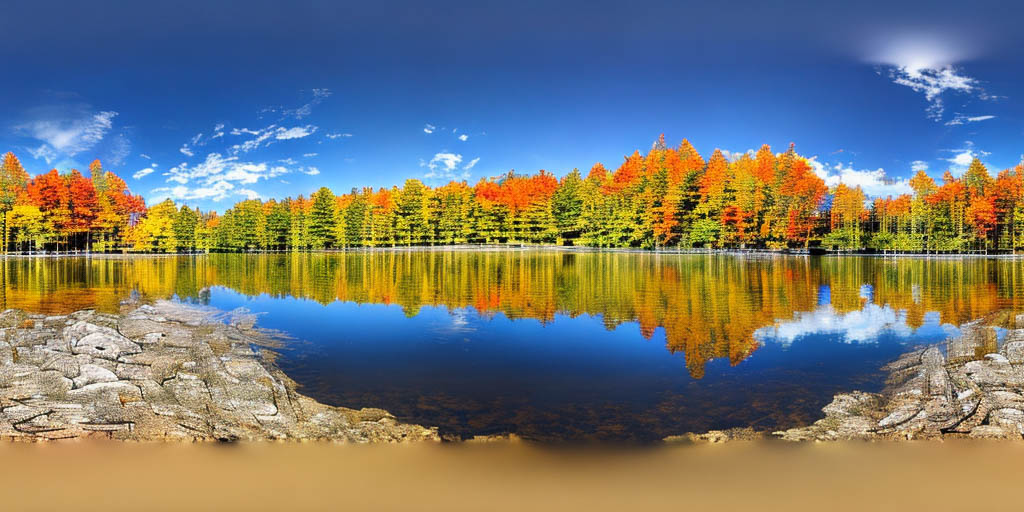}
    {A tranquil botanical garden, with exotic plants, blooming flowers, and meandering stone pathways.}
    {The calm waters of a secluded lake, reflecting the colors of the surrounding autumn foliage.}
}{1}

\genfig
{
    \row
    {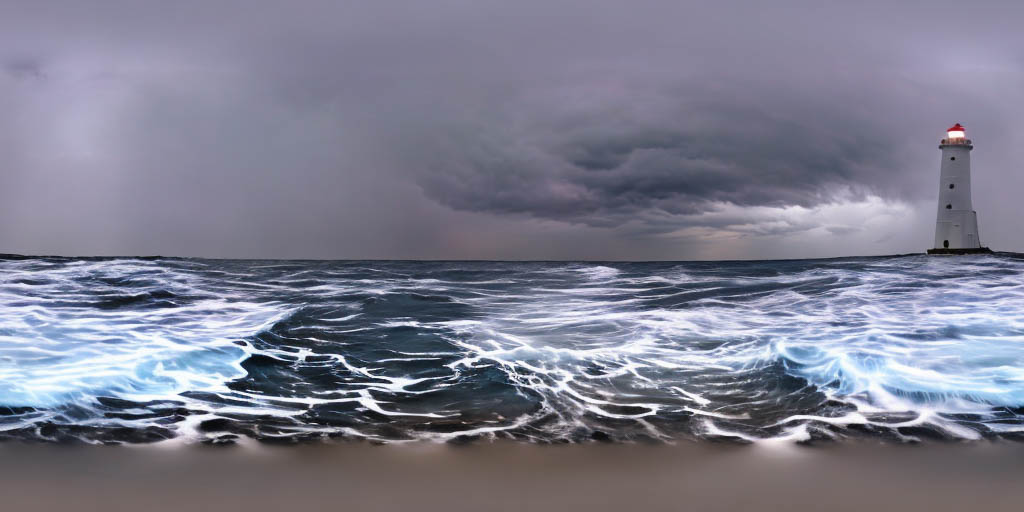}
    {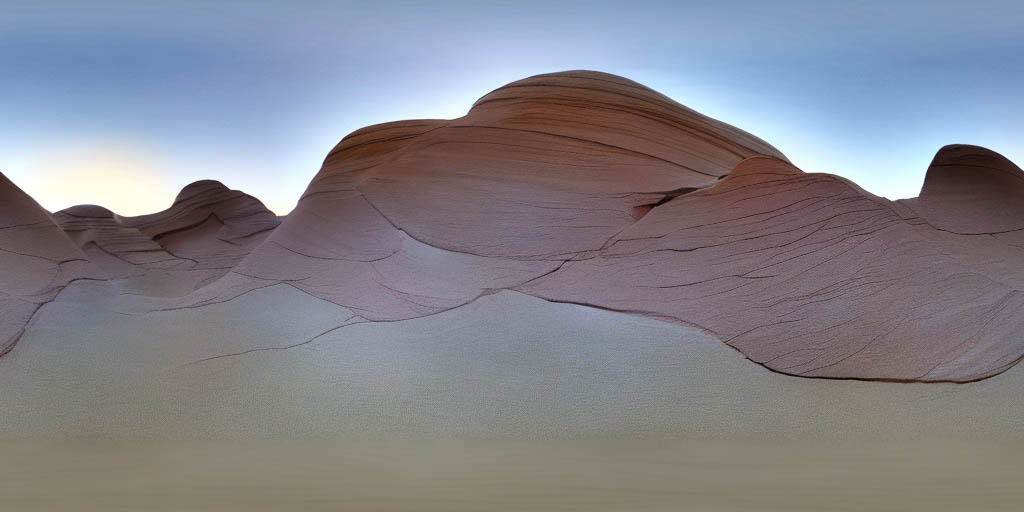}
    {Lighthouse in stormy seas.}
    {Desert canyon, sculpted sandstone.}
    \row
    {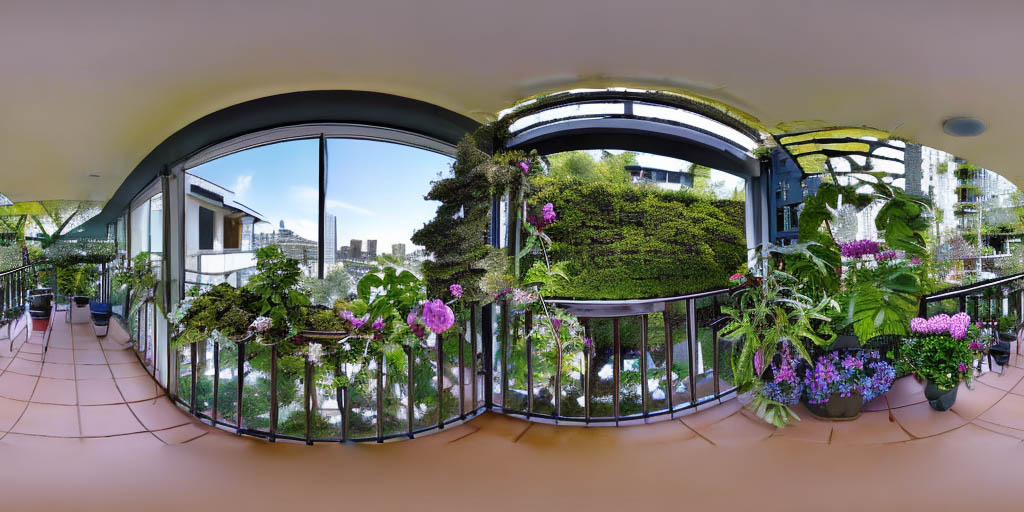}
    {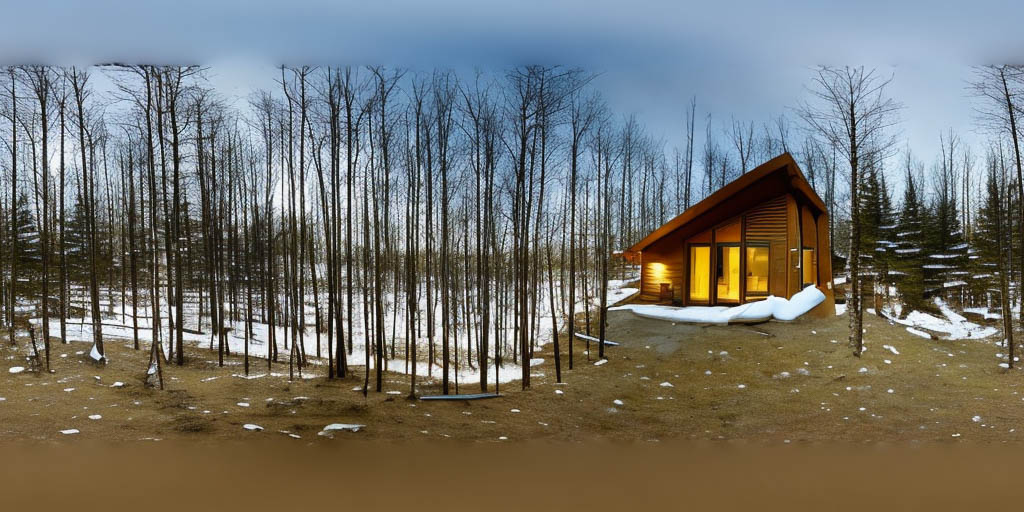}
    {Balcony garden, blooming serenity.}
    {Firelit cabin, crackling warmth amid snowy woods.}
    \row
    {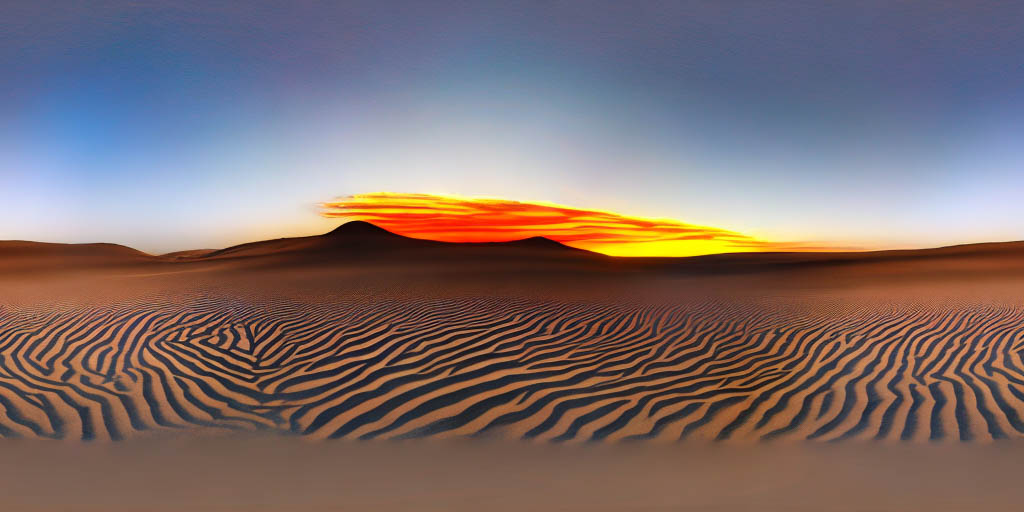}
    {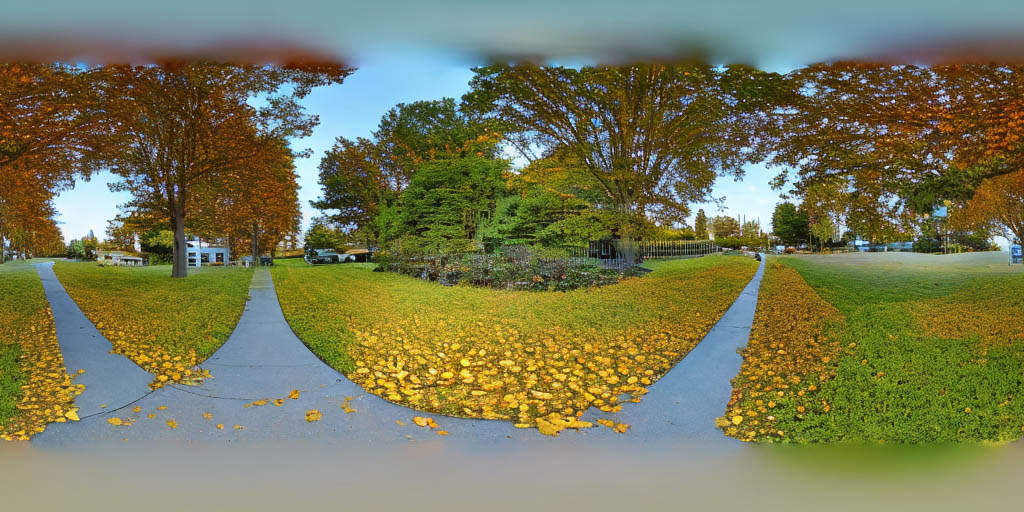}
    {Desert sunrise, silhouettes painted against the golden horizon.}
    {Suburban street, autumn leaves carpeting the sidewalk in hues.}
    \row
    {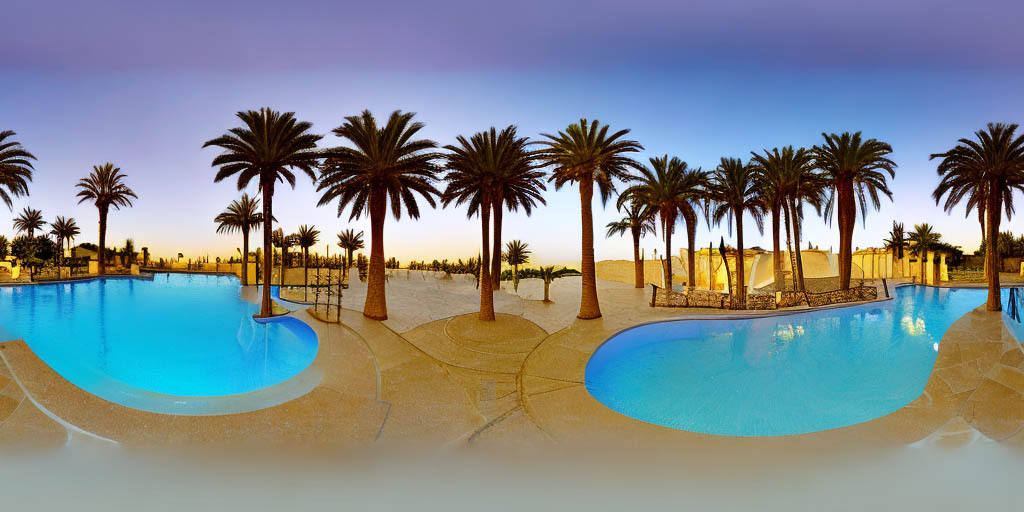}
    {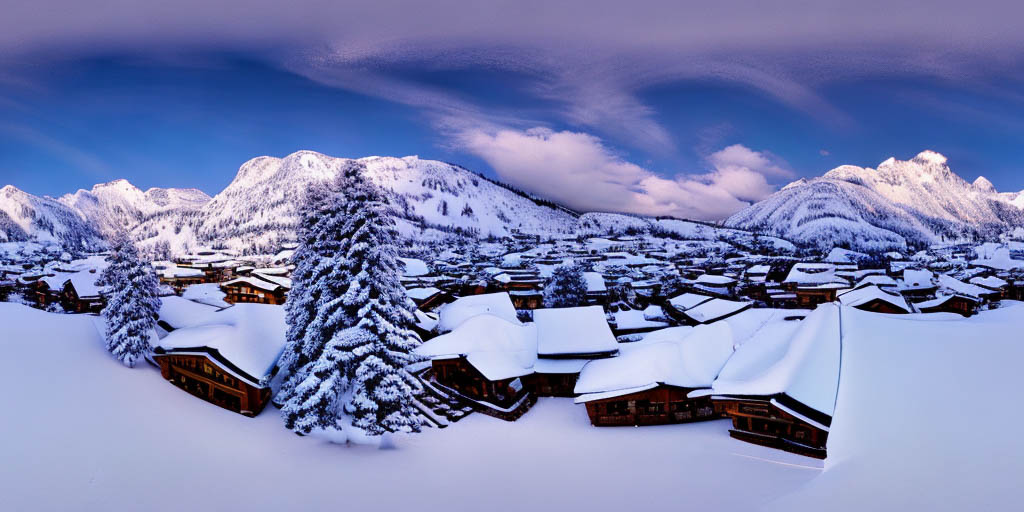}
    {Desert oasis, palm trees surrounding a pristine pool, an emerald jewel amid golden sands—an Arabian mirage.}
    {Alpine village, snow-covered rooftops, nestled between majestic peaks—a picture-perfect scene of winter tranquility.}
    \row
    {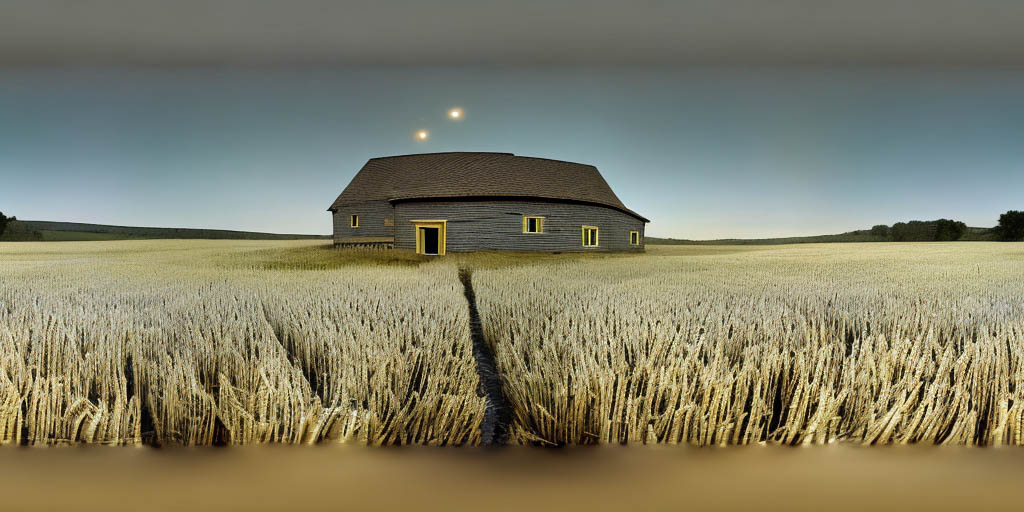}
    {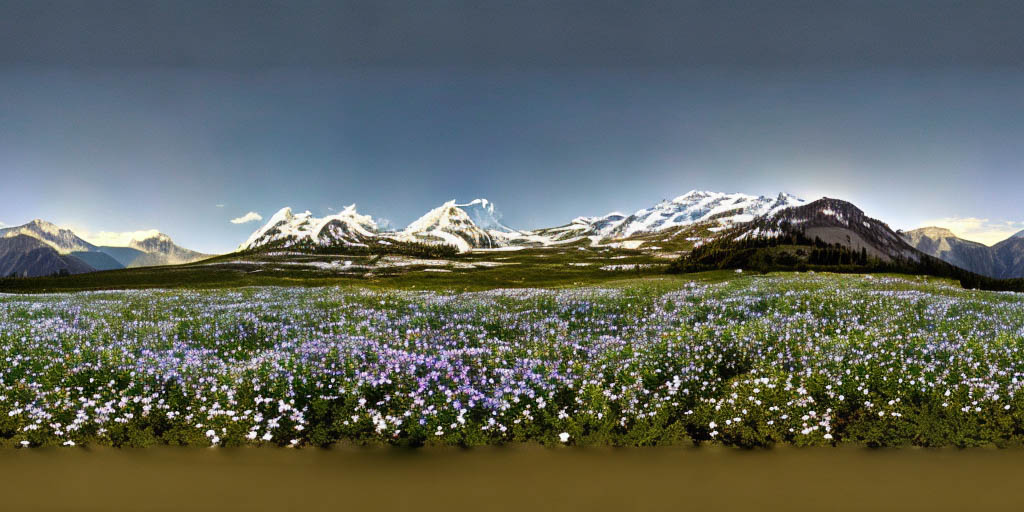}
    {Rustic farmhouse, weathered by time, surrounded by fields of golden wheat—a pastoral scene capturing the essence of simplicity.}
    {Alpine meadow, wildflowers swaying in a mountain breeze, snow-capped peaks embracing a serene panorama—a high-altitude sanctuary.}
}{2}

\genfig
{
    \row
    {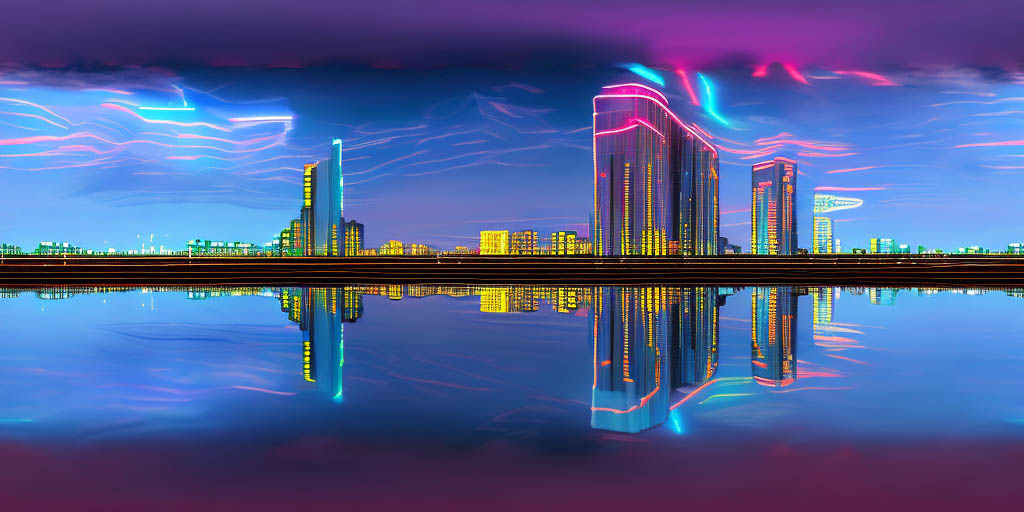}
    {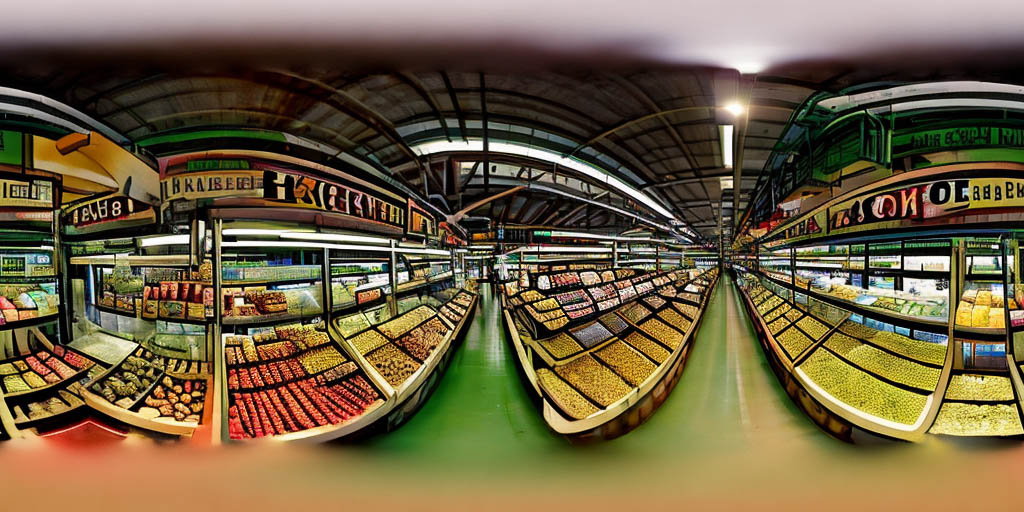}
    {A futuristic cityscape with floating skyscrapers and neon lights reflected in a calm river.}
    {In the heart of a bustling market, the aroma of exotic spices mingles with the vibrant colors of fresh produce.}
    \row
    {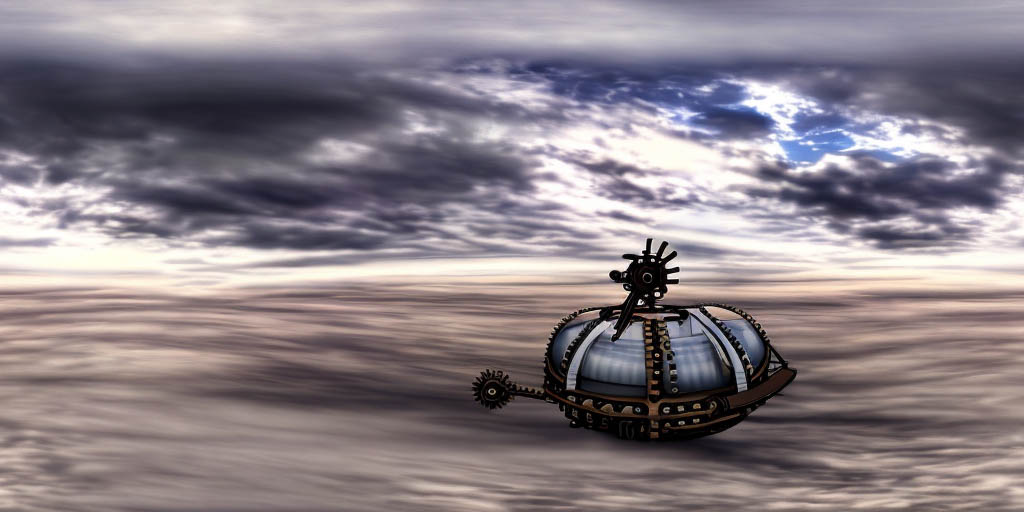}
    {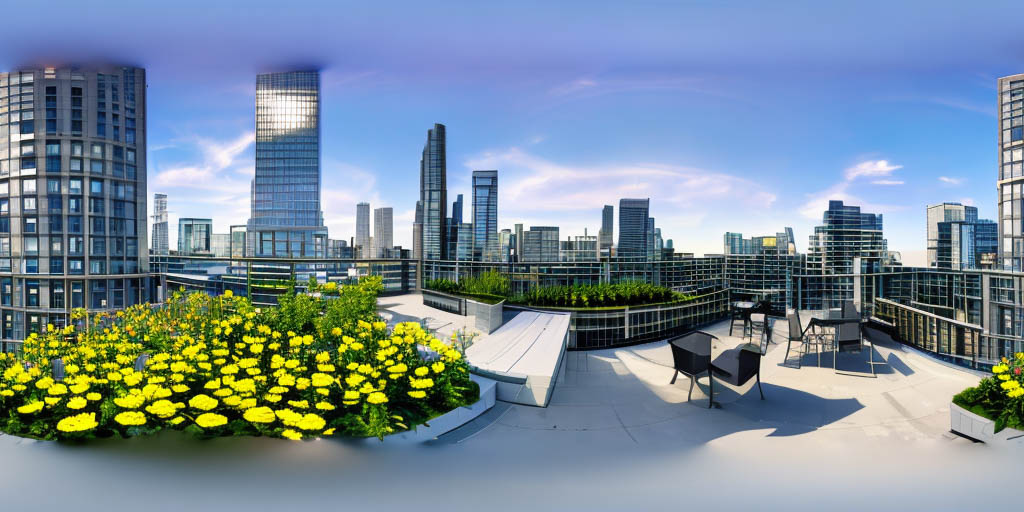}
    {Steampunk airship, navigating cloudy skies, gears turning, propellers whirring.}
    {Urban rooftop garden, vibrant blooms against a backdrop of skyscrapers, a green refuge amid concrete and steel.}
    \row
    {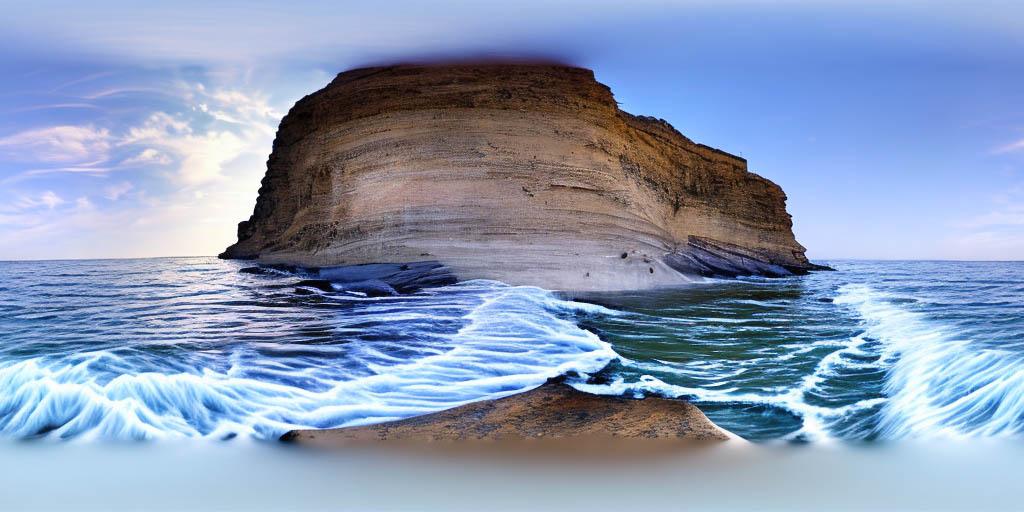}
    {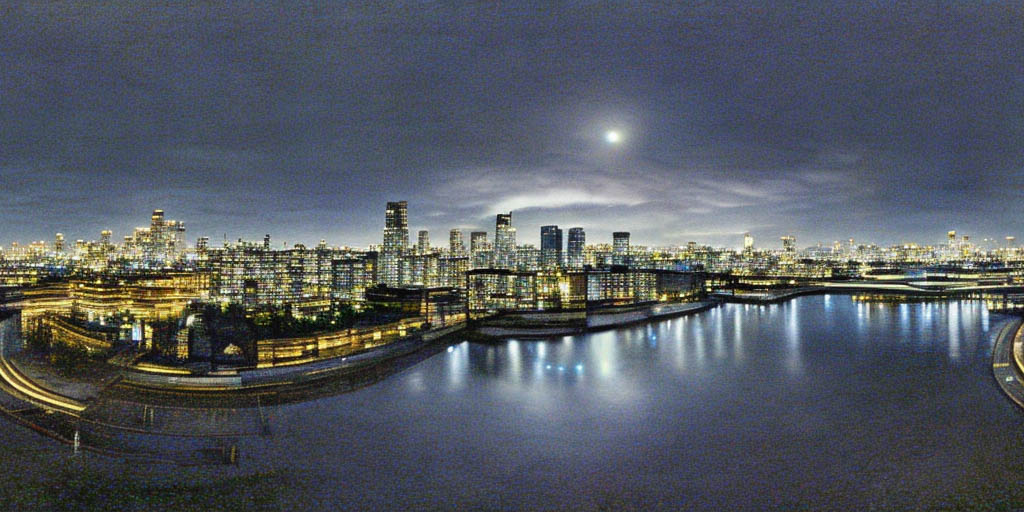}
    {Coastal cliffside, waves crashing on rugged rocks, seagulls soaring in the salty breeze—a dramatic meeting of land and sea.}
    {Moonlit cityscape, reflections shimmering on rain-kissed streets, a quiet metropolis under the night sky—an urban nocturne.}
    \row
    {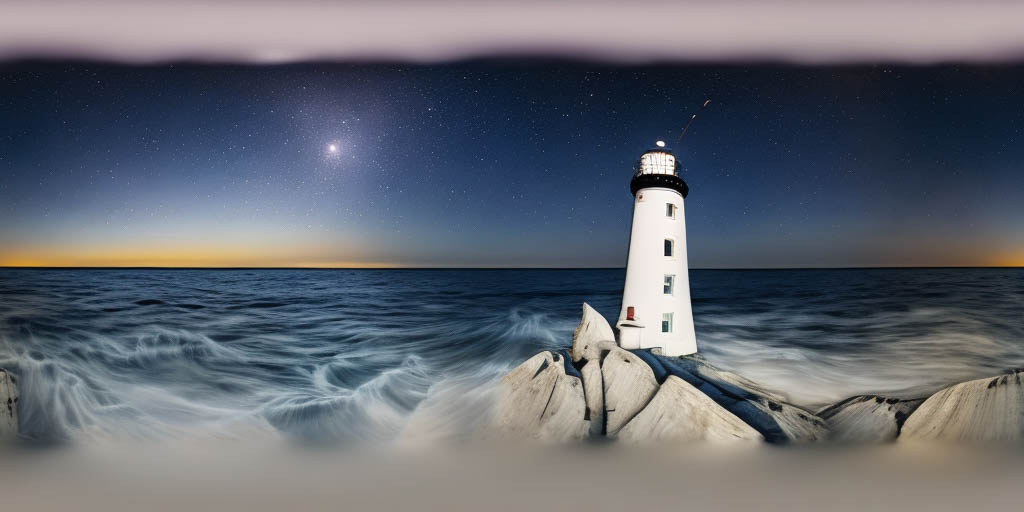}
    {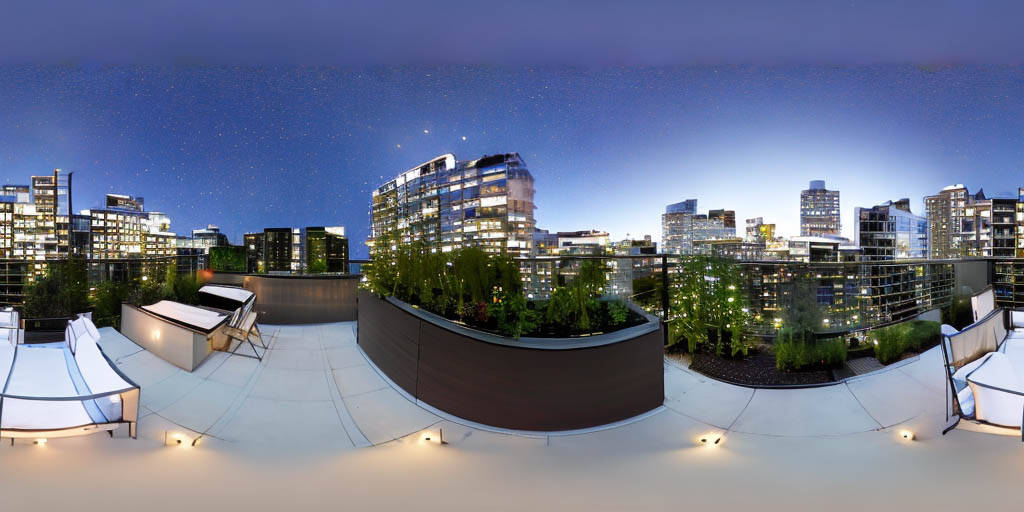}
    {Coastal lighthouse, guiding ships through the moonlit night.}
    {Rooftop garden, city lights below, a quiet urban oasis.}
    \row
    {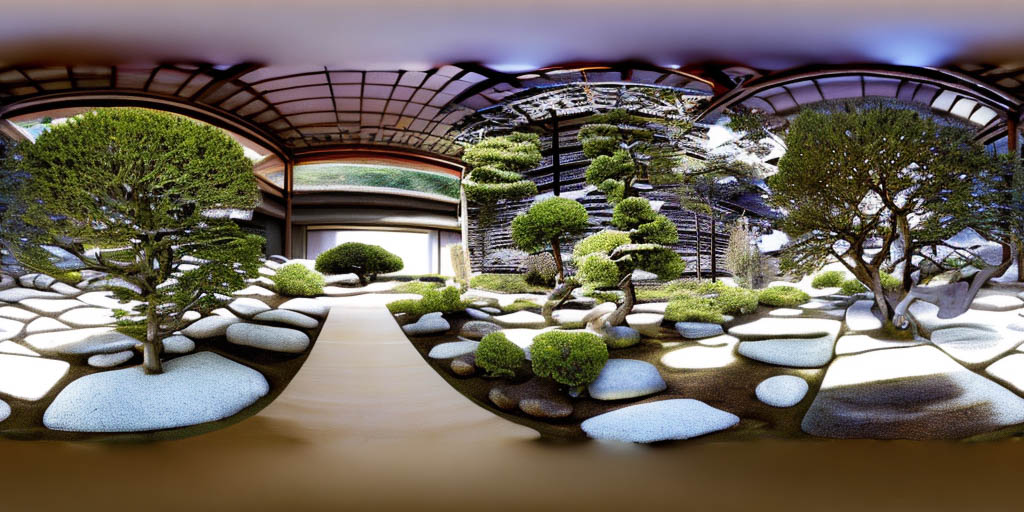}
    {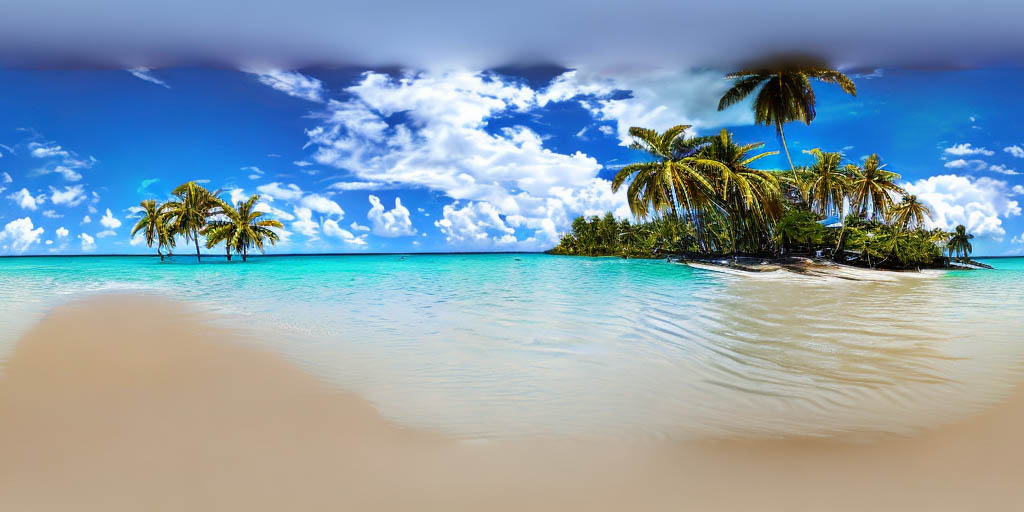}
    {Zen garden, raked pebbles, and bonsai trees—a serene oasis.}
    {Tropical paradise, palm trees swaying, turquoise waters lapping sandy shores.}
}{3}

\genfig
{
    \row
    {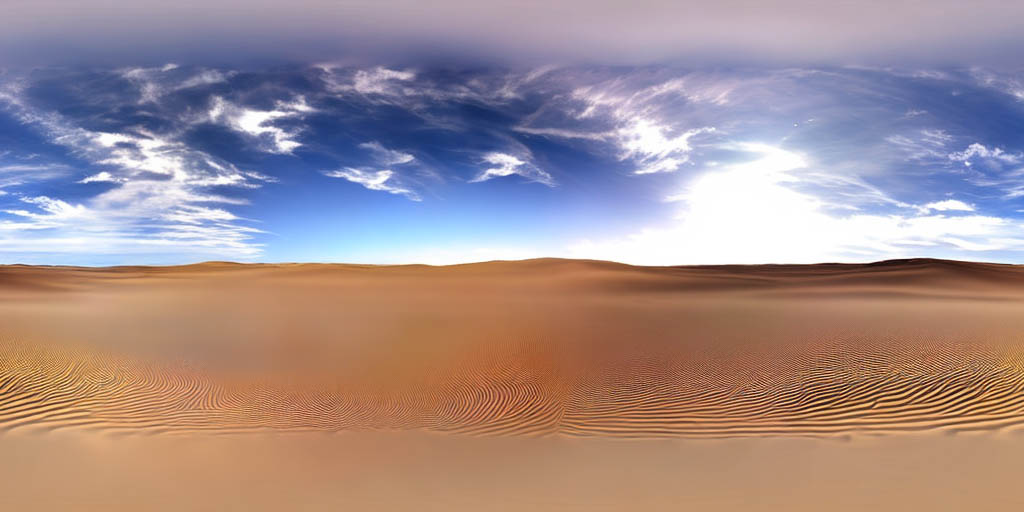}
    {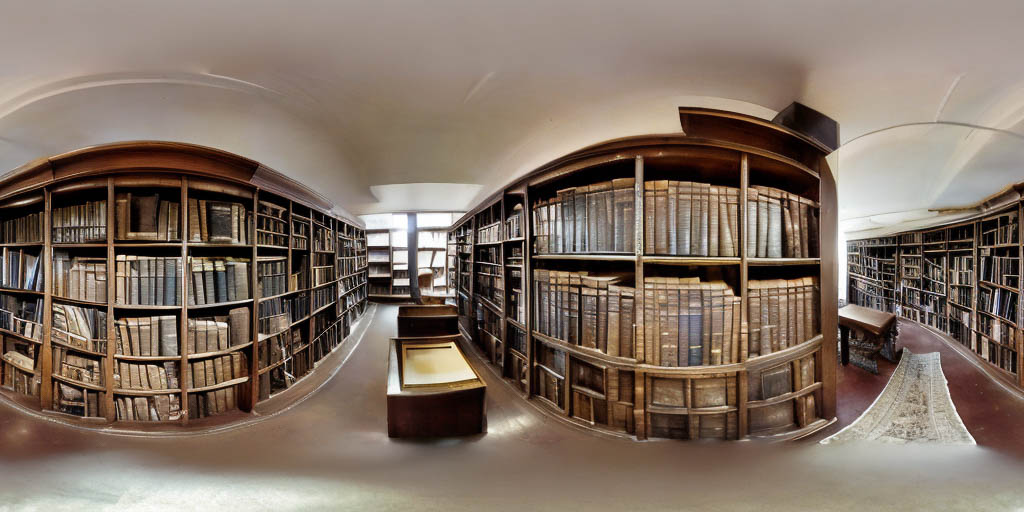}
    {Desert dunes, endless golden waves.}
    {Antique bookstore, leather-bound treasures.}
    \row
    {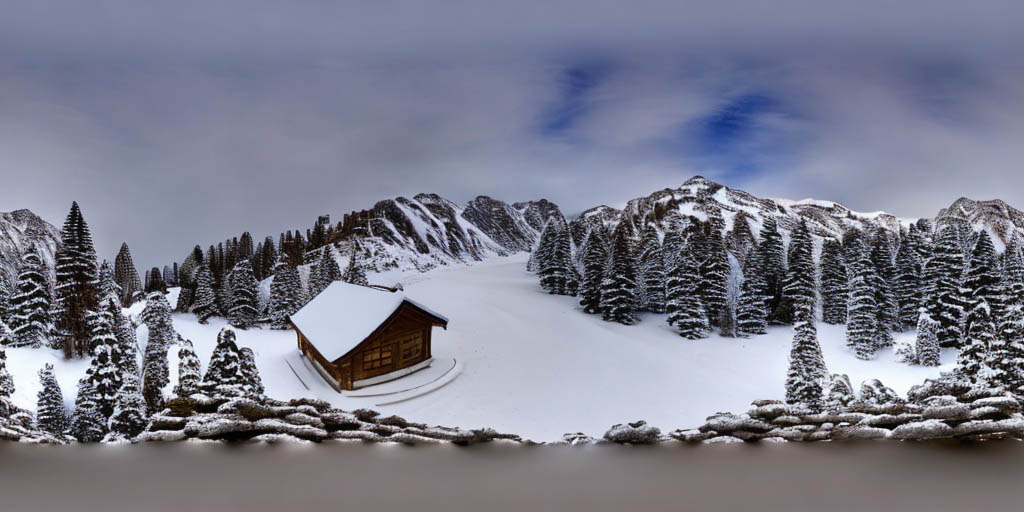}
    {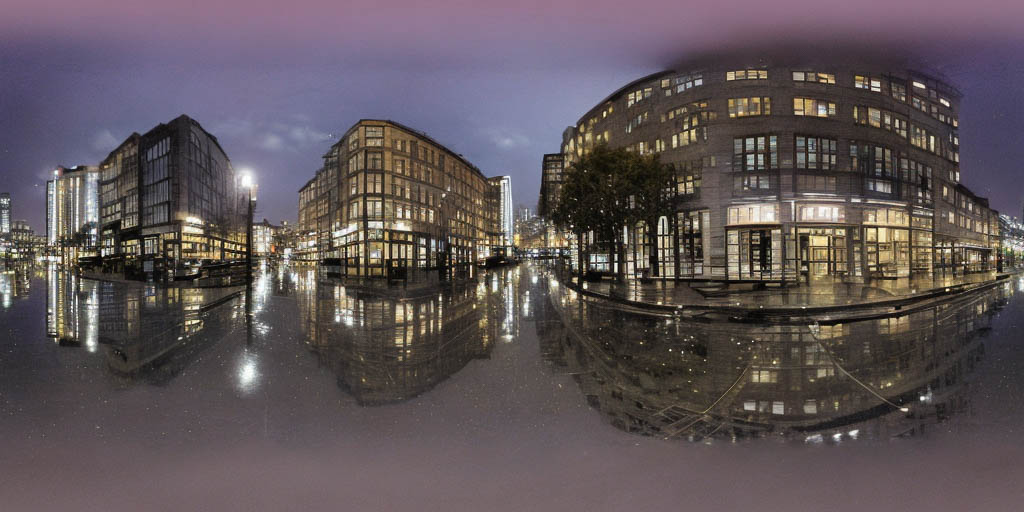}
    {Alpine cabin, snow-capped serenity.}
    {Rain-soaked city streets, glistening reflections.}
    \row
    {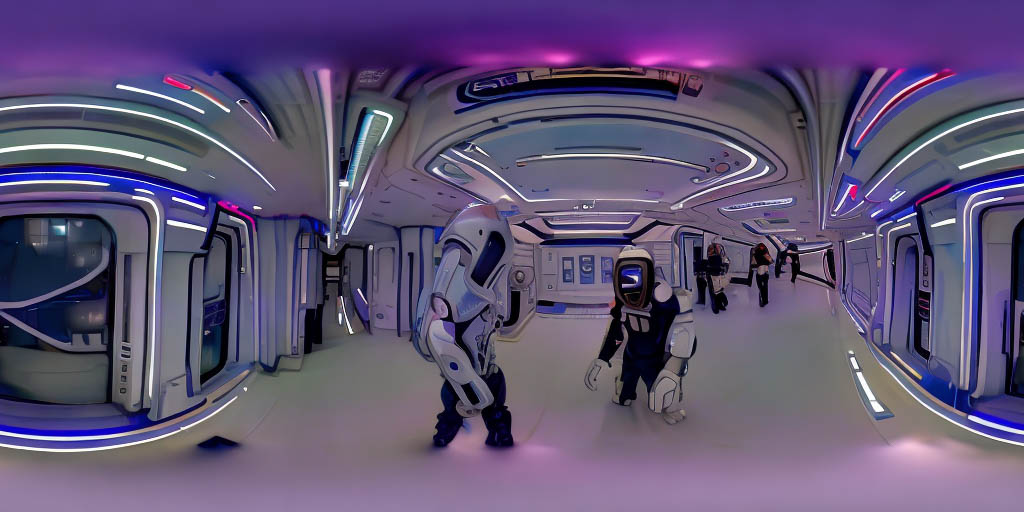}
    {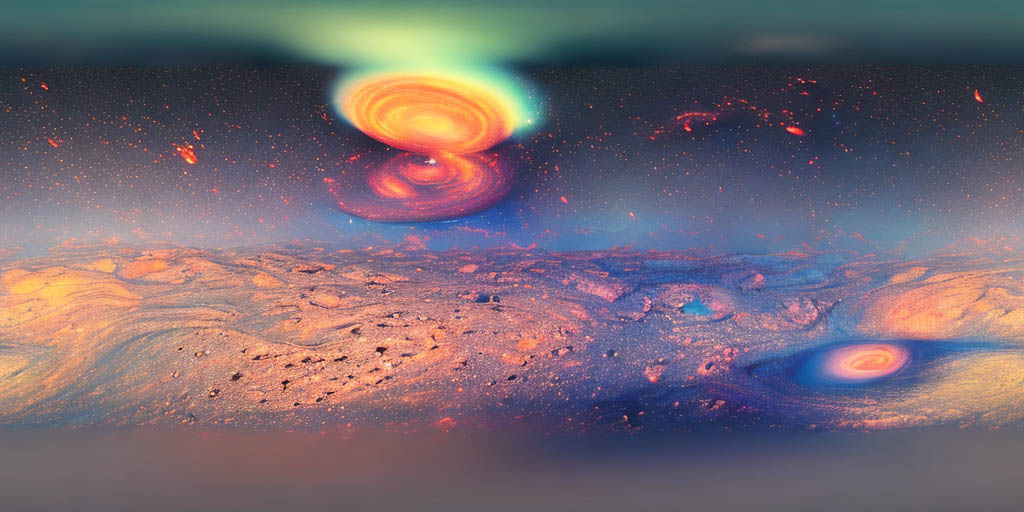}
    {Inside a bustling space station, people from different galaxies interact amid futuristic architecture and advanced robotics.}
    {On the surface of a distant planet, a landscape of alien rock formations and swirling, multicolored gases.}
    \row
    {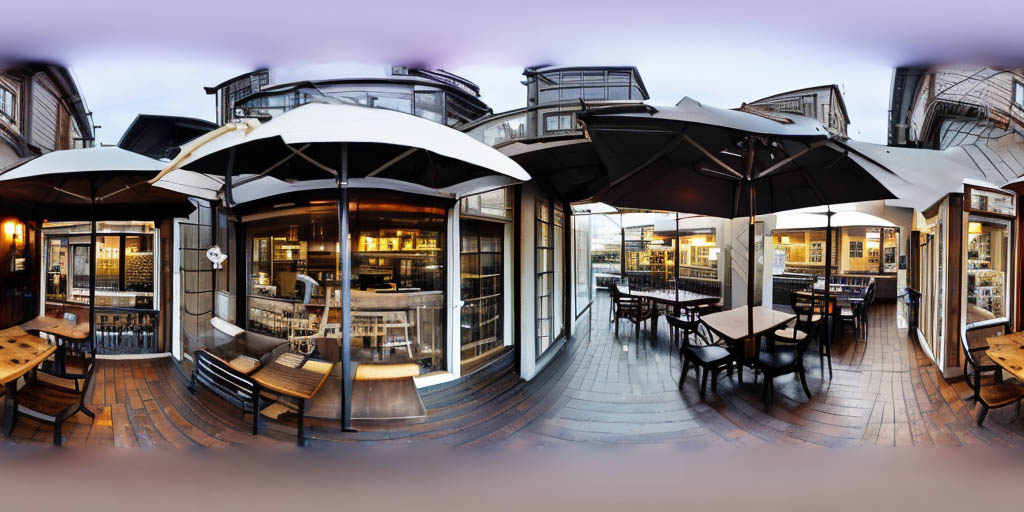}
    {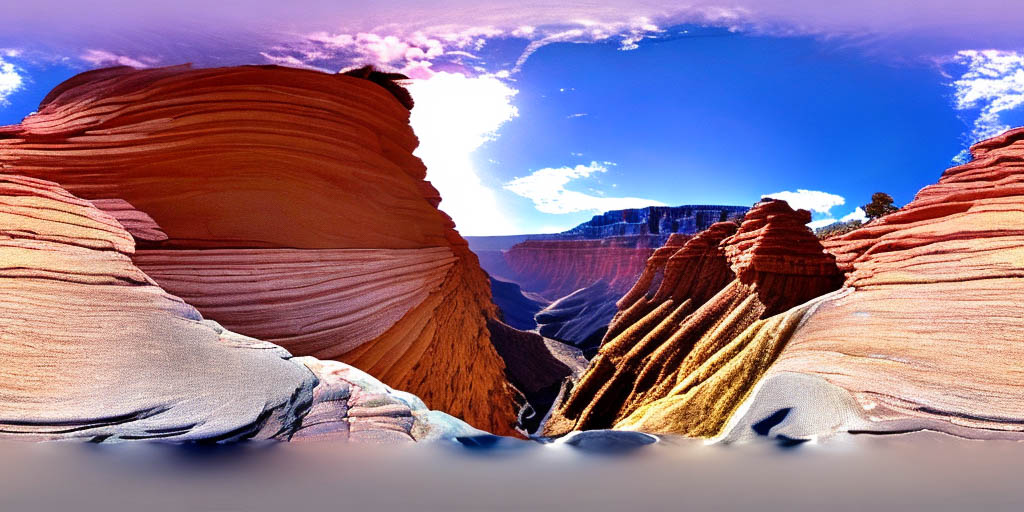}
    {A cozy coffee shop on a rainy day, with the comforting scent of freshly brewed coffee and the sound of rain on the windows.}
    {Standing on the edge of the Grand Canyon, marveling at the vastness of the canyon and the layers of colorful rock formations.}
    \row
    {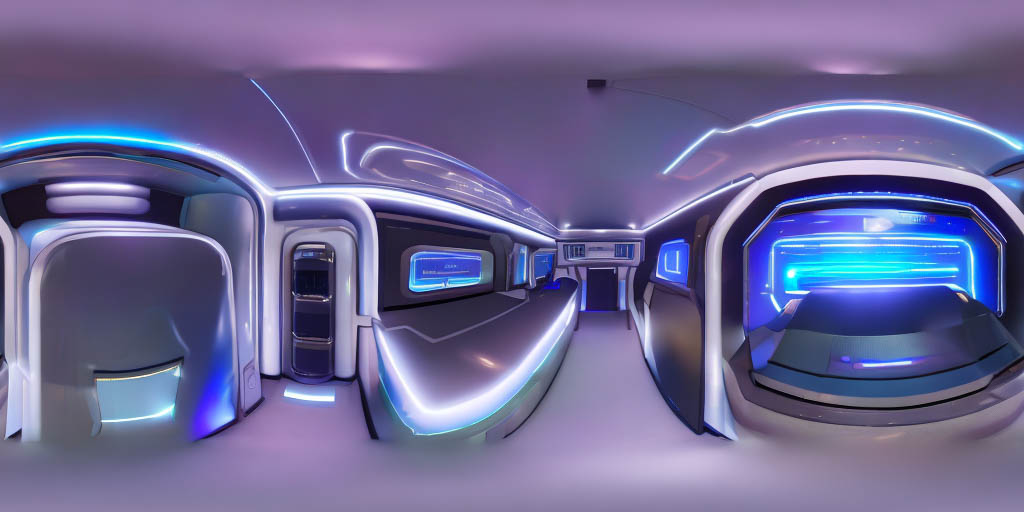}
    {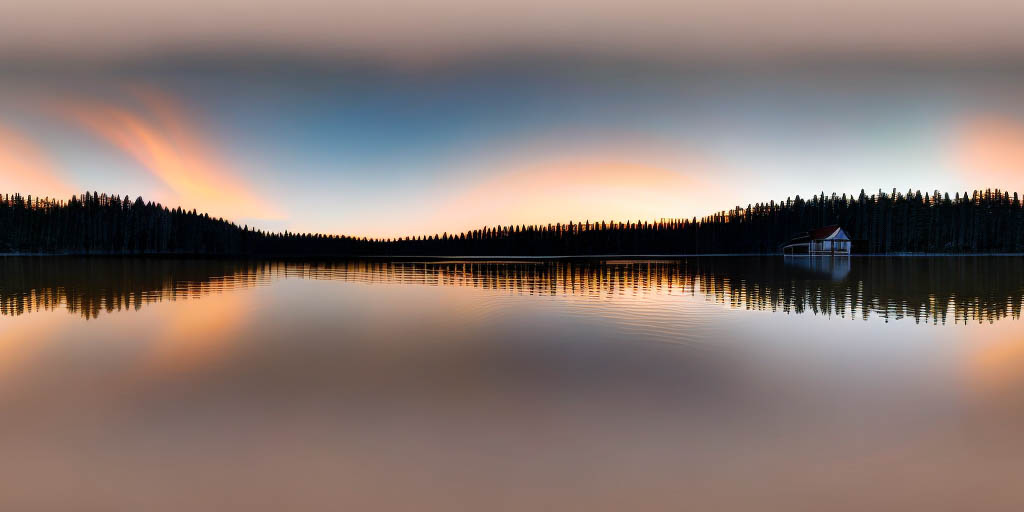}
    {A spaceship interior adorned with holographic displays, sleek metallic surfaces, and advanced technology.}
    {A serene lakeside cabin at dawn, with mist rising from the water and the first light of the day illuminating the landscape.}
}{4}

}

While our method is trained on the Matterport3D dataset, which contains mostly indoor scenes, we show that it can generalize to out-domain prompts and transfer its knowledge of layout understanding to outdoor scenes, as shown in \crefrange{fig:generalization_1}{fig:generalization_4}.

\section{Future Works}\label{sec:suppl_future}
Future works might include introducing more controls over the style and content of the generated panorama images to support applications like virtual house tour, or extending the method to enable outpainting by exploiting the perspective branch to extract guidance from the input image.
The dual-branch architecture can also potentially benefit texture generation for 3D models, where the global branch can operate on UV maps and the perspective branch can operate on rendered images.

\end{document}